\documentclass[10pt,conference,a4paper]{IEEEtran}

\usepackage{times}
\usepackage{epsfig}
\usepackage{graphicx}
\usepackage{subcaption}
\usepackage{amsmath}
\usepackage{amssymb}
\usepackage{calrsfs}
\usepackage{algorithm}
\usepackage{algorithmic}
\usepackage{microtype}
\usepackage{float}
\usepackage{alphalph}

\usepackage[noadjust]{cite}

\def\eg{\emph{e.g.}}
\def\ie{\emph{i.e.}}
\def\etal{\emph{et al.}}

\DeclareMathAlphabet{\pazocal}{OMS}{zplm}{m}{n}

\usepackage[pagebackref=true,breaklinks=true,colorlinks,bookmarks=false]{hyperref}

\begin{document}
\title{Partial Membership Latent Dirichlet Allocation for Image Segmentation}



\author{\IEEEauthorblockN{Chao Chen}
\IEEEauthorblockA{ Electrical \& Computer Engineering\\
University of Missouri\\
Email: ccwwf@mail.missouri.edu}
\and
\IEEEauthorblockN{Alina Zare}
\IEEEauthorblockA{ Electrical \& Computer Engineering\\
University of Missouri\\
Email: zarea@missouri.edu}
\and
\IEEEauthorblockN{J. Tory Cobb}
\IEEEauthorblockA{Naval Surface Warfare Center\\
Panama City, FL\\
Email: james.cobb@navy.mil}}

\maketitle


\begin{abstract}

Topic models (\eg, pLSA, LDA, SLDA) have been widely used for segmenting imagery. These models are confined to crisp segmentation.  Yet, there are many images in which some regions cannot be assigned a crisp label (\eg, transition regions between a foggy sky and the ground or between sand and water at a beach).  In these cases, a visual word is best represented with partial memberships across multiple topics.  To address this, we present a partial membership latent Dirichlet allocation (PM-LDA) model and associated  parameter estimation algorithms. 
Experimental results on two natural image datasets and one SONAR image dataset show that PM-LDA can produce both crisp and soft semantic image segmentations; a capability existing methods do not have. 

\end{abstract}

\section{Introduction}

\begin{figure}[!htb]
\centering
\begin{subfigure}[b]{0.15\textwidth}
\includegraphics[width=1\linewidth,height=0.8\linewidth]{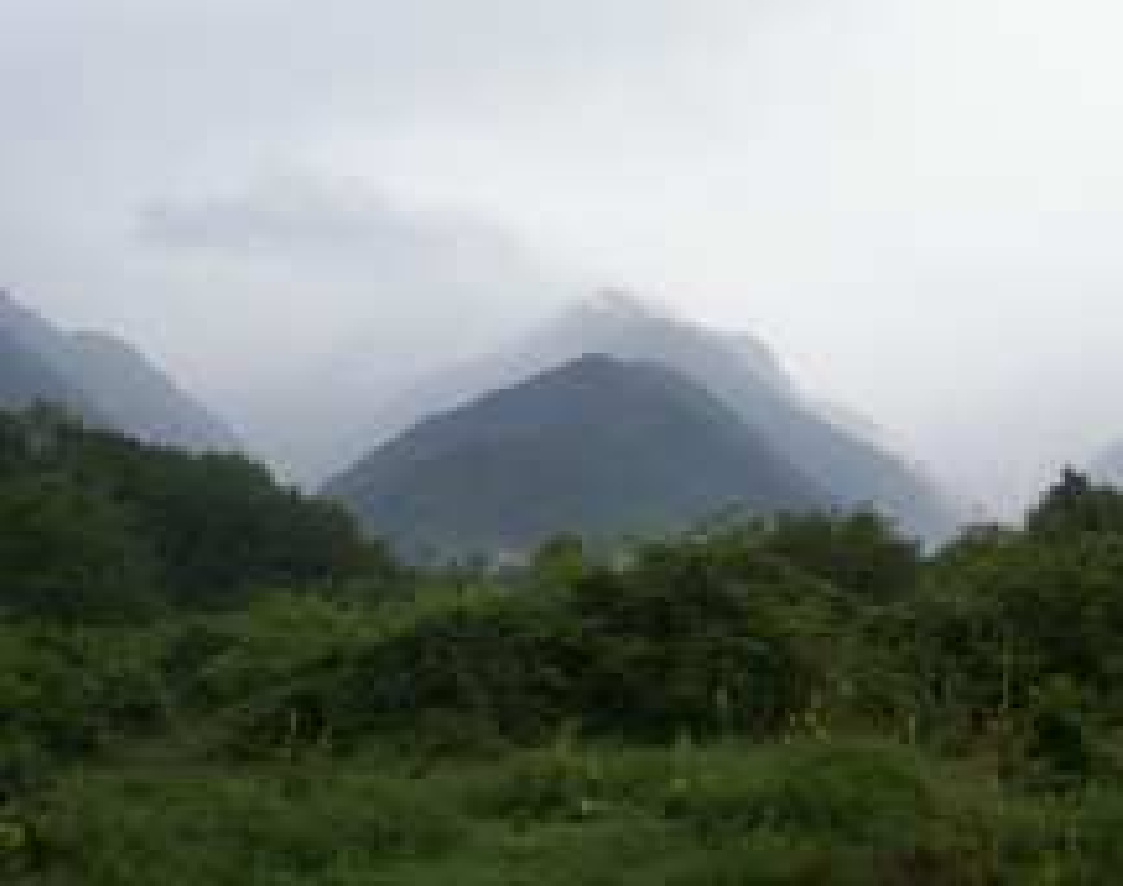}
\captionsetup{labelformat=empty,,skip=0pt}
\caption{(a)}
\label{fig:afog}
\end{subfigure}
\begin{subfigure}[b]{0.15\textwidth}
\includegraphics[width=1\linewidth,height=0.8\linewidth]{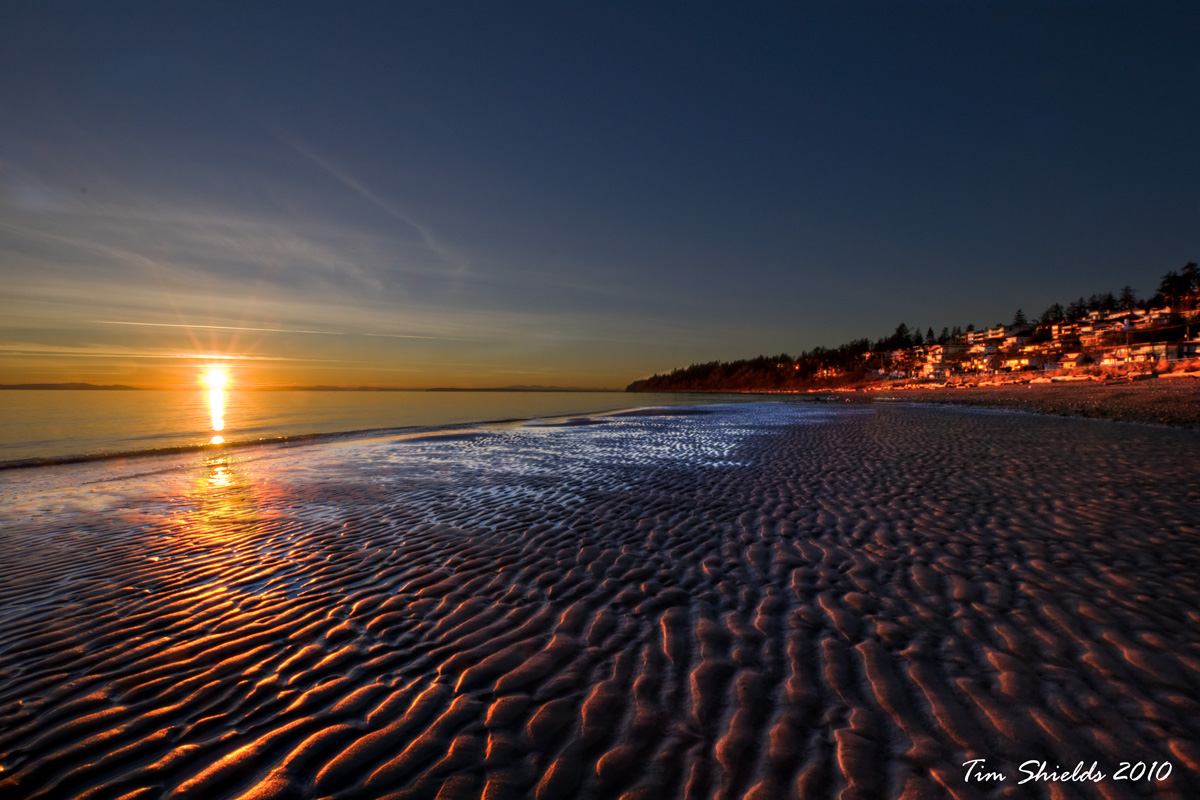}
\captionsetup{labelformat=empty,skip=0pt}
\caption{(b)}
\label{fig:bsun}
\end{subfigure}
\begin{subfigure}[b]{0.15\textwidth}
\includegraphics[width=1\linewidth,height=0.8\linewidth]{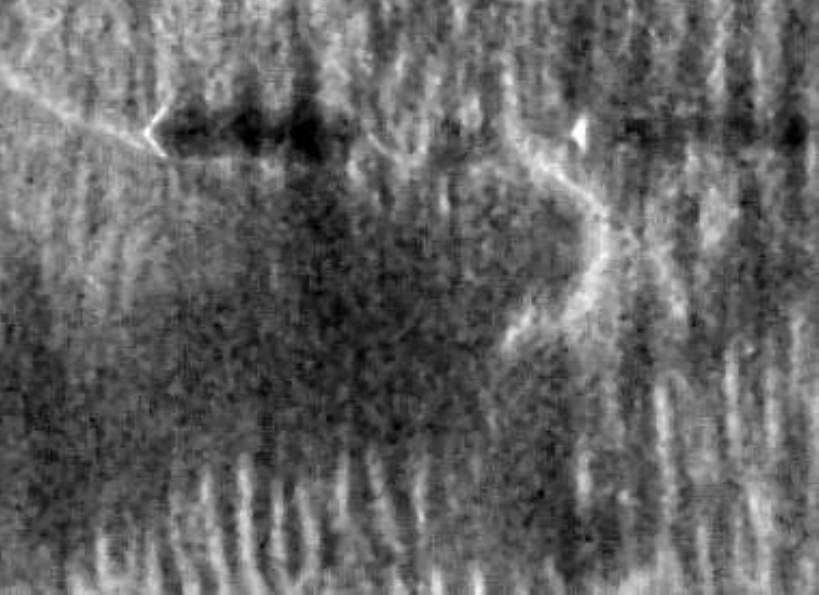}
\captionsetup{labelformat=empty,skip=0pt}
\caption{(c)}
\label{fig:csonar}
\end{subfigure}
\caption{Imagery with regions of gradual transition. (a) Image with gradual transition from fog to mountain. 
(b) Sunset image with gradual transition from sun to sky. 
(c) SONAR image with gradually vanishing sand ripples. }
\label{fig:sunset}
\vspace{-4mm}
\end{figure}

The goal of unsupervised semantic image segmentation is to divide an image into semantically distinct coherent regions, \ie, regions corresponding to objects or parts of objects. It plays an important role in a wide range of computer vision applications, such as object recognition and tracking and image retrieval \cite{zhao:2010,shi:2000,comaniciu:2002, felzenszwalb:2004}.  Yet, in many images, widely used image segmentation methods fail to perform.  Specifically, imagery in which there are smooth gradients and transition regions are poorly-addressed with the many crisp image segmentation methods in the literature.    For example, consider the photograph in Fig. \ref{fig:afog} where the gradually thinning fog blurs the boundary between the foggy sky and the mountain, a sharp boundary between the ``fog'' and ``mountain'' topics does not exist. Similarly, in Fig. \ref{fig:bsun} consider the gradually fading sunlight or in Fig. \ref{fig:csonar} consider the gradually vanishing sand ripples shown in the Synthetic Aperture SONAR image of the sea floor.  In both of these cases, sharp boundaries between the ``sun'' and ``sky'' topics or ``sand ripple'' and ``flat sand'' topics do not exist.  In this paper, we present a Partial Membership Latent Dirichlet Allocation (PM-LDA) to address image segmentation in the case of gradients and regions of transition.

%

Inspired by the success of Latent Dirichlet Allocation (LDA) \cite{blei:2003} in discovering semantically meaningful topics from document collections, many have successfully applied LDA or its variants to image segmentation \cite{russell:2006, cao:2007, wang:2008, zhao:2010,andreetto:2012}.  These unsupervised semantic image segmentation methods differ from traditional (\ie, non-hierarchical/flat) segmentation methods  (\eg, normalized cuts algorithm \cite{shi:2000}) by estimating and describing additional inter-segment relationships. Namely, unsupervised semantic image segmentation methods over-segment imagery and then group these ``visual words'' into topic clusters such that small segments from the same object class can be combined into a complete object and provide a comprehensive organization of the larger scene.   In other words, unsupervised semantic image segmentation methods cluster imagery hierarchically in which the lower level corresponds to an over-segmentation of the imagery and the higher level groups the over-segmented pieces into topic clusters.  

\begin{figure*}[htb!]
  \begin{subfigure}{0.3\textwidth}
     \centering
     \includegraphics[width=0.8\linewidth]{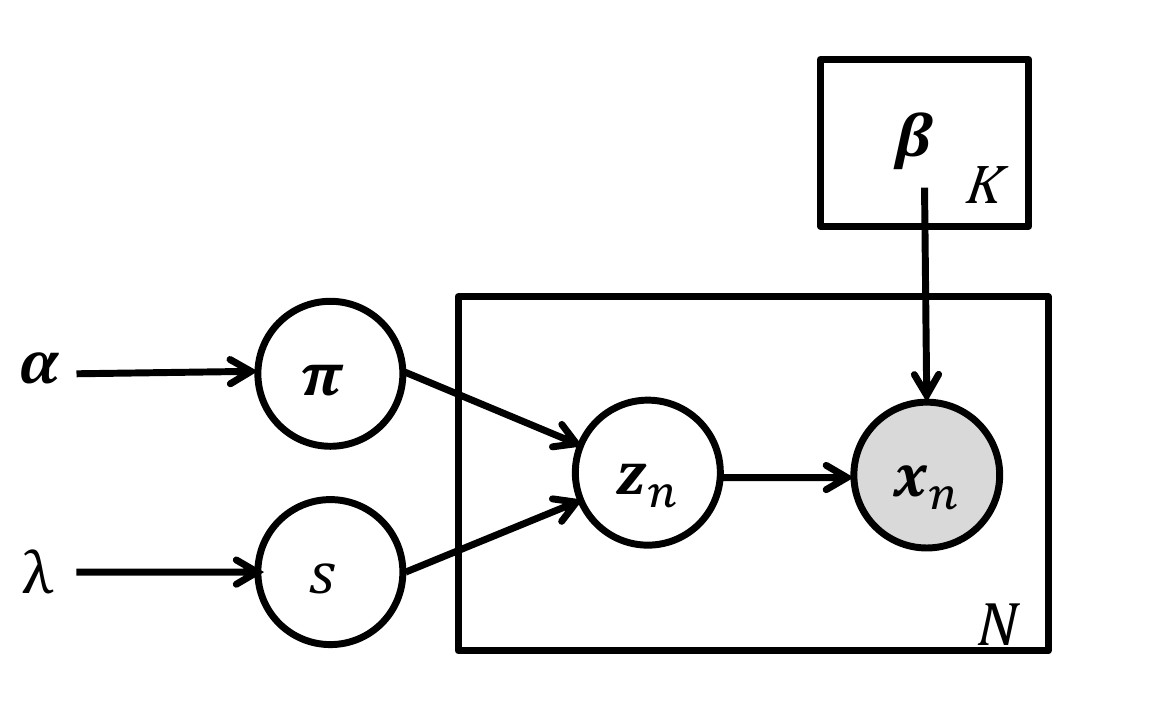}
     \captionsetup{skip=0pt}
     \caption{BPM}
     \label{fig:bmp}
  \end{subfigure}
  \begin{subfigure}{0.3\textwidth}
     \centering
     \includegraphics[width=0.8\linewidth]{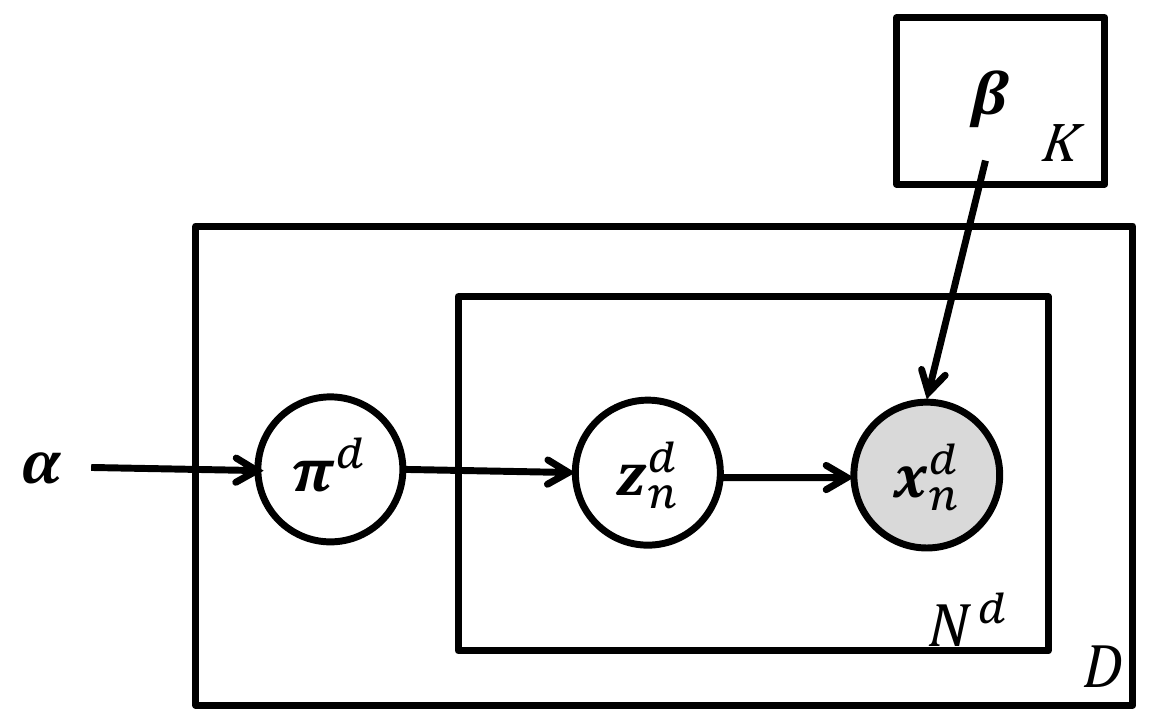}
    \captionsetup{skip=0pt}
     \caption{LDA}
     \label{fig:lda}
  \end{subfigure}
  \begin{subfigure}{0.3\textwidth}
     \centering
     \includegraphics[width=0.8\linewidth]{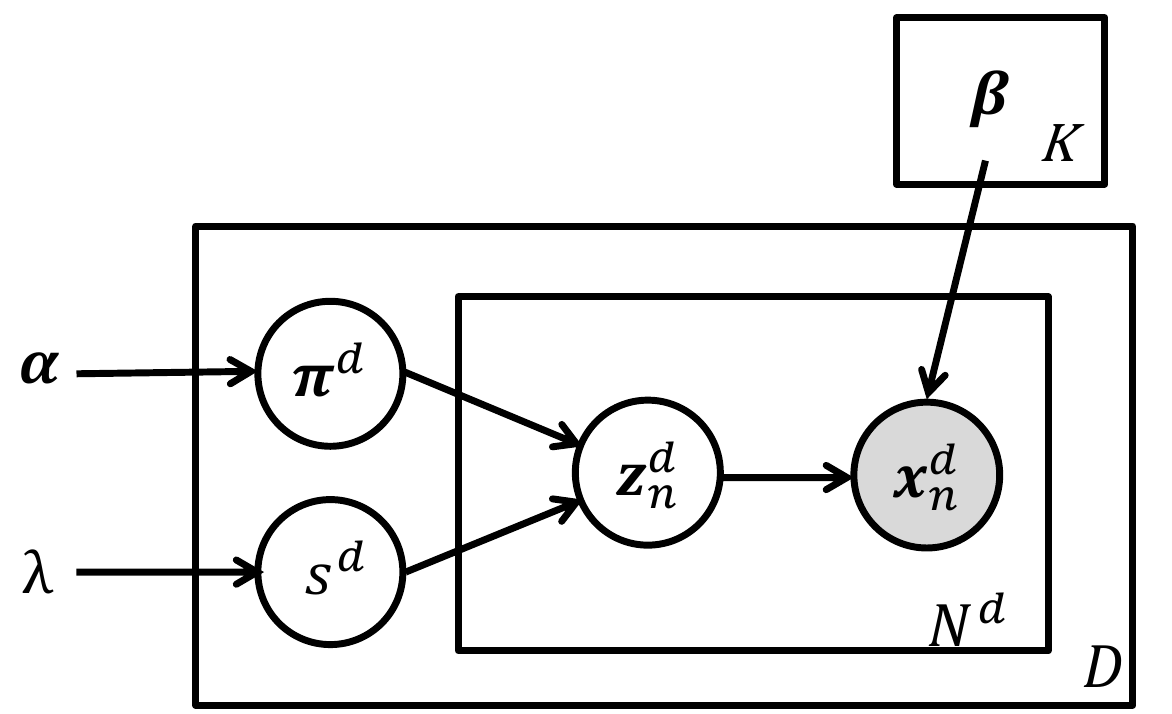}
    \captionsetup{skip=0pt}
     \caption{PM-LDA}
     \label{fig:pmlda}
  \end{subfigure}
  \caption{(a) Graphical model for BPM. (b) Graphical model for LDA. (c) Graphical model for PM-LDA}
\end{figure*}

However, under these existing topic models, a visual word is only assigned to one topic (\ie, the word-topic assignment is a binary indicator).   Yet, in many images, it is impossible to assign a crisp boundary between topics.  In this paper, we generalize LDA to allow for partial memberships.

Partial membership models and algorithms have been previously developed in the literature.  One prevalent partial membership approach is the Fuzzy C-means algorithm (FCM) \cite{bezdek:1984}. 
More recently, Heller \etal \cite{heller:2008} and Glenn \etal \cite{GlennZare:2014} proposed Bayesian partial membership models (\ie, probabilistic interpretations of FCM) along with their associated generative processes. Glenn \etal \cite{GlennZare:2014} proposed a Bayesian Fuzzy Clustering (BFC) model and associated algorithms to bridge and extend  probabilistic clustering and fuzzy clustering methods. 
Heller \etal \cite{heller:2008} derived a Bayesian partial membership model (BPM) by extending the standard mixture model. The generative processes for both the BFC and BPM models are similar. The main difference between the BPM and BFC models is that the BFC uses both a fixed \emph{fuzzifier} parameter and a scaling parameter to control the degree of mixing between topics.  In contrast, in the BPM, the degree of mixing between topics is controlled only through a scaling hyper-parameter, $s$, found in the prior distribution on partial membership values. 
These existing partial membership methods are non-hierarchical approaches.  The proposed PM-LDA extends these to hierarchical approaches to allow for semantic image segmentation. PM-LDA is an extension of both the LDA model proposed by Blei \etal \cite{blei:2003} and the partial membership model proposed by Heller \etal \cite{heller:2008}.



\section{Bayesian Partial Membership Model} \label{sec:heller}

In a finite mixture model, the data likelihood of $\mathbf{x}_n$,  is
\begin{equation}
p(\mathbf{x}_n|\boldsymbol{\beta})=\sum_{k=1}^{K}\pi_kp_k(\mathbf{x}_n|\beta_k),
\end{equation}
where $\left\{ \pi_k \right\}_{k=1}^K$  are the mixture weights and ${\boldsymbol{\beta}}=\{{\beta}_1, {\beta}_2, ..., {\beta}_K\}$ are the mixture component parameters. $p_k(\mathbf{x}_n|\beta_k)$ is the $k^{th}$ mixture component with parameters ${\beta}_k$. In this model, a data point is assumed to come from one (and only one) of the $K$ mixture components. Thus, given its component assignment, $\mathbf{z}_n$, the probability of a data point, $\mathbf{x}_n$,  is defined as $p(\mathbf{x}_n|\mathbf{z}_n,\boldsymbol{\beta})=\prod_{k=1}^{K}p_k(\mathbf{x}_n|\beta_k)^{z_{nk}}$ where $
z_{nk} \in \{0,1\}$, $\sum_{k=1}^{K}z_{nk}=1$, 
and $\mathbf{z}_n=[z_{n1}, z_{n2}, ..., z_{nK}]$ is the binary membership vector.   If $z_{nk}=1$, the data point $\mathbf{x}_n$ is assumed to have been drawn from mixture component $k$. 
 
 In order to obtain a model allowing multiple cluster memberships for a data point, the constraint $z_{nk} \in \{0, 1\}$ is relaxed to $z_{nk} \in [0,1]$.
%
%
The modified constraints and the inclusion of prior distributions for several  key parameters results in the Bayesian Partial Membership model \cite{heller:2008}, 
\begin{eqnarray}
p(\boldsymbol{\pi},s,\mathbf{z}_n, \mathbf{x}_n|\boldsymbol{\alpha},\lambda,\boldsymbol{\beta})&=&p(\boldsymbol{\pi}|\boldsymbol{\alpha})p(s|\lambda)p(\mathbf{z}_n|\boldsymbol{\pi} s) \nonumber \\
&&\prod_{k=1}^{K}p_k(\mathbf{x}_n|\beta_k)^{z_{nk}}, 
\label{eqn:bpm}
\end{eqnarray}
where $z_{nk}\in[0,1]$, $\sum_{k=1}^{K}z_{nk}=1$, and $\boldsymbol{\pi}$ is the cluster mixing proportion assumed to be distributed according to a Dirichlet distribution with parameter $\boldsymbol{\alpha}$, \emph{i.e.,} $\boldsymbol{\pi} \sim \text{Dir}(\boldsymbol{\alpha})$. 
 The scaling factor, $s$, determines the level of cluster mixing and is distributed according to an exponential distribution with mean $1/\lambda$, $s \sim \text{exp}(\lambda)$, and
$\mathbf{z}_n \sim \text{Dir}(\boldsymbol{\pi} s)$ is the membership vector for data point $\mathbf{x}_n$. 


As shown in \cite{heller:2008}, if each of the mixture components are exponential family distributions of the same type, then $p(\mathbf{x}_n|\mathbf{z}_n,\boldsymbol{\beta})=\prod_{k=1}^{K}p_k(\mathbf{x}_n|\beta_k)^{z_{nk}}$ with $z_{nk} \in [0,1]$ and $\sum_{k=1}^{K}z_{nk}=1$, can be written as:
\begin{equation}
p(\mathbf{x}_n|\mathbf{z}_n,\boldsymbol{\beta}) = \text{Expon}\left(\sum_k z_{nk}{\eta}_k\right).
\label{eqn:expfamprod}
\end{equation}
This indicates that the data generating distribution for $\mathbf{x}_n$ is of the same exponential family distribution as the original $K$ clusters, but with new natural parameters $\sum_k z_{nk}{\eta}_k$.  The new parameters are a convex combination of the natural parameters, $\eta_k$, of the original clusters weighted by $z_{nk}$. This provides the powerful (and convenient) ability to sample directly from the unique mixture distribution for each data point if the natural parameters of the original clusters and the membership vector for the data point are known. A graphical model of BPM is shown in Fig. \ref{fig:bmp}. 

\section{Partial Membership LDA}  \label{sec:PMLDA}

In the BPM,  data points are organized at only one level, where each data point is indexed by its corresponding component distribution. In our proposed model, PM-LDA, (and in LDA) data is organized at two levels: the word level and the document level as illustrated in Fig. \ref{fig:lda} and \ref{fig:pmlda}.  In the proposed PM-LDA model, the random variable associated with a data point is assumed to be distributed according to multiple topics with a continuous partial membership in each topic. Specifically, the PM-LDA model is
\begin{eqnarray}
p(\boldsymbol{\pi}^d,s^d, \mathbf{z}^d_n,\mathbf{x}^d_n|\boldsymbol{\alpha},\lambda,\boldsymbol{\beta})&=&p(\boldsymbol{\pi}^d|\boldsymbol{\alpha})p(s^d|\lambda)p(\mathbf{z}^d_n|\boldsymbol{\pi}^ds^d)\nonumber \\
&& \prod_{k=1}^{K}p_k(\mathbf{x}_n^d|\beta_k)^{{z}^d_{nk}} 
\label{eqn:pmlda}
\end{eqnarray} 
where $\mathbf{x}^d_n$ is the $n$th word in document $d$, $\mathbf{z}^d_n$ is the partial membership vector of $\mathbf{x}^d_n$, $\boldsymbol{\pi}^d \sim \text{Dir}(\boldsymbol{\alpha})$ and $s^d \sim \text{exp}(\lambda)$ are the topic proportion and  the level of topic mixing in document $d$, respectively. The parameter $\boldsymbol{\alpha}$ gives the topic composition across a document. For example, in  Fig. \ref{fig:csonar}, the image may be composed of $40\%$ ``sand ripple'' topic and $60\%$ ``flat sand'' topic (\ie, $\boldsymbol{\alpha} = [0.4, 0.6]$).  The  parameter $\lambda$ controls how similar the partial membership vector of each word is expected to be to the topic distribution of the document. For example, a small $\lambda$ would correspond to most words in an document to have partial membership vectors very close to $\boldsymbol{\pi}^{d}$. During image segmentation, a small $\lambda$ generally corresponds to large transition regions (\eg, transition from ``flat sand'' to ``sand ripple'' comprises most of the image). For a large $\lambda$, the partial membership vectors for each word can vary significantly from the document mixing proportions.  In general, a large $\lambda$ corresponds to very narrow (tending towards crisp) transition regions during image segmentation (\eg, the SAS image may have $39\%$ of the visual words as pure ``sand ripple'', $59\%$ as pure ``flat sand'', and only $2\%$ mixed). 
The vector  $\mathbf{z}^d_{n} \sim \text{Dir}(\boldsymbol{\pi}^{d} s^{d})$ represents the partial memberships of data point $\mathbf{x}^{d}_{n}$ in each of the $K$ topics. 
If each topic distribution is assumed to be of the exponential family, $p_k(\cdot|\beta_k) = \text{Expon}(\eta_k)$, then using the result in \eqref{eqn:expfamprod}, $p(\mathbf{x}_n^d|\mathbf{z}_n^d,\boldsymbol{\beta}) = \text{Expon}(\sum_k z_{nk}^d{\eta}_k)$. The  graphical model  for PM-LDA is shown in Fig. \ref{fig:pmlda}.

\begin{figure}[!tb]
\centering
 \begin{subfigure}[b]{0.22\textwidth}
\centering
\includegraphics[width=0.8\linewidth]{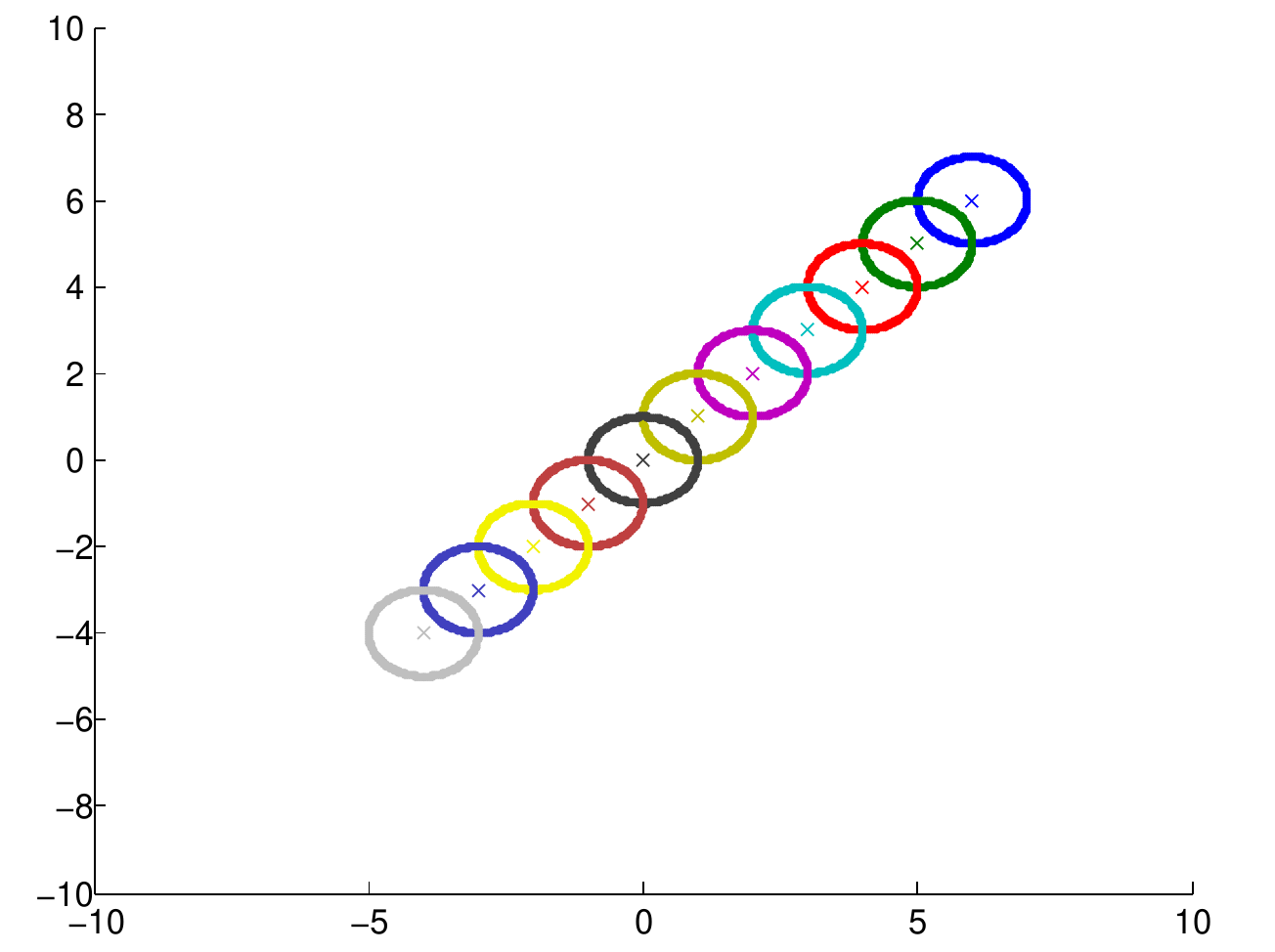}
\captionsetup{labelformat=empty,skip=0pt}
\caption{(a)}
\end{subfigure}
 \begin{subfigure}[b]{0.22\textwidth}
\centering
\includegraphics[width=0.8\linewidth]{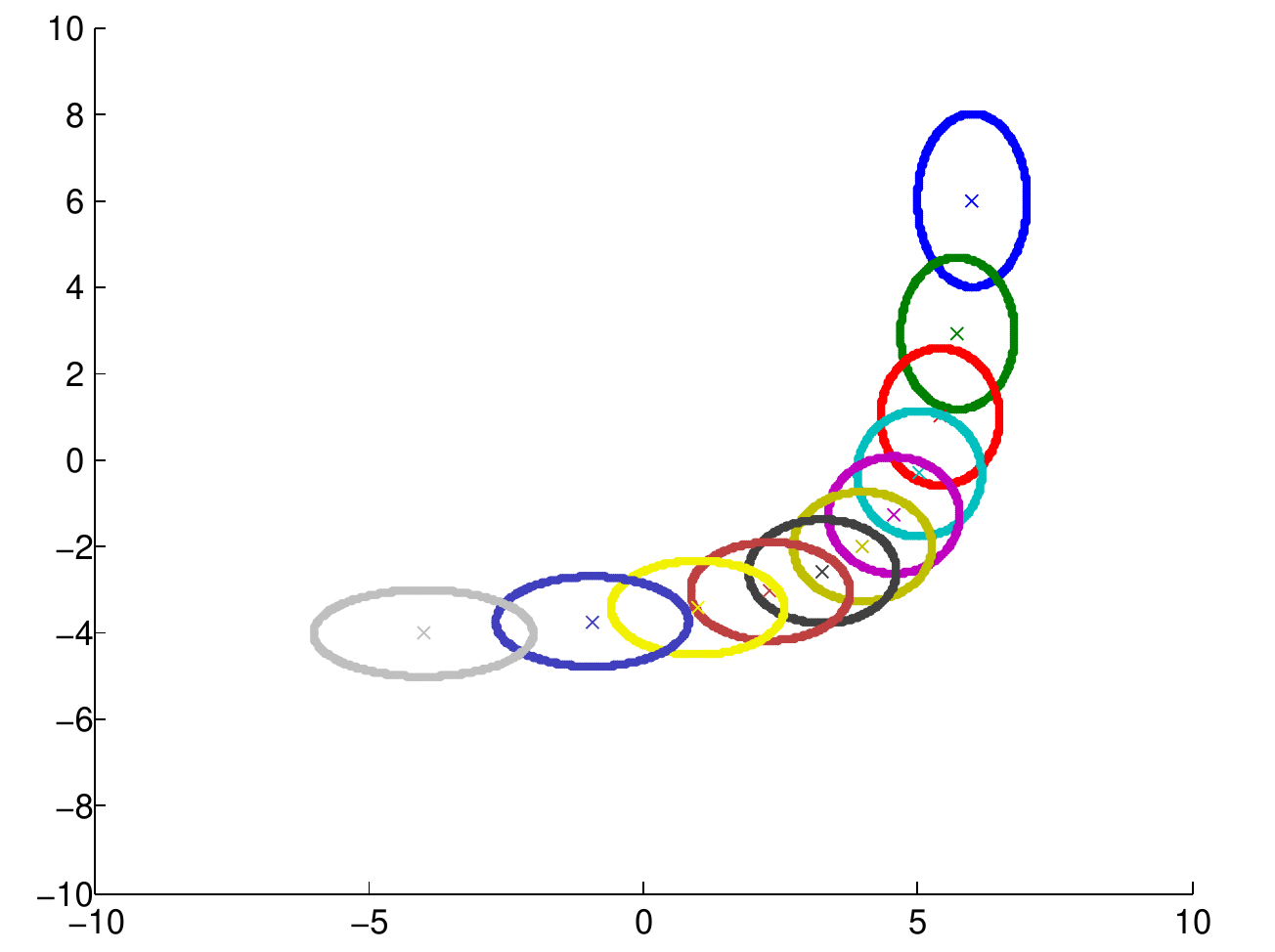}
\captionsetup{labelformat=empty,skip=0pt}
\caption{(b)}
\end{subfigure}
\caption{Partial membership data generating distributions. In (a), two Gaussian topics with $\mu_1=[-4, -4]; \mu_2=[6, 6]$, $\Sigma_1=\Sigma_2=\mathbf{I}$. In (b), two Gaussian topics with  $\mu_1=[-4,-4]; \mu_2=[6, 6]$, $\Sigma_1=[4, 0; 0, 1],\Sigma_2=[1, 0; 0, 4]$ \cite{heller:2008}.}
\vspace{-4mm}
\label{fig:topics}
\end{figure}

 In PM-LDA, the membership $\mathbf{z}^d_{n}$ is drawn from a Dirichlet distribution which is in contrast to a multinomial distribution as used in LDA. With the infinite number of possible values for $\mathbf{z}^{d}_n$, the word generating distributions in PM-LDA are expanded from only $K$ generating distributions (as in LDA) to infinitely many. Fig. \ref{fig:topics}  illustrates this using two Gaussian topic distributions, where the membership value to one topic is varied from $0$ to $1$ with an increment $0.1$. The two original topics are shown as the Gaussian distributions at either end. In LDA,  words are generated from only the two original topic distributions. In PM-LDA, words can be generated from any (of the infinitely many) convex combinations of the topic distributions.  As the scaling factor $s \rightarrow 0$, the PM-LDA model will degrade to the LDA model.

Given the hyperparameters $\boldsymbol{\Psi}=\{\boldsymbol{\alpha},\lambda,\boldsymbol{\beta}\}$, the full PM-LDA model over all words in the $d^{th}$ document is:
\begin{eqnarray}
\label{eqn:joint}
&&p(\boldsymbol{\pi}^d,s^d,\mathbf{Z}^d,\mathbf{X}^d|\boldsymbol{\alpha},\lambda,\boldsymbol{\beta})   \\ \nonumber
&=&p(\boldsymbol{\pi}^d|\boldsymbol{\alpha}) p(s^d|\lambda) \prod_{n=1}^{N^d}p(\mathbf{x}^{d}_{n}|\mathbf{z}^{d}_{n},\boldsymbol{\beta})p(\mathbf{z}^{d}_{n}|\boldsymbol{\pi}^ds^d). \nonumber 
\end{eqnarray}
where $\boldsymbol{\pi}^d$ are the topic proportions, scaling factor $s^d$, partial membership vectors $\mathbf{Z}^d=\{\mathbf{z}^{d}_{n}\}_{n=1}^{N^d}$ and a set of $N^d$  words $\mathbf{X}^d$ for document $d$. 
The log of \eqref{eqn:joint} when considering the specific forms chosen in our model, is shown in \eqref{eqn:onedocument}.
\begin{figure*}[ht!]
\begin{eqnarray}  \label{eqn:onedocument}
\pazocal{L}^{d} &=& \ln(p(\boldsymbol{\pi}^d,s^d,\mathbf{Z}^d,\mathbf{X}^d|\boldsymbol{\alpha},\lambda,\boldsymbol{\beta}) ) =  \ln \Gamma \left(\sum_{k=1}^{K}\alpha_k\right)-\sum_{k=1}^{K}\ln\Gamma \left(\alpha_k \right) + \sum_{k=1}^{K}(\alpha_k-1)\ln \pi^d_k +\ln \lambda - \lambda s^d \nonumber \\
&+&\sum_{n=1}^{N^d}\ln p(\mathbf{x}^{d}_{n}|\mathbf{z}^{d}_{n},\boldsymbol{\beta}) +\sum_{n=1}^{N^d}\bigg\{\ln \Gamma \left(\sum_{k=1}^{K}s^d\pi^{d}_{k}\right) -\sum_{k=1}^{K}\ln\Gamma(s^d\pi^{d}_{k})+\sum_{k=1}^{K}(s^d\pi^{d}_{k}-1)\ln z^{d}_{nk}\bigg\}
\end{eqnarray}
\end{figure*}
%
%
During parameter estimation, our goal is to maximize $\pazocal{L}=\sum_{d=1}^{D}\pazocal{L}^{d}$ by estimating all the model parameters ${\boldsymbol{\pi}^{d},s^{d}, \mathbf{z}^{d}_n, \boldsymbol{\beta}}$. 
In this paper, we employ an Metropolis within Gibbs  \cite{robert:2013, GlennZare:2014} sampling approach.  

\section{Parameter Estimation for PM-LDA}
The goal of parameter estimation is to maximize the following  posterior distribution,
\begin{equation}
p(\boldsymbol{\Pi},\mathbf{S},\mathbf{M},\boldsymbol{\beta}|\mathbf{D},\boldsymbol{\alpha},\lambda) \propto p({\boldsymbol{\Pi},\mathbf{S},\mathbf{M},\mathbf{D}|\boldsymbol{\alpha},\lambda,\boldsymbol{\beta})}, 
\label{eqn:post}
\end{equation}
where $\mathbf{D} =\left\{\mathbf{X}^1, \mathbf{X}^2, ..., \mathbf{X}^D \right\}$ includes all training documents and $\boldsymbol{\Pi}, \mathbf{S}, \mathbf{M}$ include all of the topic proportions, scaling factors and membership vectors, respectively.

A Metropolis within Gibbs sampler is employed to perform the MAP inference which can generate samples from the posterior distribution in \eqref{eqn:post}, \cite{robert:2013,GlennZare:2014}. An outline of the sampler is provided in Alg. \ref{alg:clustercenter}. The sampler is simple and straight-forward implement composed only of a series of draws from candidate distributions for each parameter and then evaluation of the candidate in the appropriate acceptance ratio. Our implementation of the sampler has been posted online.\footnote[1]{Code can be found at: https://github.com/TigerSense/PMLDA}   In our current implementation, we consider the topic distributions to be Gaussian with different means, $\mu_k$, but identical diagonal and isotropic covariance matrices, $\Sigma_k = \sigma^2\mathbf{I}$. 


\begin{algorithm}[ht]
\begin{algorithmic}[1]
\REQUIRE{A corpus $\mathbf{D}$, the number of topics $K$, and the number of sampling iterations $T$}
\ENSURE{Collection of all samples: $\boldsymbol{\Pi}^{(t)}, \mathbf{S}^{(t)}, \mathbf{M}^{(t)}$,  $\boldsymbol{\beta}^{(t)}=\left\{\mu^{(t)}_1, \Sigma^{(t)}_1, \mu^{(t)}_2, \Sigma^{(t)}_2..., \mu^{(t)}_K, \Sigma^{(t)}_K\right\}$.
\FOR{$t=1:T$}
\FOR{$d=1:D$}
\STATE \underline{Sample $\boldsymbol{\pi}^d$:} Draw candidate: $\boldsymbol{\pi}^\dagger \sim \text{Dir}(\boldsymbol{\alpha})$ \\
 Accept candidate with probability:\\ $a_{\boldsymbol{\pi}}=\min \left \{ 1, \frac{p(\boldsymbol{\pi}^{\dagger}, s^{(t-1)}, \mathbf{Z}^{(t-1)}, \mathbf{X}|\boldsymbol{\Psi}) p(\boldsymbol{\pi}^{(t-1)}|\boldsymbol{\alpha})}{ p(\boldsymbol{\pi}^{(t-1)}, s^{(t-1)}, \mathbf{Z}^{(t-1)}, \mathbf{X}|\boldsymbol{\Psi}) p(\boldsymbol{\pi}^\dagger|\boldsymbol{\alpha})}\right \}$

\STATE \underline{Sample $s^d$:} Draw candidate: $s^\dagger \sim \text{exp}(\lambda)$\\
Accept candidate with probability:\\
$a_s=\min \left \{ 1, \frac{p(\boldsymbol{\pi}^{(t)}, s^{\dagger}, \mathbf{Z}^{(t-1)}, \mathbf{X}|\boldsymbol{\Psi})p(s^{(t-1)}|\lambda)}{p(\boldsymbol{\pi}^{(t)}, s^{(t-1)}, \mathbf{Z}^{(t-1)}, \mathbf{X}|\boldsymbol{\Psi})p(s^{\dagger}|\lambda)}\right \}$

\FOR{$n=1:N^d$}
\STATE \underline{Sample $\mathbf{z}^d_n$:} Draw candidate: $\mathbf{z}_n^\dagger \sim \text{Dir}(\mathbf{1}_K)$\\
Accept candidate with probability:\\
$a_{\mathbf{z}}=\min \left\{1, \frac{p(\boldsymbol{\pi}^{(t)}, s^{(t)}, \mathbf{z}_n^\dagger, \mathbf{x}_n|\boldsymbol{\Psi})}{p(\boldsymbol{\pi}^{(t)}, s^{(t)}, \mathbf{z}_n^{(t-1)}, \mathbf{x}_n|\boldsymbol{\Psi})}\right\} $
\ENDFOR
\ENDFOR

\FOR{$k=1:K$}
\STATE \underline{Sample $\mu_k$:} Draw proposal: ${\mu}_k^{\dagger}\sim\mathcal{N}(\cdot|{\mu}_{\mathbf{D}},f{\Sigma}_{\mathbf{D}})$\\ 
${\mu}_{\mathbf{D}}$ and ${\Sigma}_{\mathbf{D}}$ are  mean and covariance of the data\\
Accept candidate with probability:\\
$a_k=\small{\min\left\{1, \frac{p\left(\boldsymbol{\Pi}^{(t)}, \mathbf{S}^{(t)}, \mathbf{M}^{(t)}, \mathbf{D}|{\mu}_k^{\dagger}\right)\mathcal{N}(\mu_k^{(t-1)}|\mu_\mathbf{D}, \Sigma_\mathbf{D})}{p\left(\boldsymbol{\Pi}^{(t)}, \mathbf{S}^{(t)}, \mathbf{M}^{(t)}, \mathbf{D}|\boldsymbol{\mu}_k^{(t-1)}\right)\mathcal{N}(\mu_k^{\dagger}|\mu_\mathbf{D}, \Sigma_\mathbf{D})} \right\}}$

\ENDFOR
\STATE \underline{Sample covariance matrices $\boldsymbol{\Sigma}= \sigma^2\mathbf{I}$:}\\
Draw candidate from:\\ $\sigma^2 = \frac{1}{2}\left\{\max_{\mathbf{x}_n}d^2(\mathbf{x}_n-{\mu}_{\mathbf{D}})-\min_{\mathbf{x}_n}d^2(\mathbf{x}_n-{\mu}_{\mathbf{D}}) \right\}$\\
Accept candidate with probability: \\
$a_{\boldsymbol{\Sigma}}= \min\left\{1, \frac{p\left(\boldsymbol{\Pi}^{(t)}, \mathbf{S}^{(t)}, \mathbf{M}^{(t)}, \mathbf{D}|\boldsymbol{\Sigma}^\dagger\right)}{p\left(\boldsymbol{\Pi}^{(t)}, \mathbf{S}^{(t)}, \mathbf{M}^{(t)}, \mathbf{D}|\boldsymbol{\Sigma}^{(t-1)}\right)} \right\}.$

\ENDFOR}
\end{algorithmic}
\caption{Metropolis-within-Gibbs Sampling Method for Parameter Estimation}
\label{alg:clustercenter}
\end{algorithm}

The proposed Metropolis within Gibbs scheme will return the full distribution of parameter values given the desired posterior. We use the MAP sample (i.e., the sample with the largest log posterior value) as the final estimate, $\left\{\boldsymbol{\Pi}^{*}, \mathbf{S}^{*},\mathbf{M}^{*},\boldsymbol{\mu}^{*}, \boldsymbol{\Sigma}^{*}\right\}$.


\section{Data \& Experimental Results}
In this section, we show  results of image segmentation on three datasets: (i) Synthetic Aperture SONAR (SAS) imagery, (ii) Sunset imagery, and the (iii) MSRC dataset \cite{v1:2004}. 

\begin{figure*}
\centering
 \begin{subfigure}[t]{0.11\textwidth}
     \centering
     \includegraphics[width=1\linewidth,,height=0.8\linewidth]{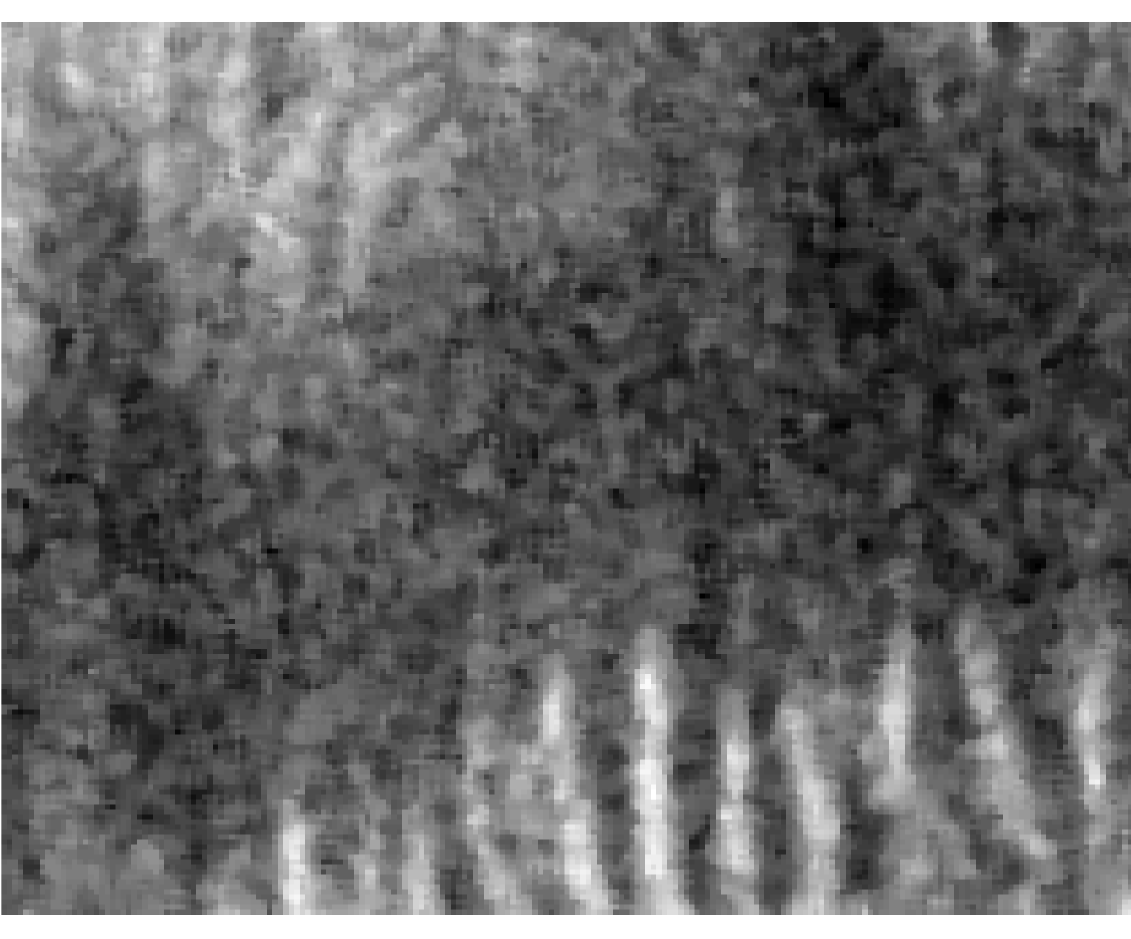}
     \captionsetup{labelformat=empty,width=1\linewidth,justification=centering,skip=0pt} 
    \caption{{(a) Image 1}} \label{fig:c1}
 \end{subfigure}
 \begin{subfigure}[t]{0.11\textwidth}
     \centering
     \includegraphics[width=1\linewidth,,height=0.8\linewidth]{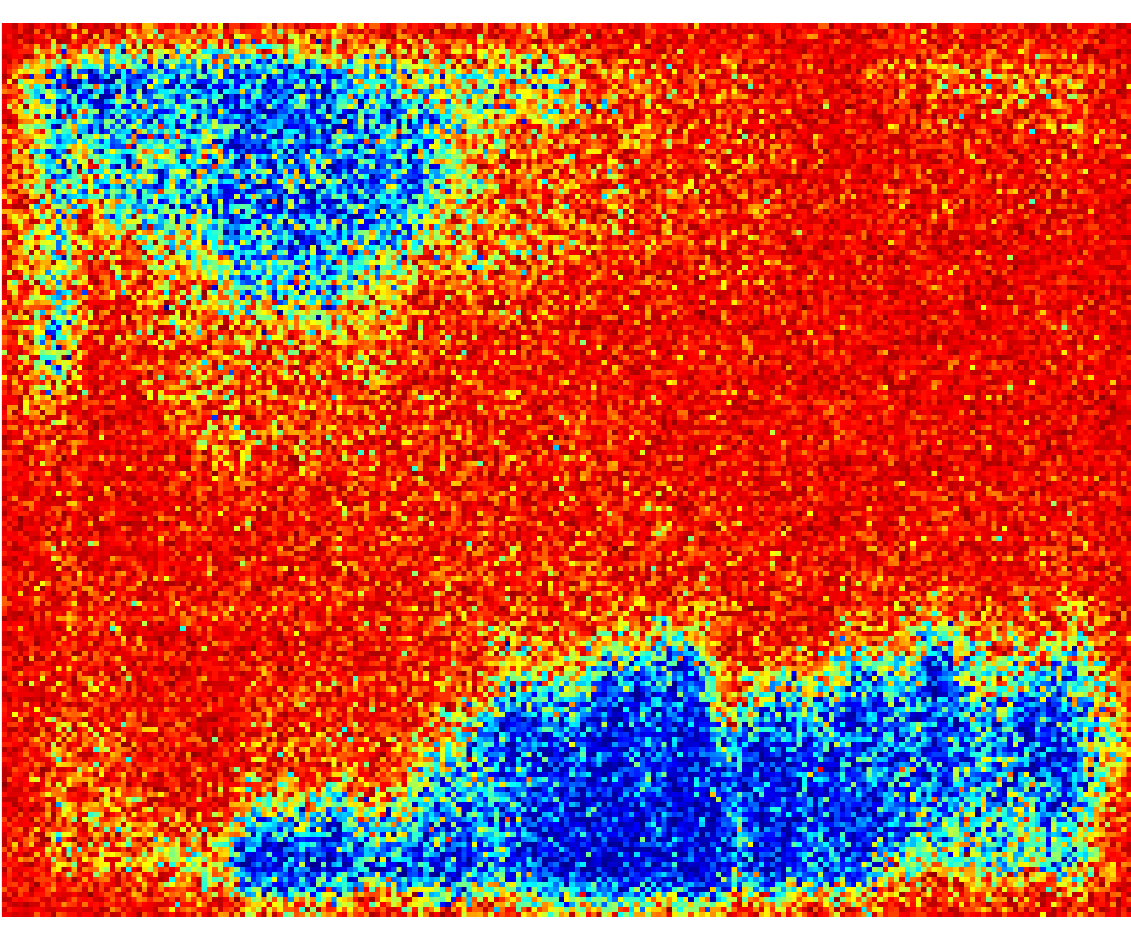}
     \captionsetup{labelformat=empty,width=1\linewidth,justification=centering,skip=0pt} 
    \caption{{(b) PM-LDA:1}}
    \end{subfigure}
     \begin{subfigure}[t]{0.11\textwidth}
     \centering
     \includegraphics[width=1\linewidth,,height=0.8\linewidth]{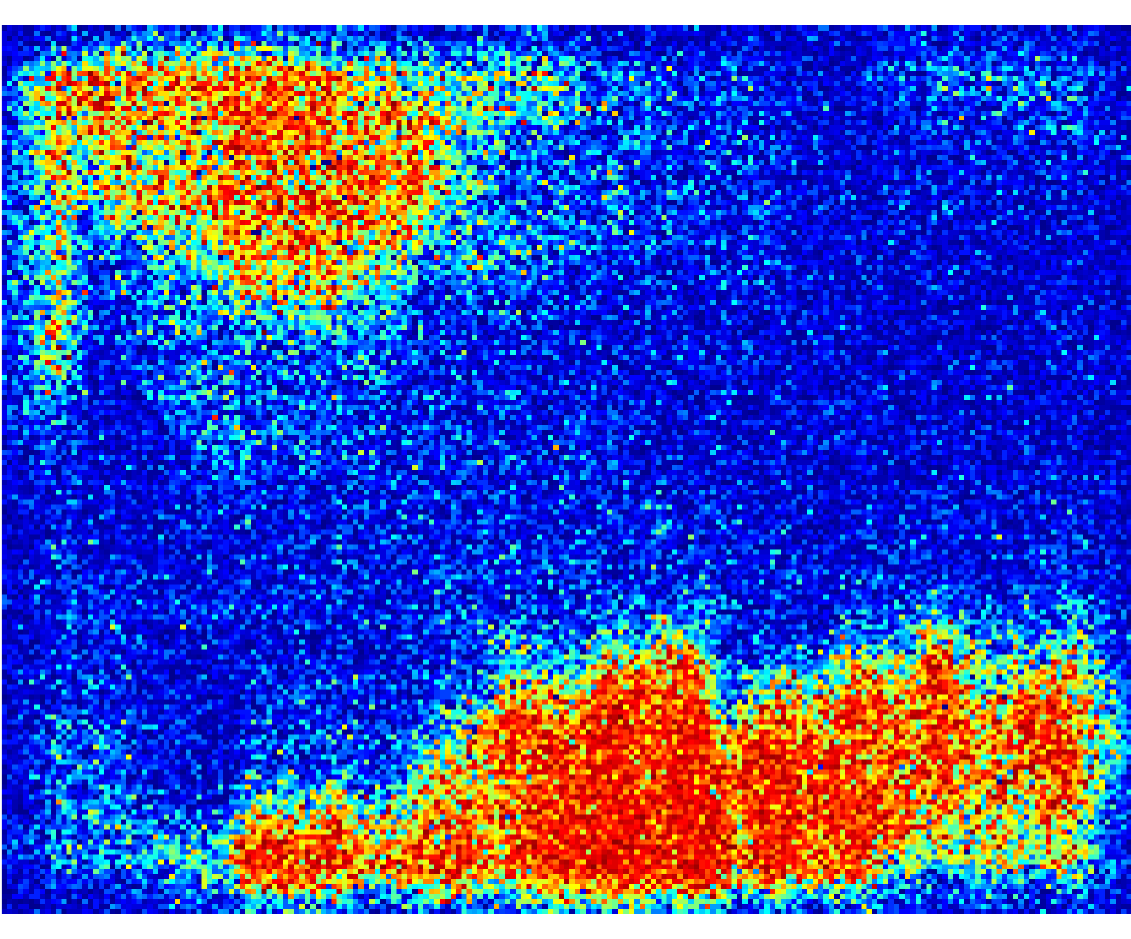}
    \captionsetup{labelformat=empty,width=1\linewidth,justification=centering,skip=0pt} 
    \caption{{(c) PM-LDA:2}} \label{fig:1c}
    \end{subfigure}
     \begin{subfigure}[t]{0.11\textwidth}
     \centering
     \includegraphics[width=1\linewidth,,height=0.8\linewidth]{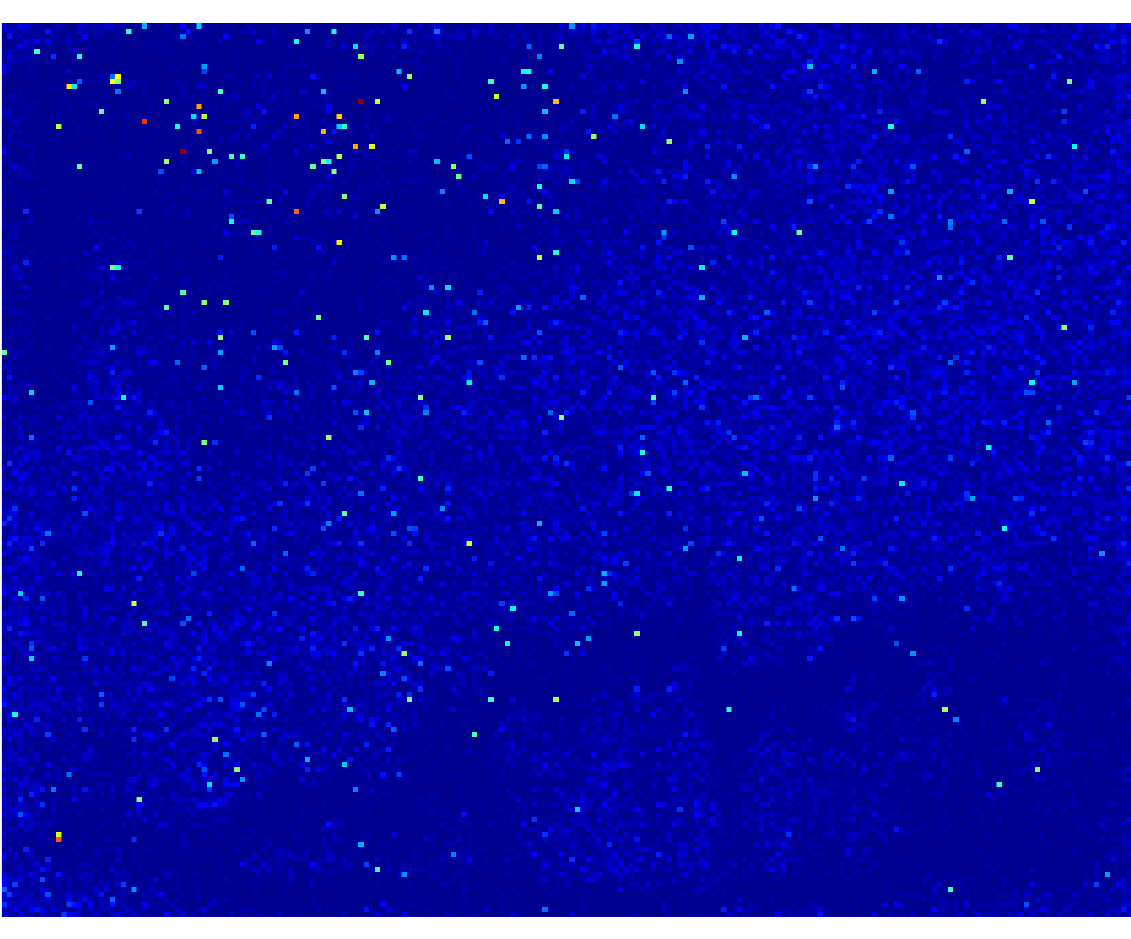}
     \captionsetup{labelformat=empty,width=1.1\linewidth,justification=centering,skip=-5pt} 
    \caption{{(d) PM-LDA:3}}
 \end{subfigure}  
\begin{subfigure}[t]{0.015\textwidth}
   \centering
     \includegraphics[width=0.91\linewidth]{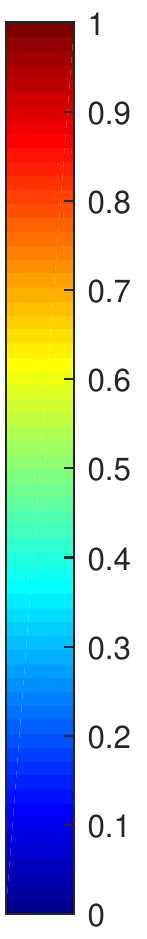}
     \captionsetup{labelformat=empty,width=1\linewidth,justification=centering,skip=-5pt} 
     \caption{}
\end{subfigure}
\setcounter{subfigure}{4}
 \begin{subfigure}[t]{0.11\textwidth}
     \centering
     \includegraphics[width=1\linewidth,,height=0.8\linewidth]{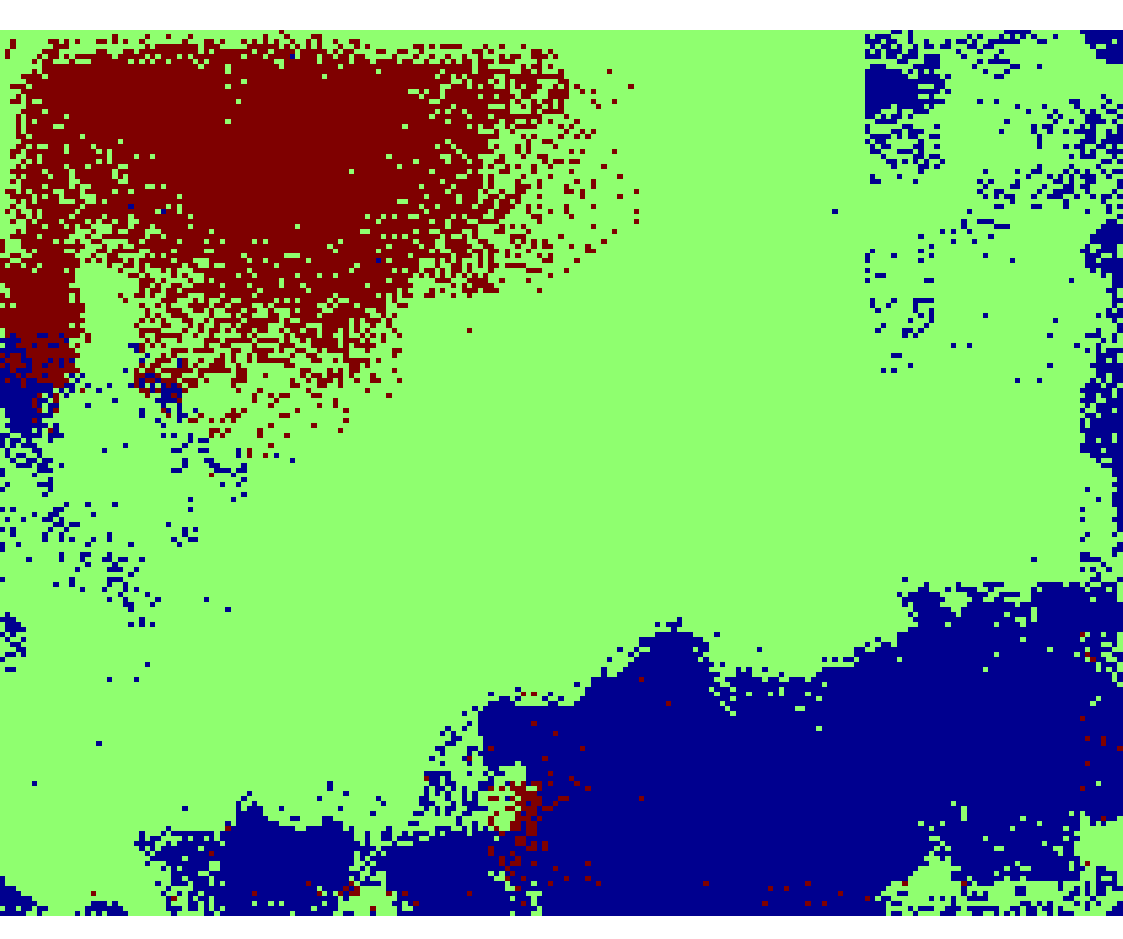}
     \captionsetup{labelformat=empty,width=1\linewidth,justification=centering,skip=0pt} 
    \caption{{(e) LDA}} \label{fig:1e}
    \end{subfigure}   
 \begin{subfigure}[t]{0.11\textwidth}
     \centering
     \includegraphics[width=1\linewidth,,height=0.8\linewidth]{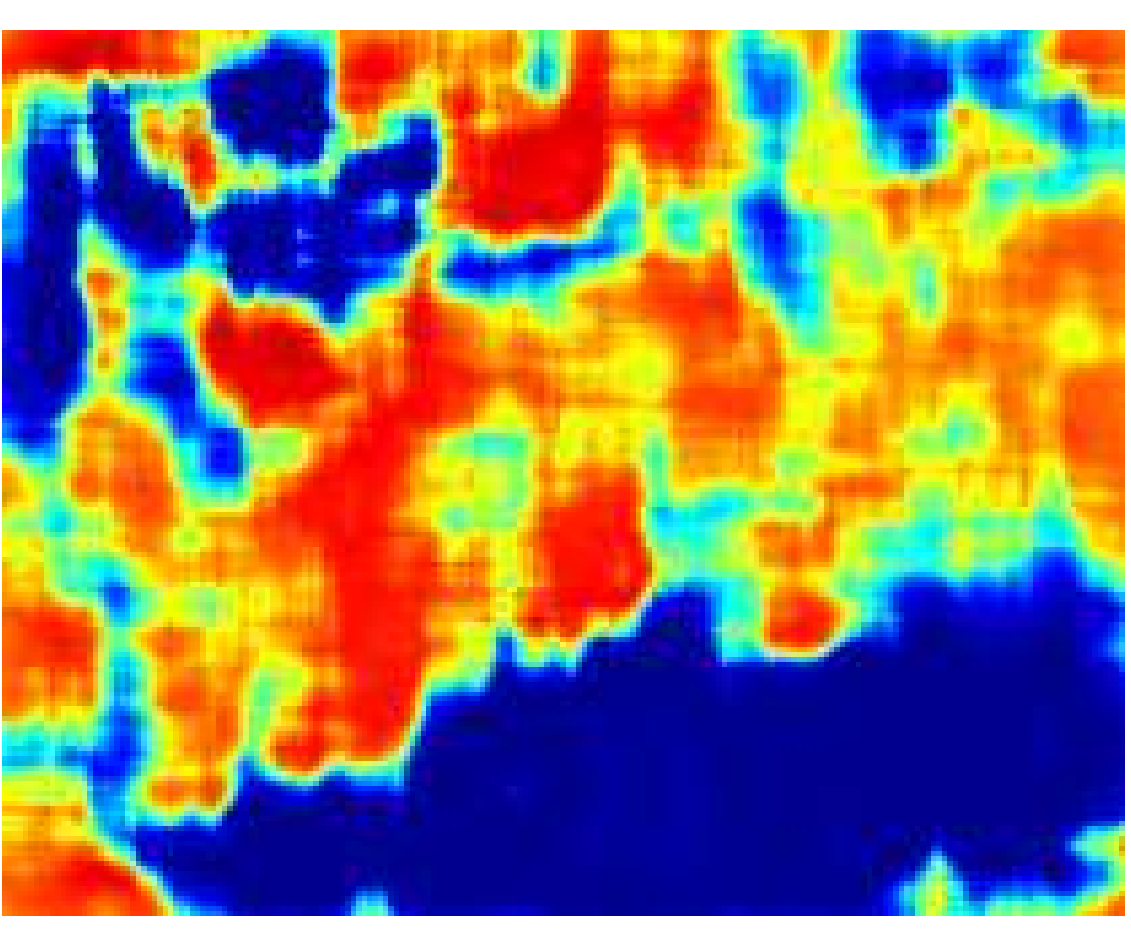}
    \captionsetup{labelformat=empty,width=1\linewidth,justification=centering,skip=0pt} 
      \caption{{(f) FCM:1}}
 \end{subfigure}
 \begin{subfigure}[t]{0.11\textwidth}
     \centering
     \includegraphics[width=1\linewidth,,height=0.8\linewidth]{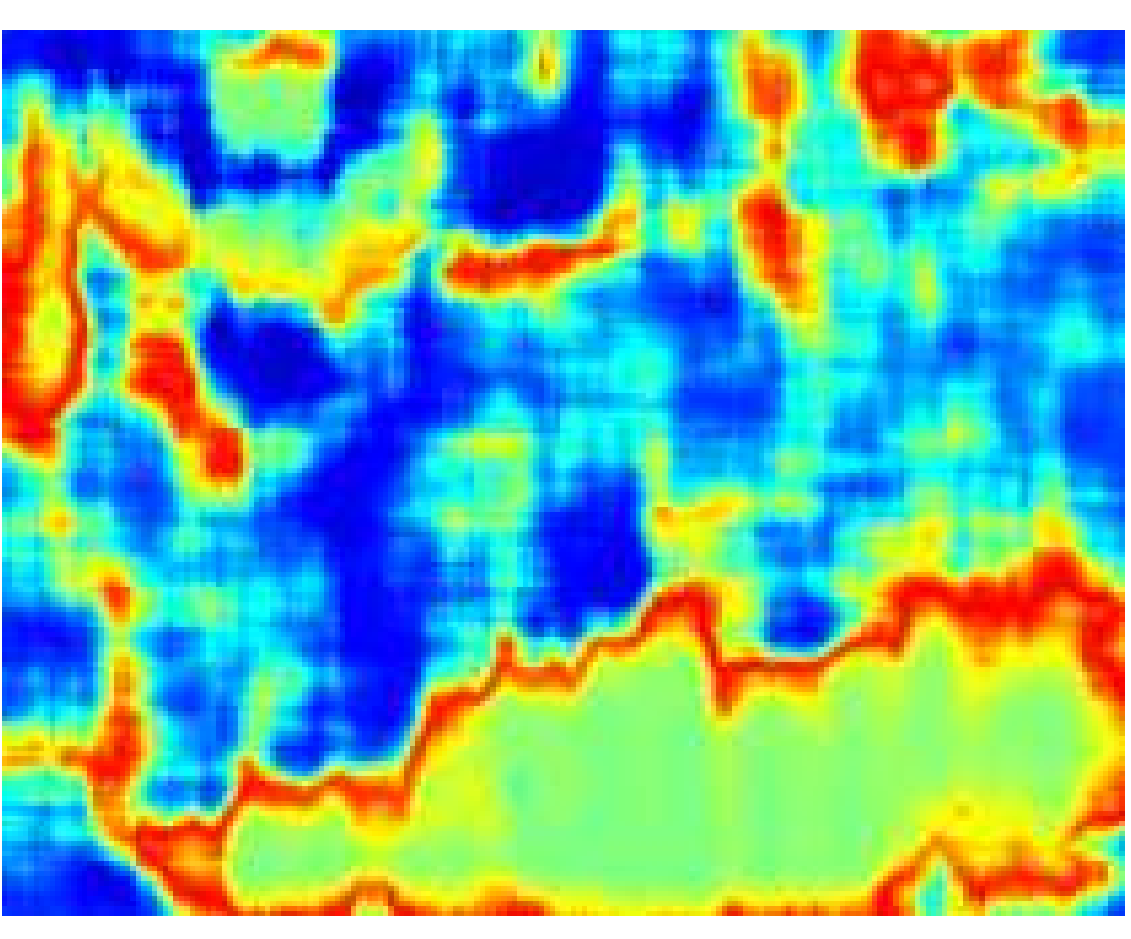}
     \captionsetup{labelformat=empty,width=1\linewidth,justification=centering,skip=0pt} 
        \caption{{(g) FCM:2}}
    \end{subfigure}
     \begin{subfigure}[t]{0.11\textwidth}
     \centering
     \includegraphics[width=1\linewidth,height=0.8\linewidth]{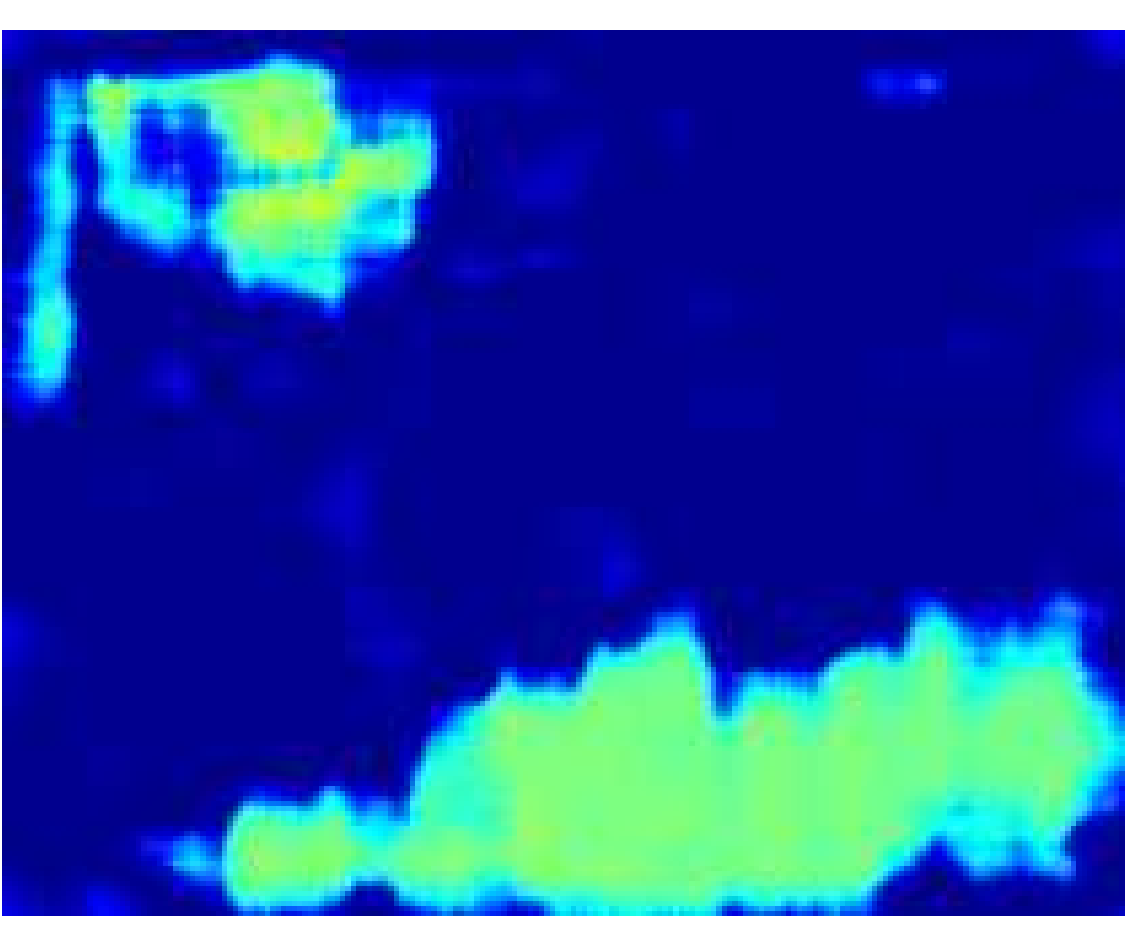}
     \captionsetup{labelformat=empty,width=1\linewidth,justification=centering,skip=0pt} 
        \caption{{(h) FCM:3}}
    \end{subfigure}
    \begin{subfigure}[t]{0.015\textwidth}
   \centering
     \includegraphics[width=0.91\linewidth]{jetcolorbar.pdf}
     \captionsetup{labelformat=empty,width=1\linewidth,justification=centering,skip=0pt} 
\end{subfigure}
\setcounter{subfigure}{8}
 \begin{subfigure}[t]{0.11\textwidth}
     \centering
     \includegraphics[width=1\linewidth,height=0.8\linewidth]{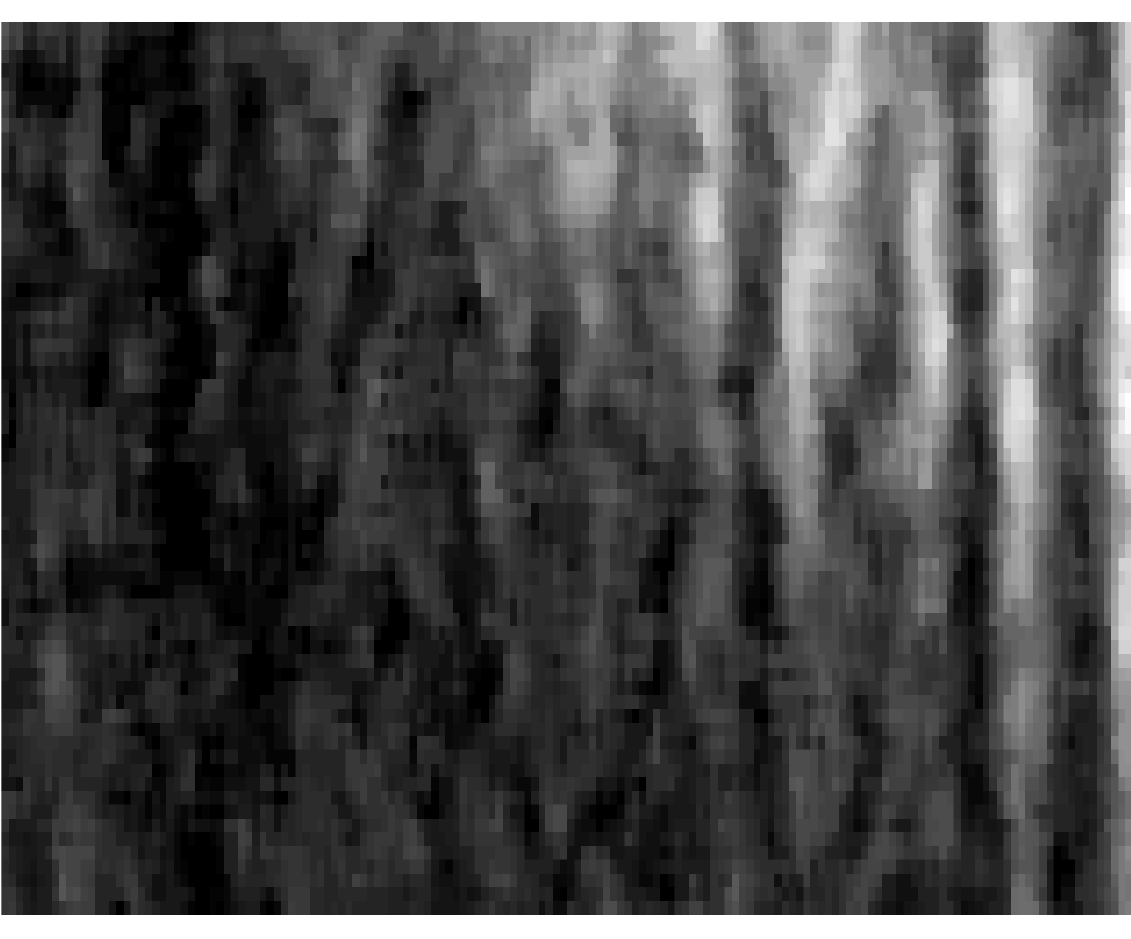}
			 \captionsetup{labelformat=empty,skip=0pt}
		\caption{(i) Image 2} \label{fig:c2}
 \end{subfigure}
 \begin{subfigure}[t]{0.11\textwidth}
     \centering
     \includegraphics[width=1\linewidth,,height=0.8\linewidth]{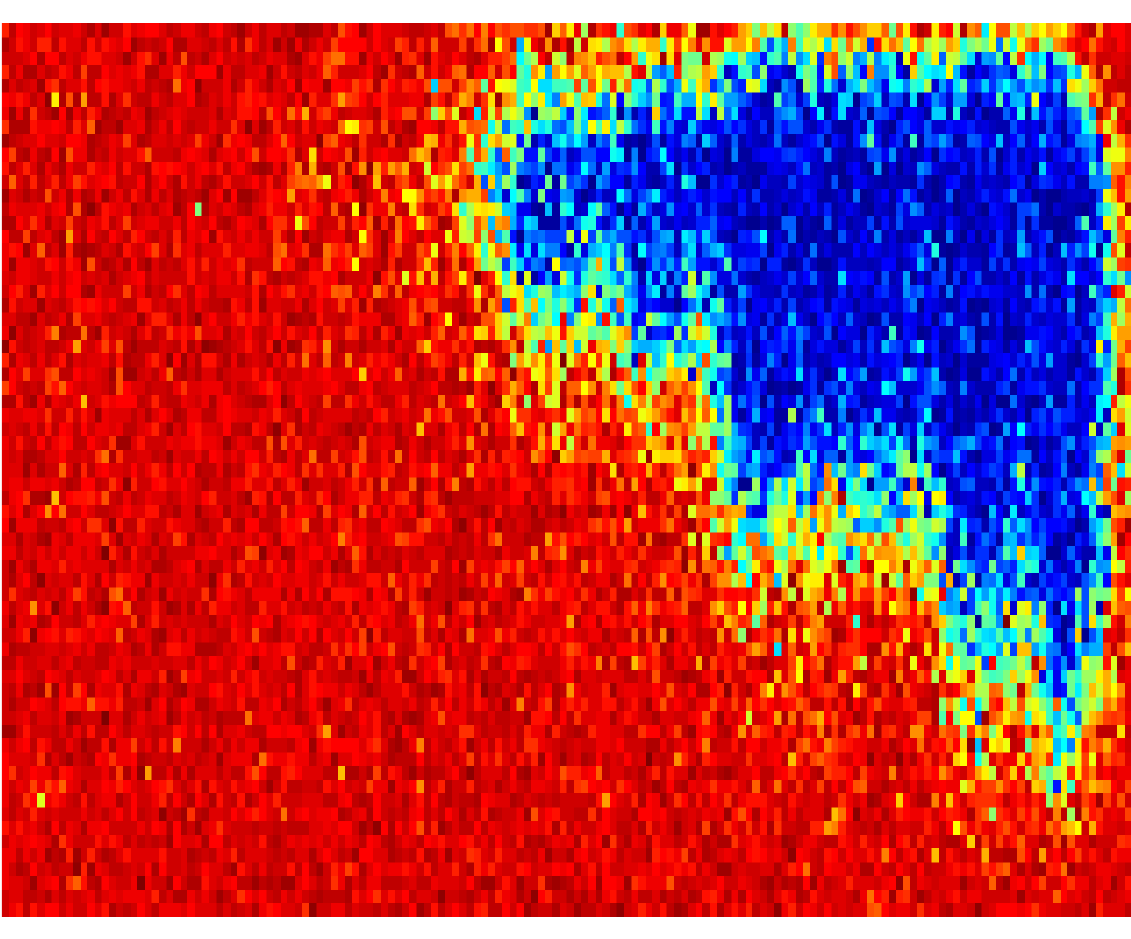} 
				 \captionsetup{labelformat=empty,skip=0pt}
		\caption{(j) PM-LDA:1}
		\end{subfigure}
		 \begin{subfigure}[t]{0.11\textwidth}
     \centering
     \includegraphics[width=1\linewidth,,height=0.8\linewidth]{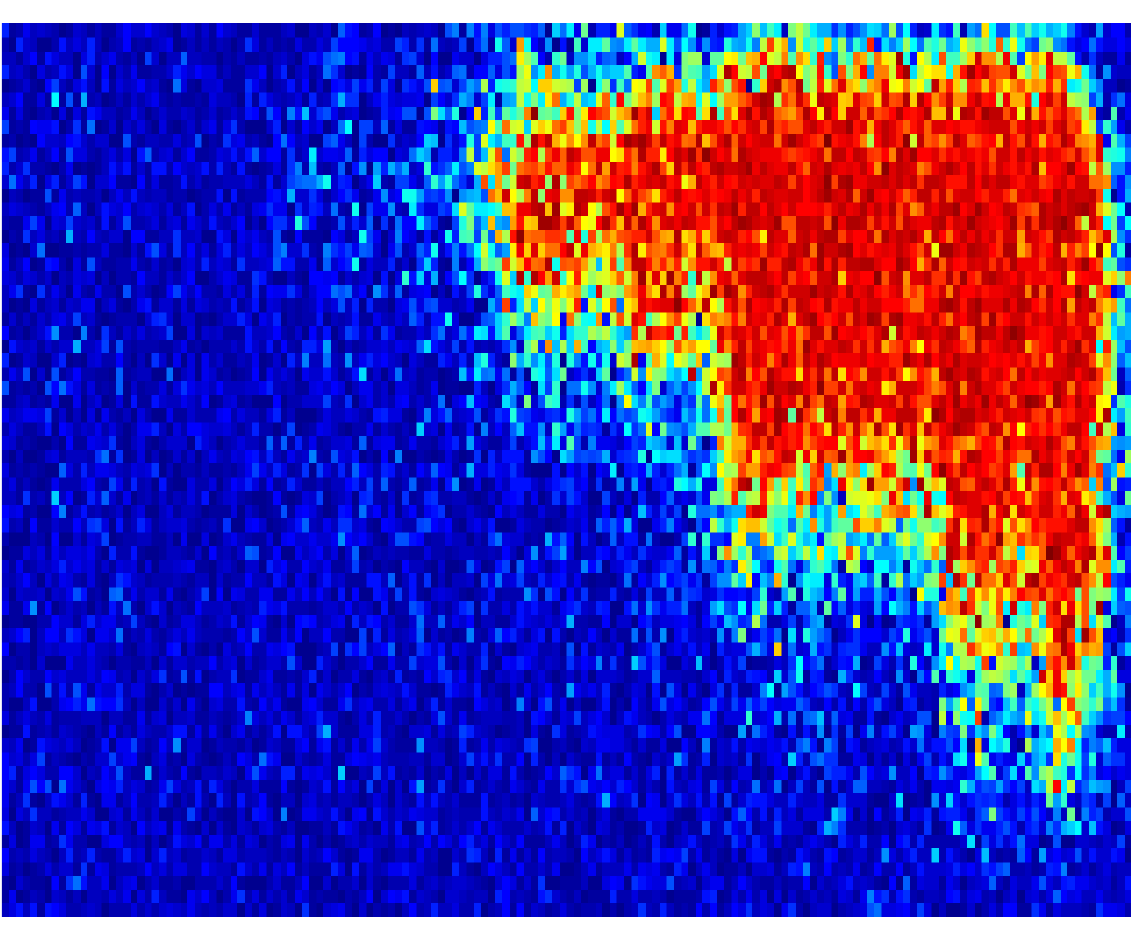}
				 \captionsetup{labelformat=empty,skip=0pt}
		\caption{(k) PM-LDA:2} \label{fig:2c}
		\end{subfigure}
		 \begin{subfigure}[t]{0.11\textwidth}
     \centering
     \includegraphics[width=1\linewidth,,height=0.8\linewidth]{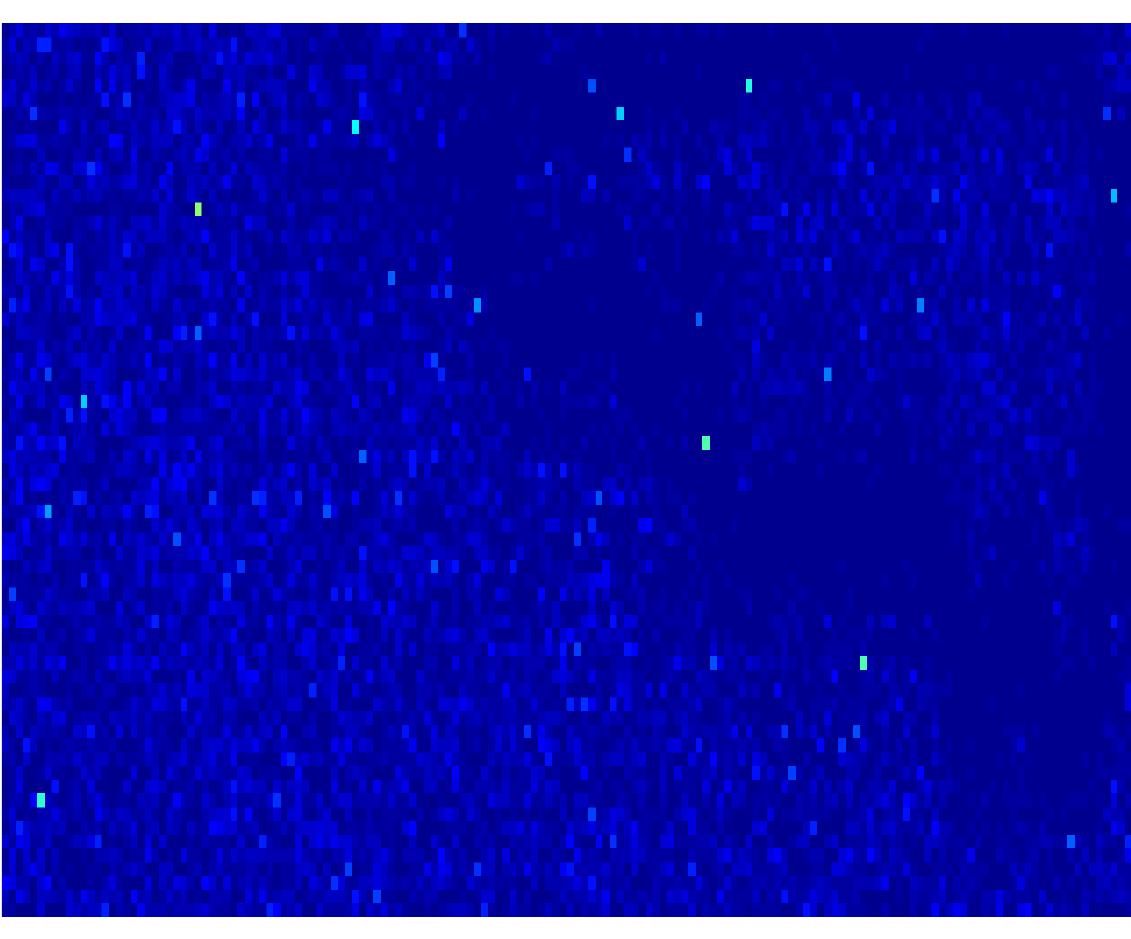}
				 \captionsetup{labelformat=empty,skip=0pt}
		\caption{(l) PM-LDA:3}
 \end{subfigure}
\begin{subfigure}[t]{0.015\textwidth}
   \centering
     \includegraphics[width=0.91\linewidth]{jetcolorbar.pdf}
				 \captionsetup{labelformat=empty,skip=0pt}
				 \caption{}
\end{subfigure}
\setcounter{subfigure}{12}
\begin{subfigure}[t]{0.11\textwidth}
     \centering
     \includegraphics[width=1\linewidth,,height=0.8\linewidth]{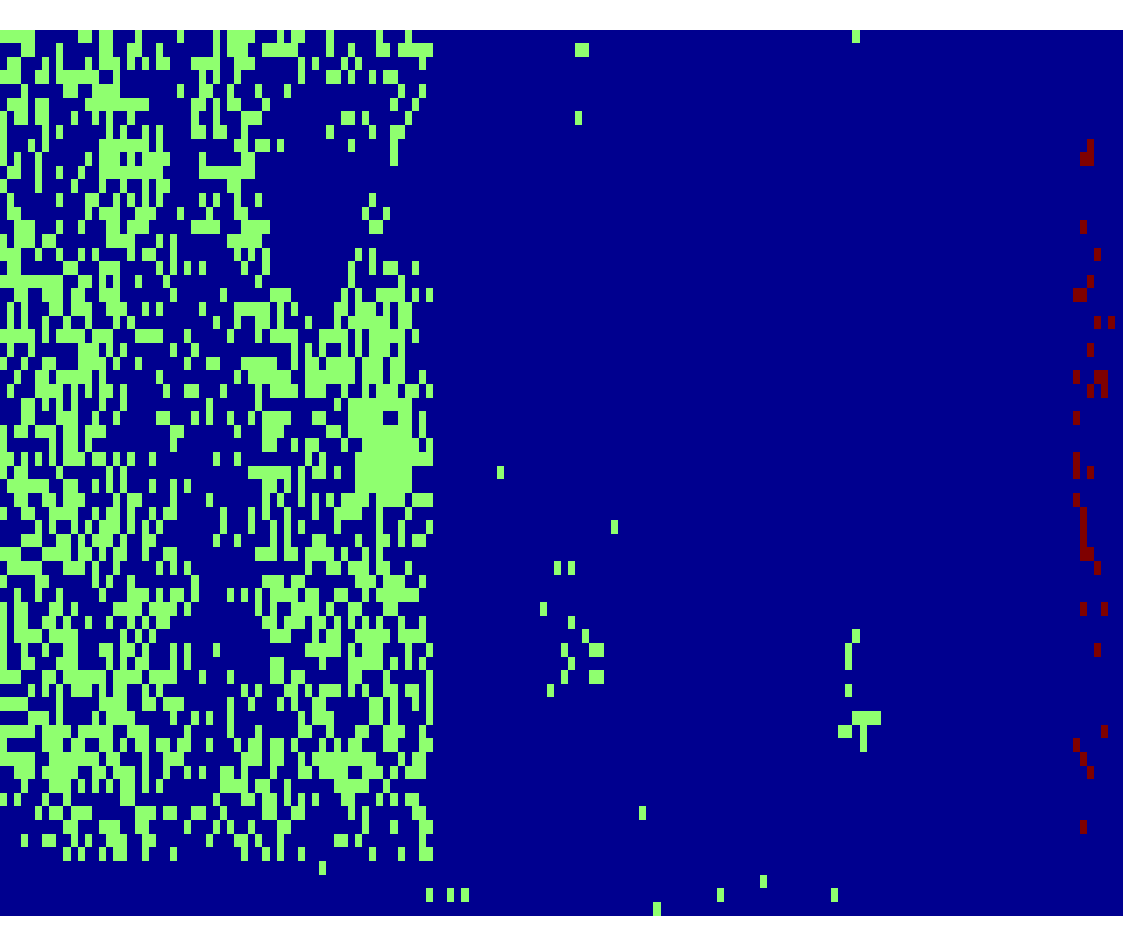}
					 \captionsetup{labelformat=empty,skip=0pt}
		\caption{(m) LDA} \label{fig:2e}
		\end{subfigure} 
 \begin{subfigure}[t]{0.11\textwidth}
     \centering
     \includegraphics[width=1\linewidth,,height=0.8\linewidth]{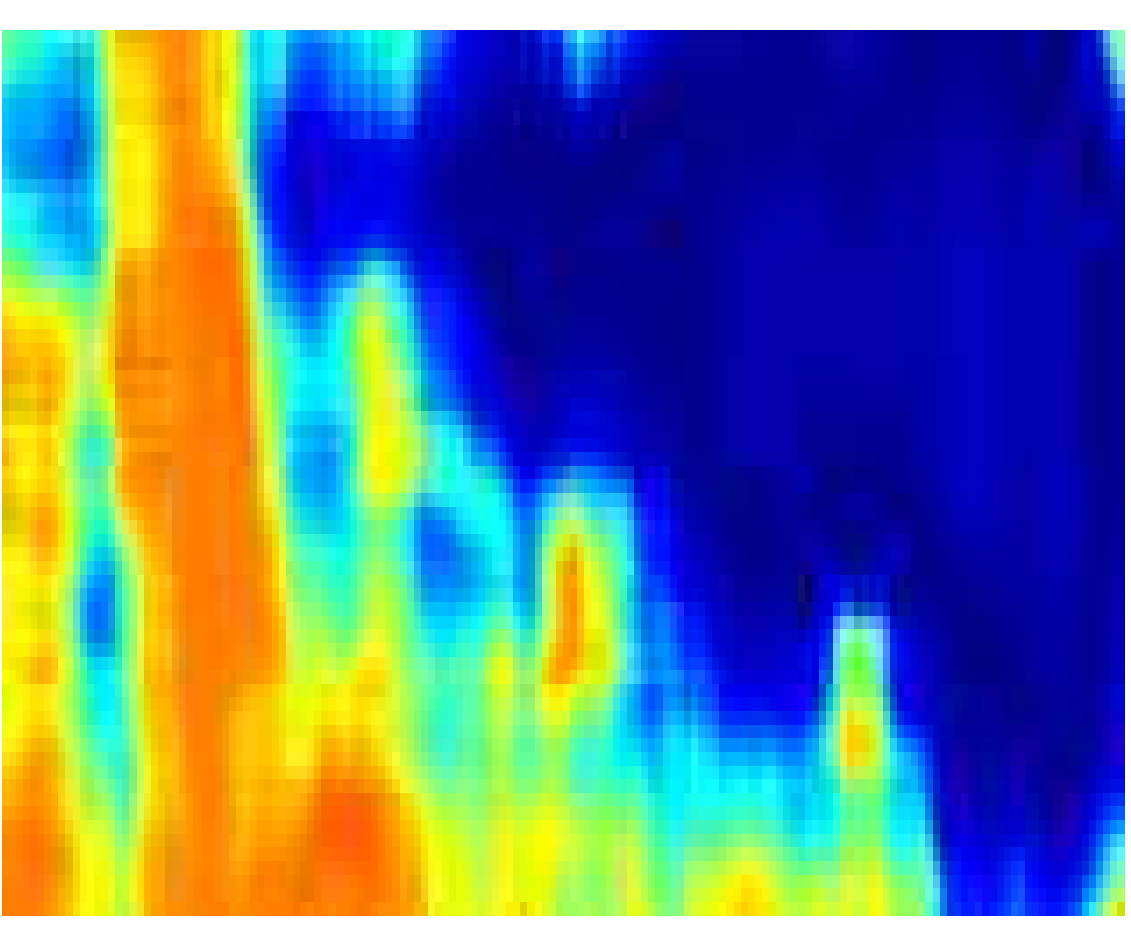}
		\captionsetup{labelformat=empty,skip=0pt}
		\caption{(n) FCM:1} 
 \end{subfigure}
 \begin{subfigure}[t]{0.11\textwidth}
     \centering
     \includegraphics[width=1\linewidth,,height=0.8\linewidth]{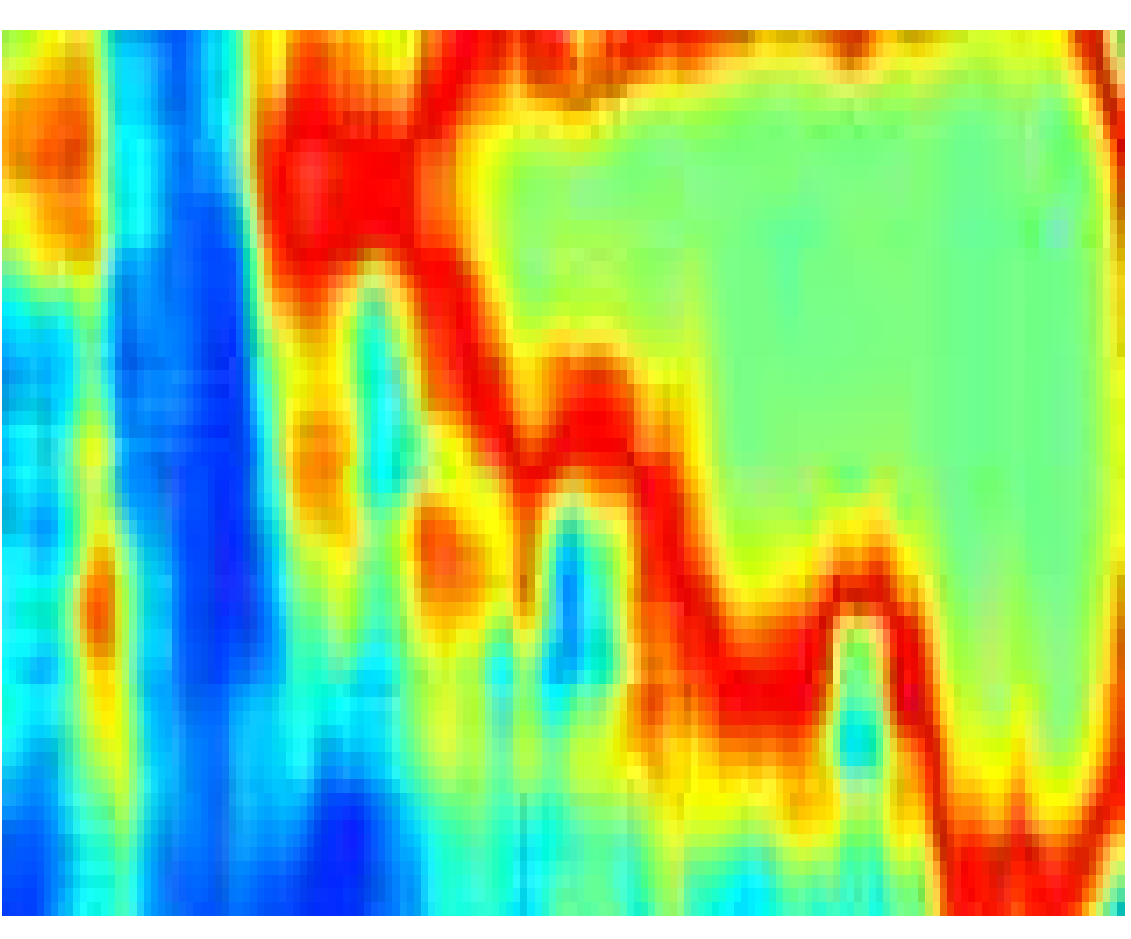}
		\captionsetup{labelformat=empty,skip=0pt}
		\caption{(o) FCM:2} \label{fig:2g}
		\end{subfigure}
		 \begin{subfigure}[t]{0.11\textwidth}
     \centering
     \includegraphics[width=1\linewidth,,height=0.8\linewidth]{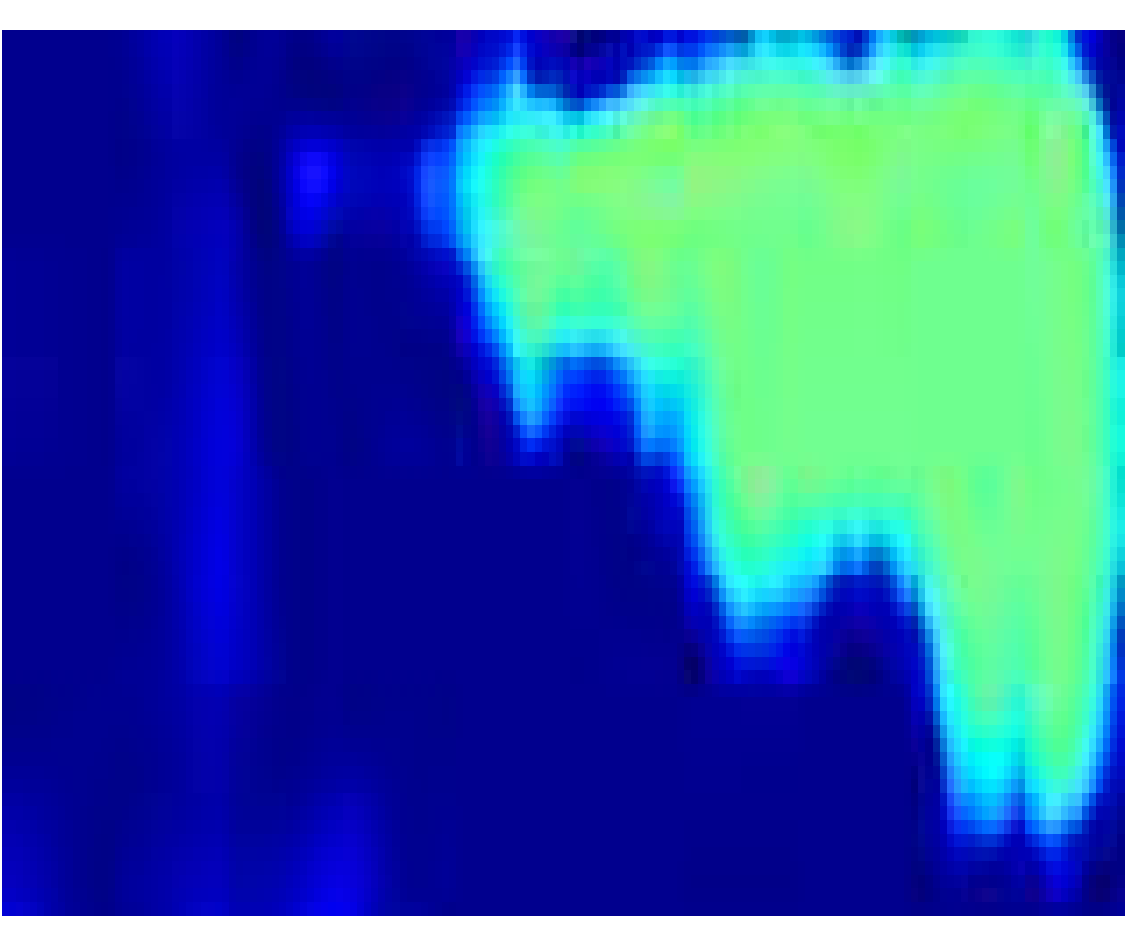}
		 \captionsetup{labelformat=empty,skip=0pt}
		\caption{(p) FCM:3} \label{fig:2h}
 \end{subfigure}	
\begin{subfigure}[t]{0.015\textwidth}
   \centering
     \includegraphics[width=0.91\linewidth]{jetcolorbar.pdf}
		 \captionsetup{labelformat=empty,skip=0pt}
\end{subfigure}
\setcounter{subfigure}{16}
\begin{subfigure}[t]{0.11\textwidth}
	\centering
	\includegraphics[width=1\linewidth,height=0.8\linewidth]{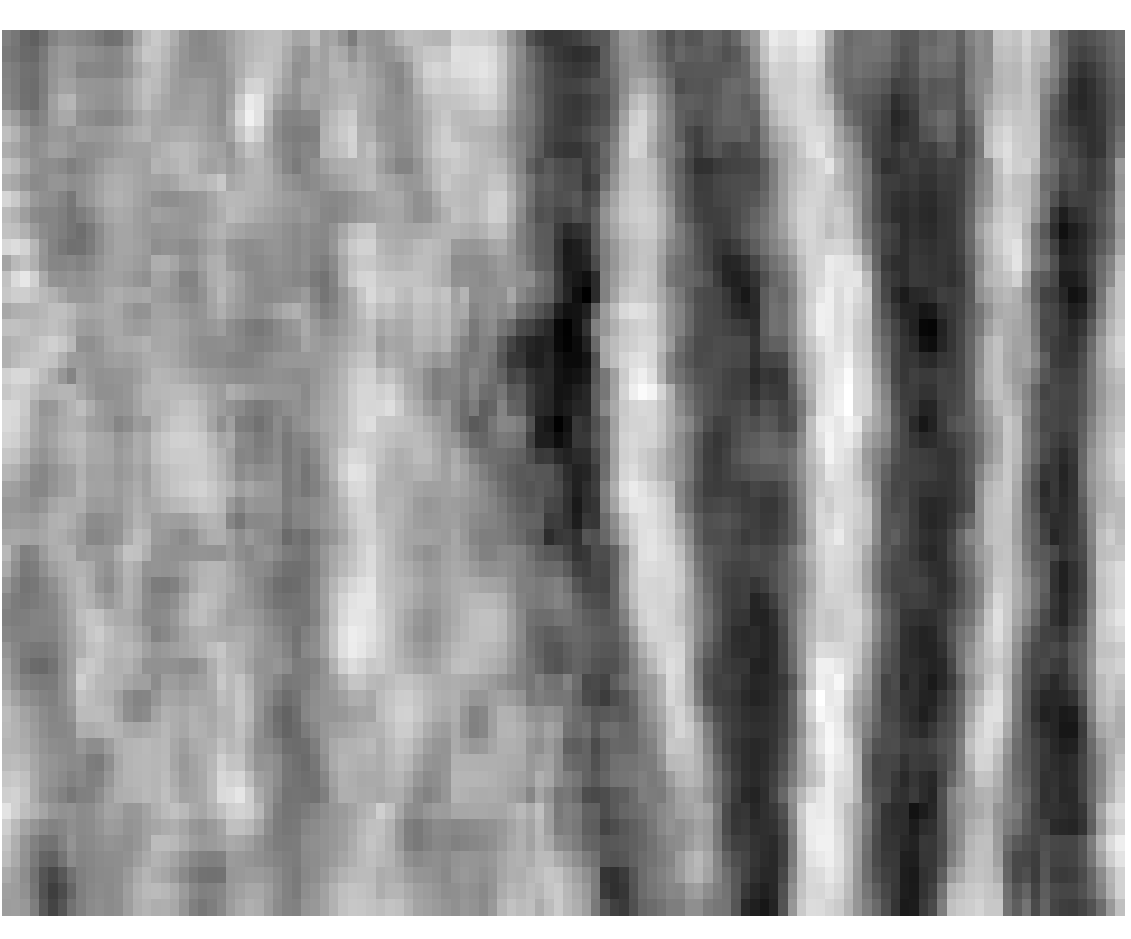}
	\captionsetup{labelformat=empty,skip=0pt}
	\caption{(q) Image 3} \label{fig:c3}
\end{subfigure}
\begin{subfigure}[t]{0.11\textwidth}
	\centering
	\includegraphics[width=1\linewidth,height=0.8\linewidth]{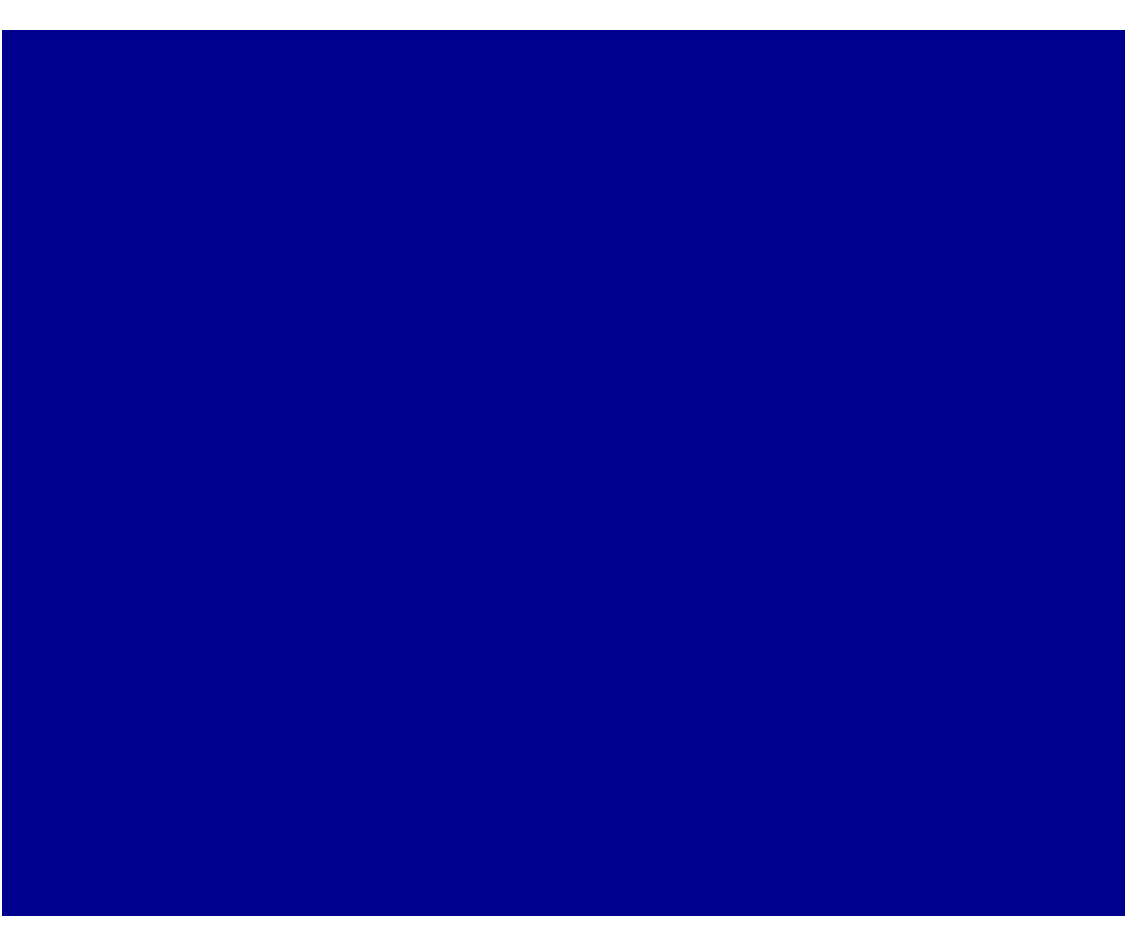}
	\captionsetup{labelformat=empty,skip=0pt}
	\caption{(r) PM-LDA:1}
\end{subfigure}
\begin{subfigure}[t]{0.11\textwidth}
	\centering
	\includegraphics[width=1\linewidth,height=0.8\linewidth]{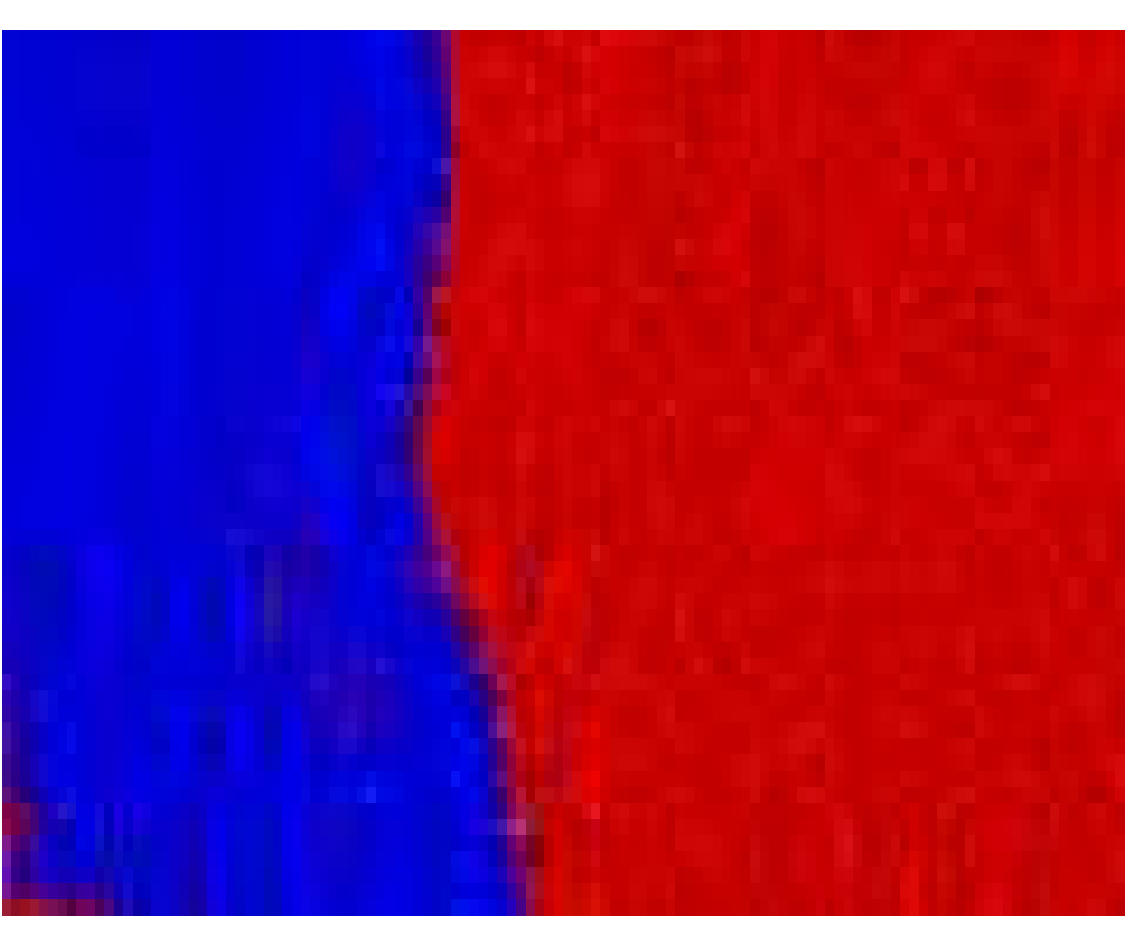}
	\captionsetup{labelformat=empty,skip=0pt}
	\caption{(s) PM-LDA:2}
\end{subfigure}
\begin{subfigure}[t]{0.11\textwidth}
	\includegraphics[width=1\linewidth,height=0.8\linewidth]{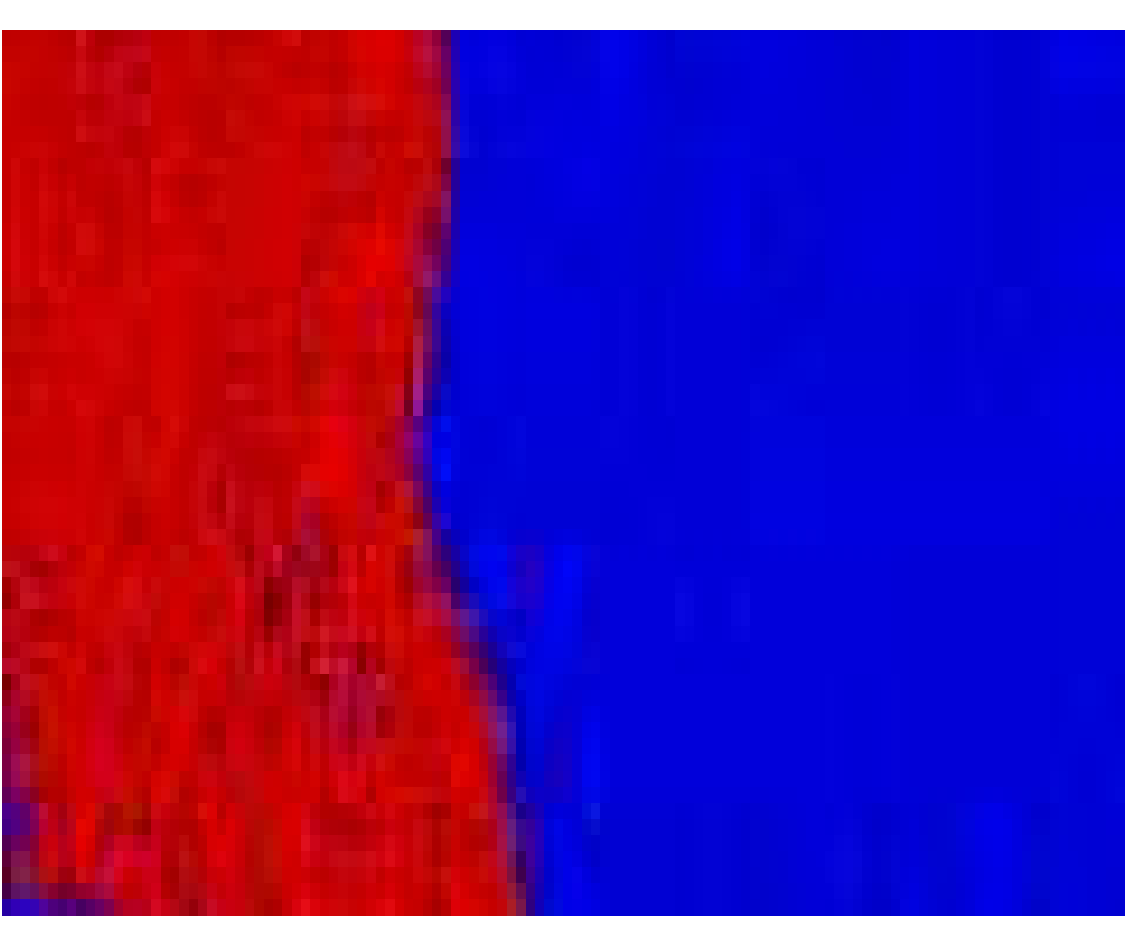}
	\centering
	\captionsetup{labelformat=empty,skip=0pt}
	\caption{(t) PM-LDA:3}\label{fig:3d}
\end{subfigure}
\begin{subfigure}[t]{0.015\textwidth}
	\centering
	\includegraphics[width=0.91\linewidth]{jetcolorbar.pdf}
	\captionsetup{labelformat=empty,skip=0pt}
\end{subfigure}
\setcounter{subfigure}{20}
\begin{subfigure}[t]{0.11\textwidth}
	\centering
	\includegraphics[width=1\linewidth,height=0.8\linewidth]{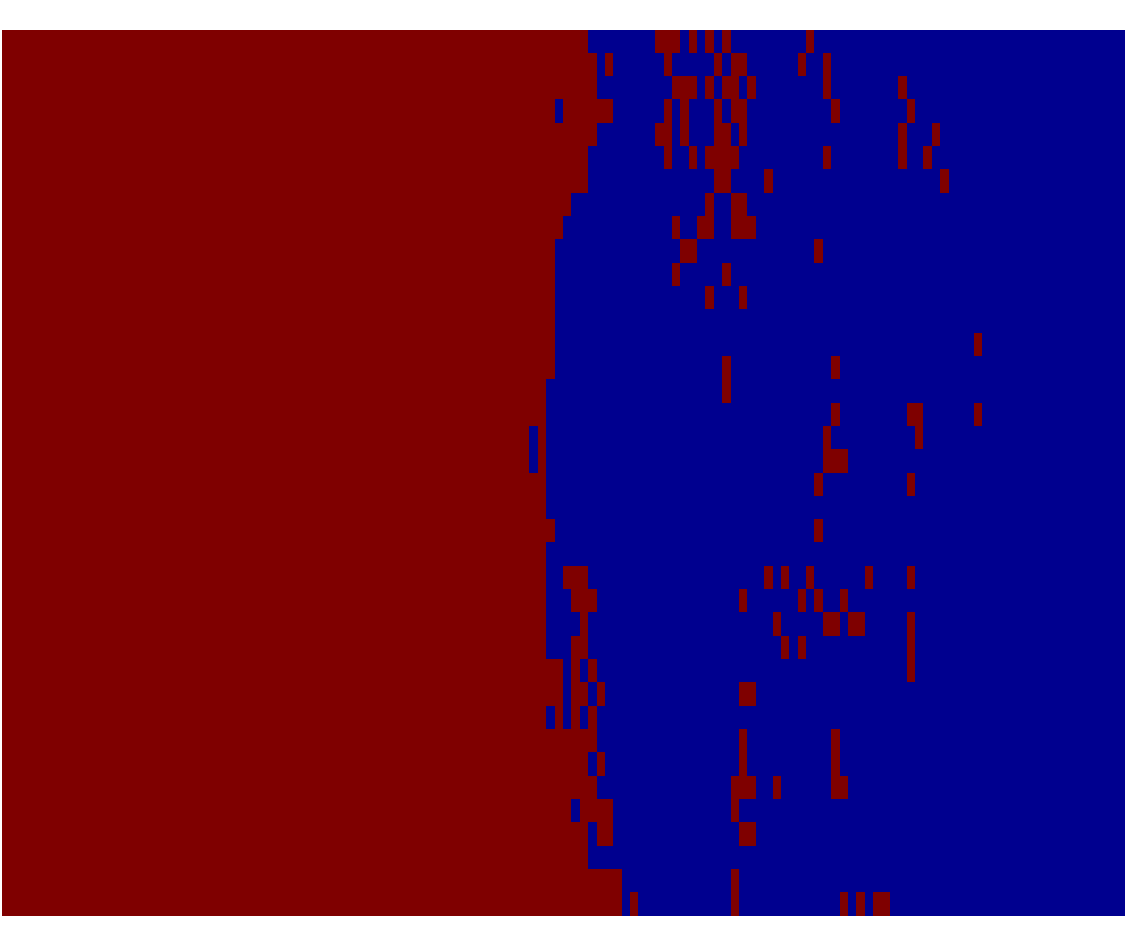}
	\captionsetup{labelformat=empty,skip=0pt}
	\caption{(u) LDA} \label{fig:3e}
\end{subfigure} 
\begin{subfigure}[t]{0.11\textwidth}
	\centering
	\includegraphics[width=1\linewidth,height=0.8\linewidth]{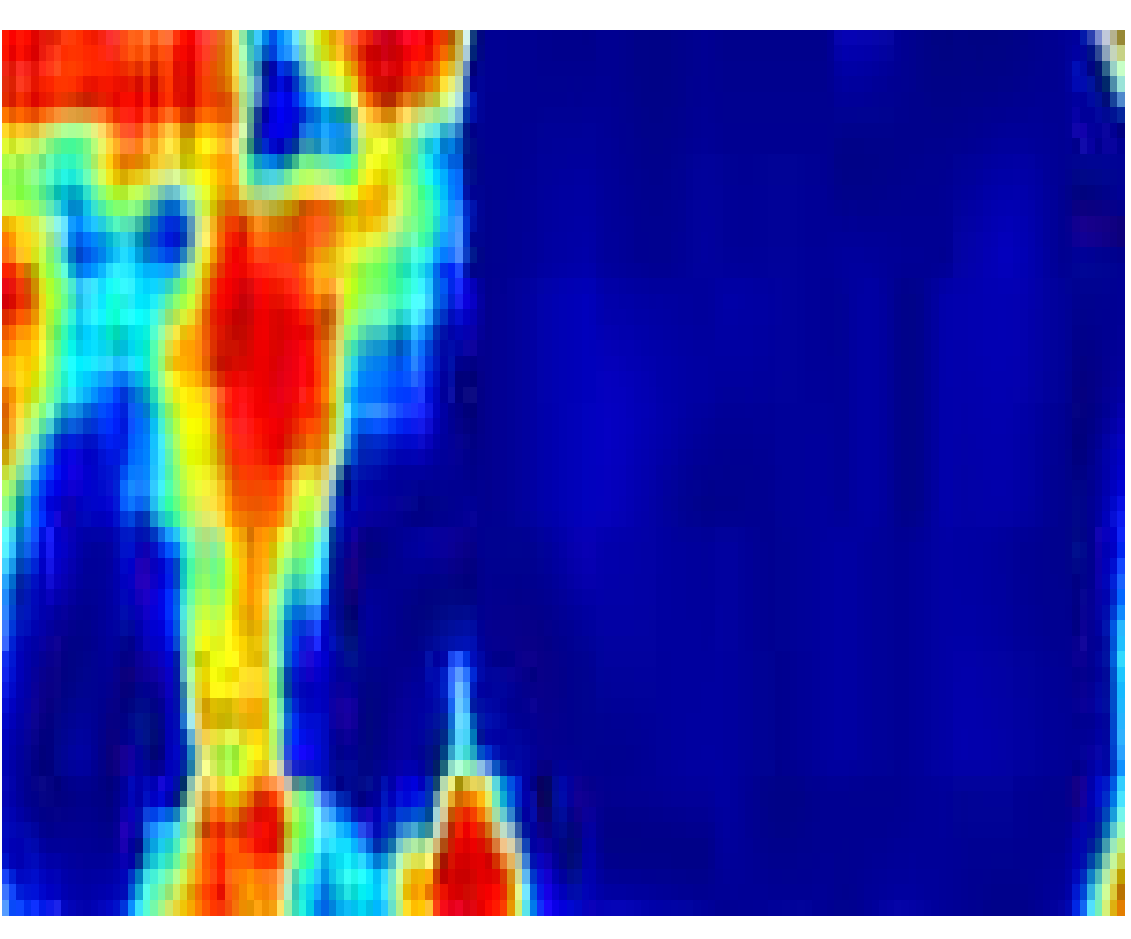}
	\captionsetup{labelformat=empty,skip=0pt}
	\caption{(v) FCM:1}
\end{subfigure}
\begin{subfigure}[t]{0.11\textwidth}
	\centering
	\includegraphics[width=1\linewidth,height=0.8\linewidth]{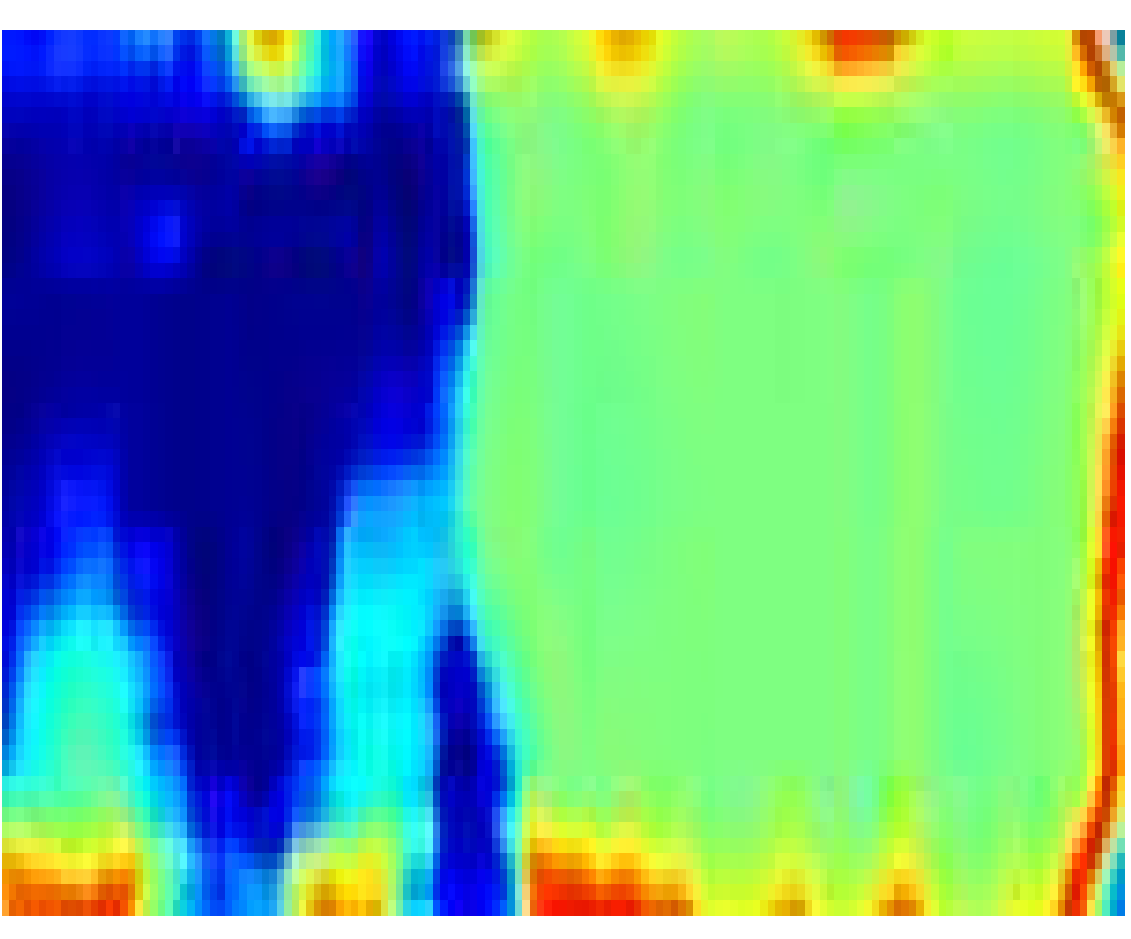}
	\captionsetup{labelformat=empty,skip=0pt}
	\caption{(w) FCM:2} \label{fig:3g}
\end{subfigure}
\begin{subfigure}[t]{0.11\textwidth}
	\centering
	\includegraphics[width=1\linewidth,height=0.81\linewidth]{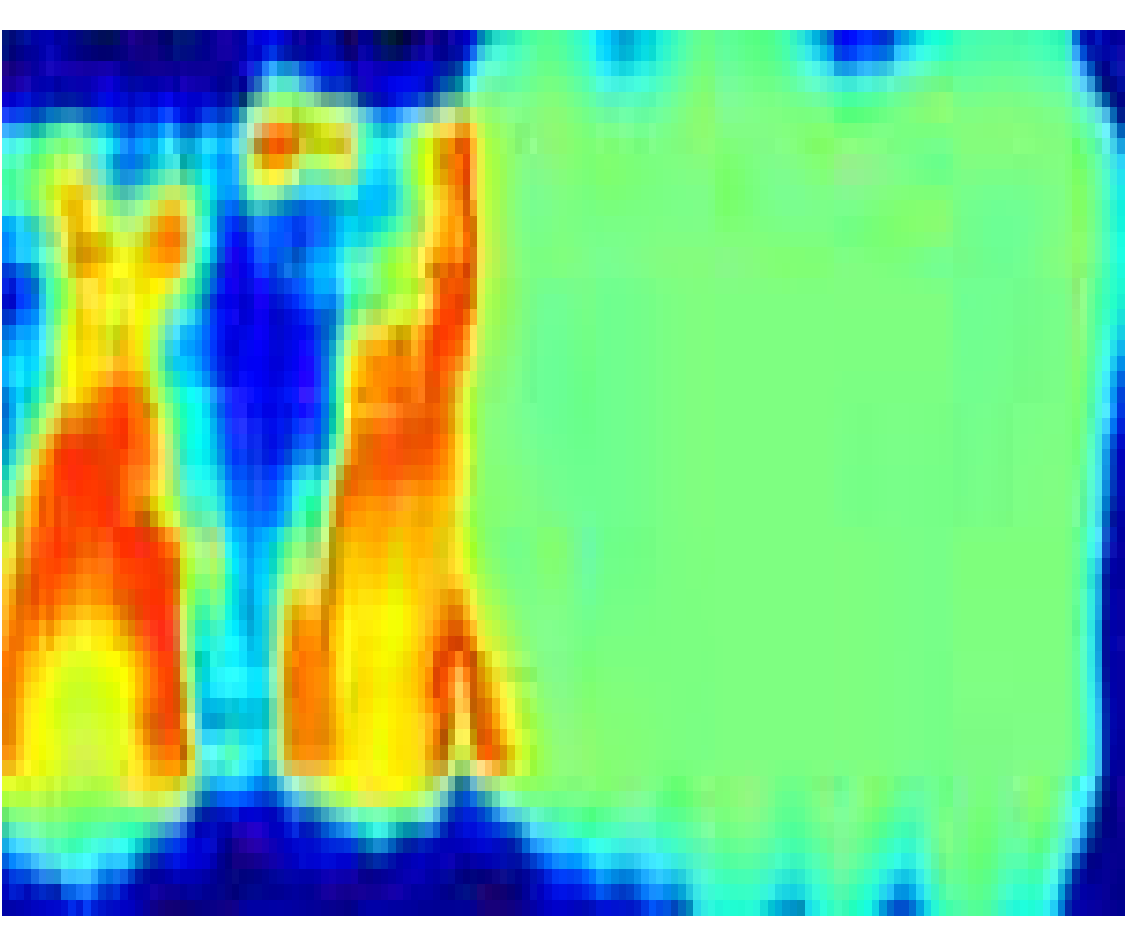}
	\captionsetup{labelformat=empty,skip=0pt}
	\caption{(x) FCM:3}	\label{fig:3h}
\end{subfigure}	
\begin{subfigure}[t]{0.015\textwidth}
	\centering
	\includegraphics[width=0.91\linewidth]{jetcolorbar.pdf}
	\captionsetup{labelformat=empty,skip=0pt}
\end{subfigure}
	\setcounter{subfigure}{24}
	
\begin{subfigure}[t]{0.11\textwidth}
		\centering
		\includegraphics[width=1\linewidth,height=0.8\linewidth]{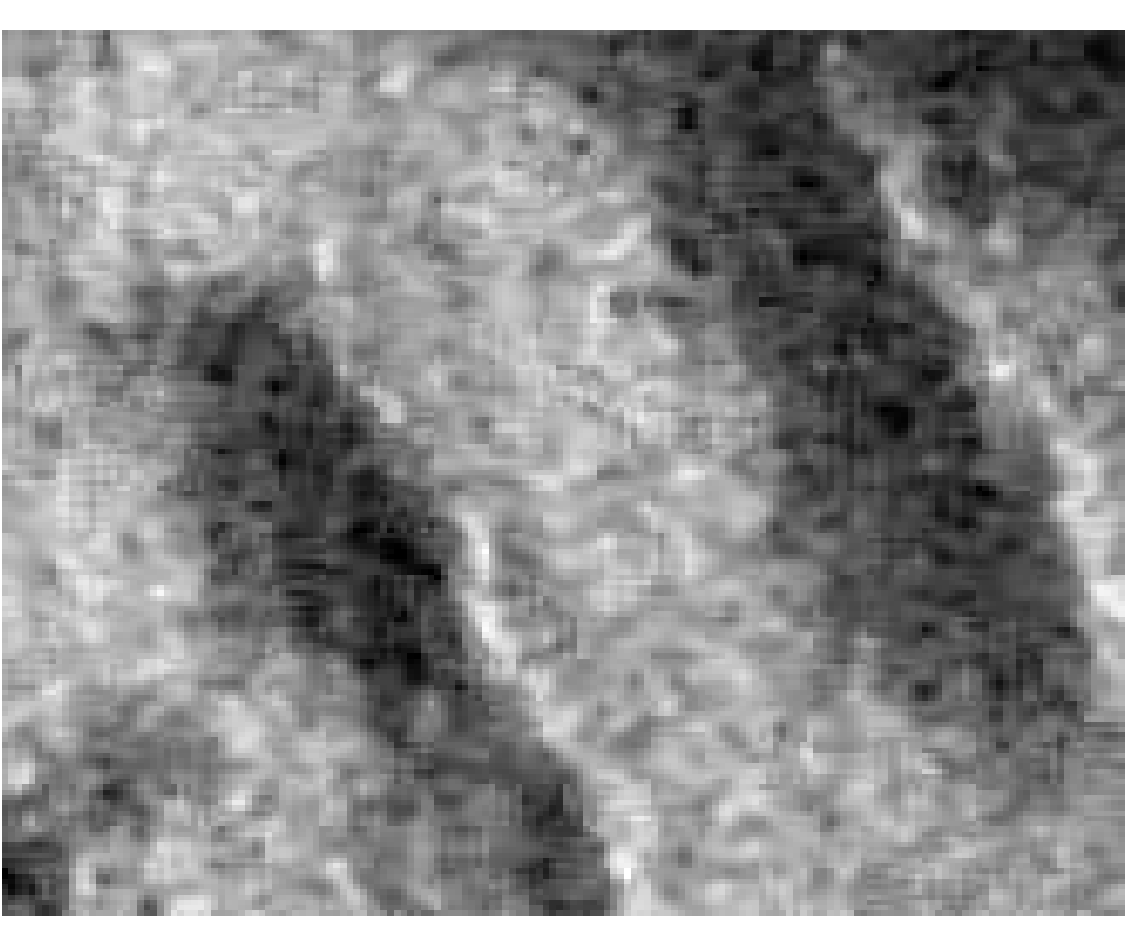}
		\captionsetup{labelformat=empty,skip=0pt}
		\caption{(y) Image 4} \label{fig:c4}
	\end{subfigure}
	\begin{subfigure}[t]{0.11\textwidth}
		\centering
		\includegraphics[width=1\linewidth,height=0.8\linewidth]{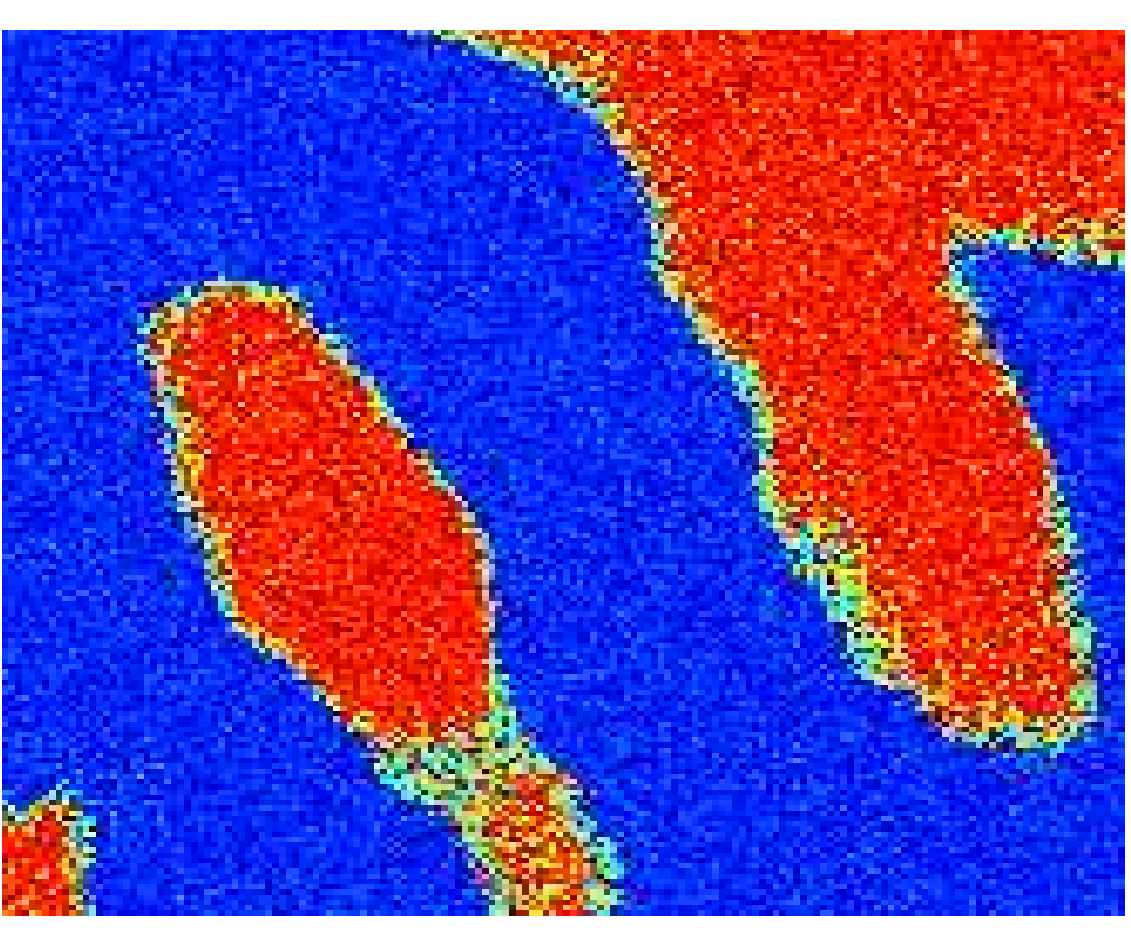}
		\captionsetup{labelformat=empty,skip=0pt}
		\caption{(z) PM-LDA:1}
	\end{subfigure}
	\begin{subfigure}[t]{0.11\textwidth}
		\centering
		\includegraphics[width=1\linewidth,height=0.8\linewidth]{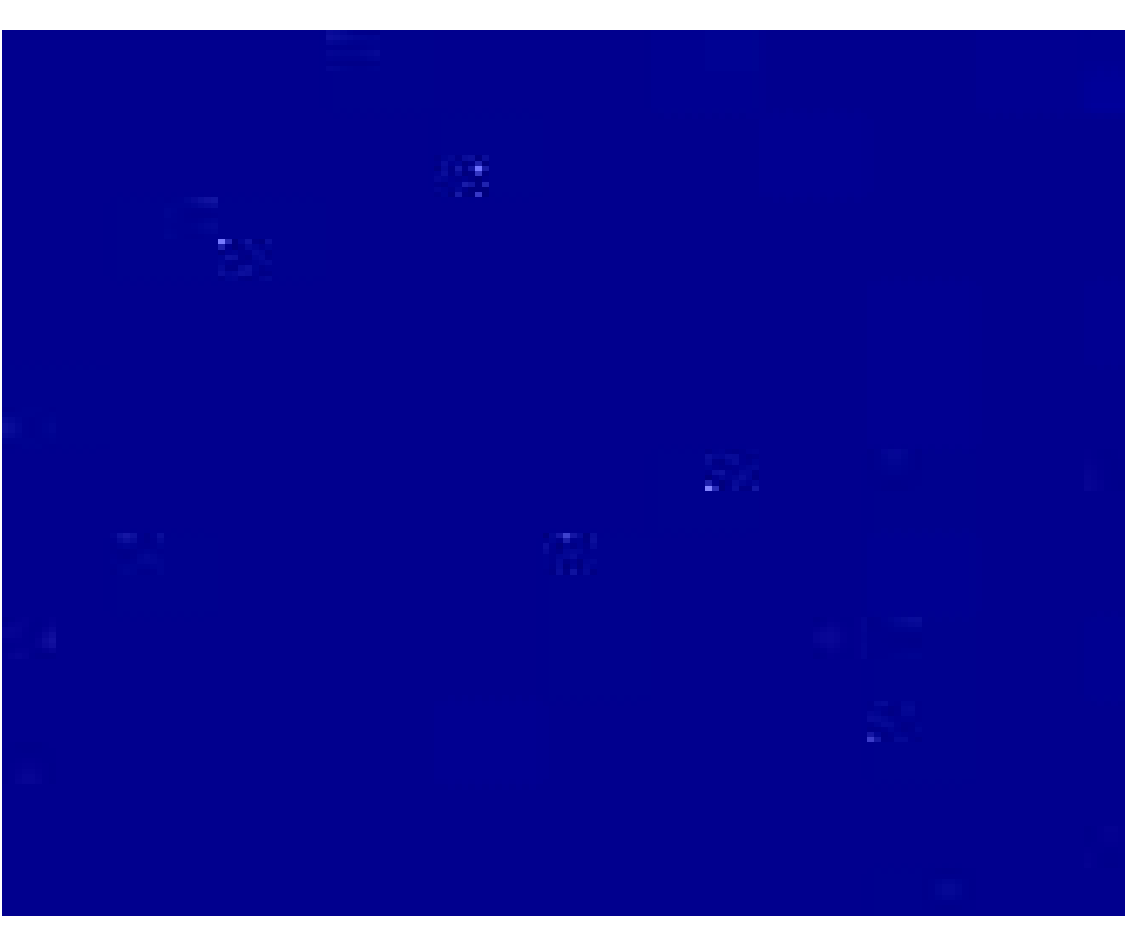}
		\captionsetup{labelformat=empty,skip=0pt}
		\caption{(aa) PM-LDA:2}
	\end{subfigure}
	\begin{subfigure}[t]{0.11\textwidth}
		\centering
		\includegraphics[width=1\linewidth,height=0.8\linewidth]{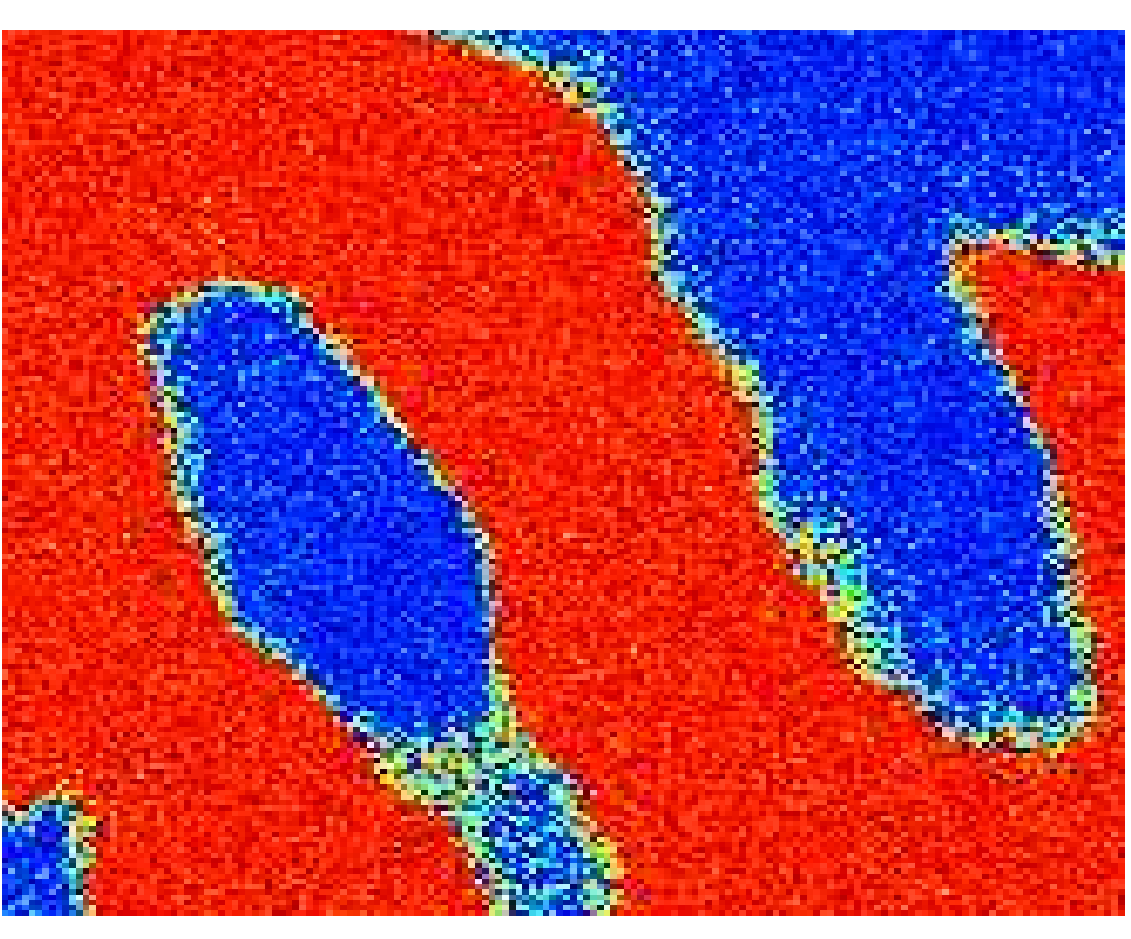}
		\captionsetup{labelformat=empty,skip=0pt}
		\caption{(ab) PM-LDA:3} \label{fig:4d}
	\end{subfigure}
	\begin{subfigure}[t]{0.015\textwidth}
		\centering
		\includegraphics[width=0.91\linewidth]{jetcolorbar.pdf}
		\captionsetup{labelformat=empty,skip=0pt}
	\end{subfigure}
	\setcounter{subfigure}{28}
	\begin{subfigure}[t]{0.11\textwidth}
		\centering
		\includegraphics[width=1\linewidth,height=0.8\linewidth]{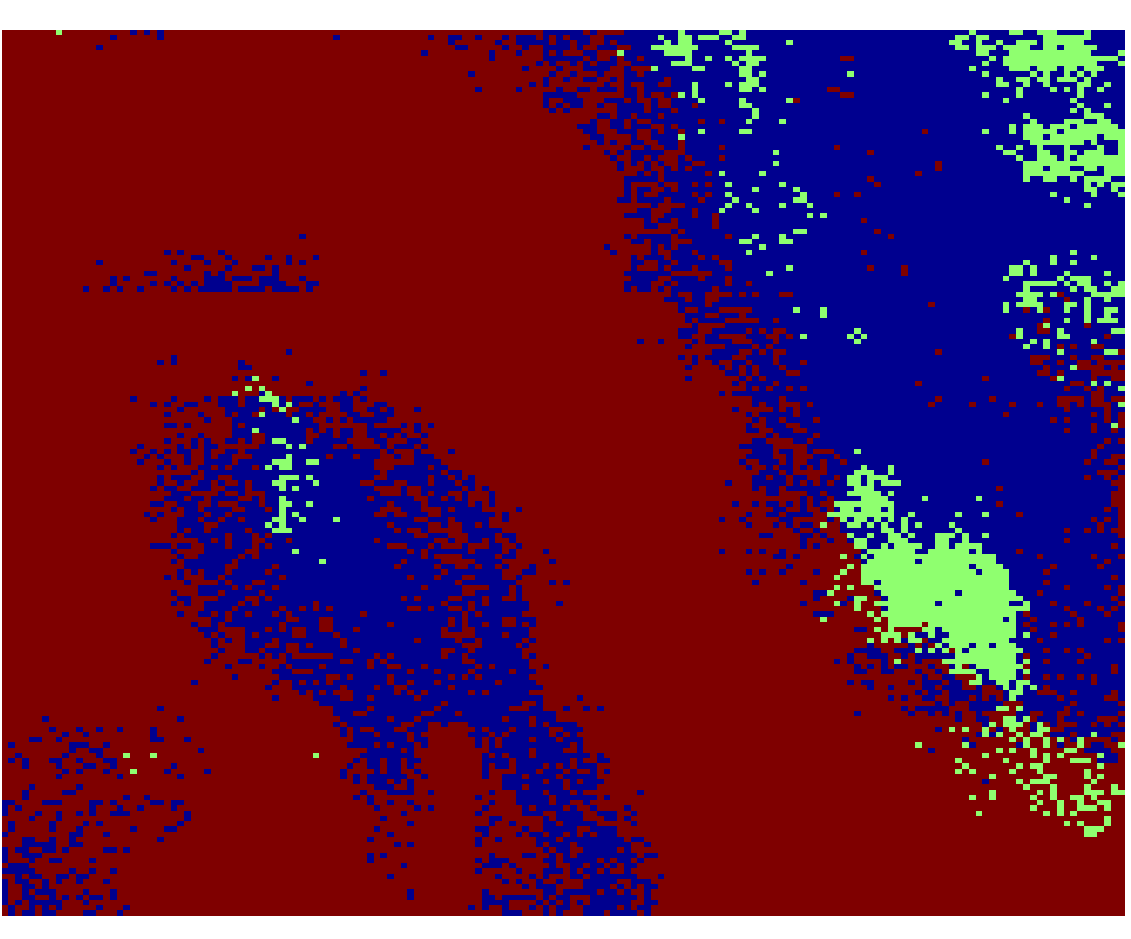}
		\captionsetup{labelformat=empty,skip=0pt}
		\caption{(ac) LDA}\label{fig:4e}
	\end{subfigure} 
	\begin{subfigure}[t]{0.11\textwidth}
		\centering
		\includegraphics[width=1\linewidth,height=0.8\linewidth]{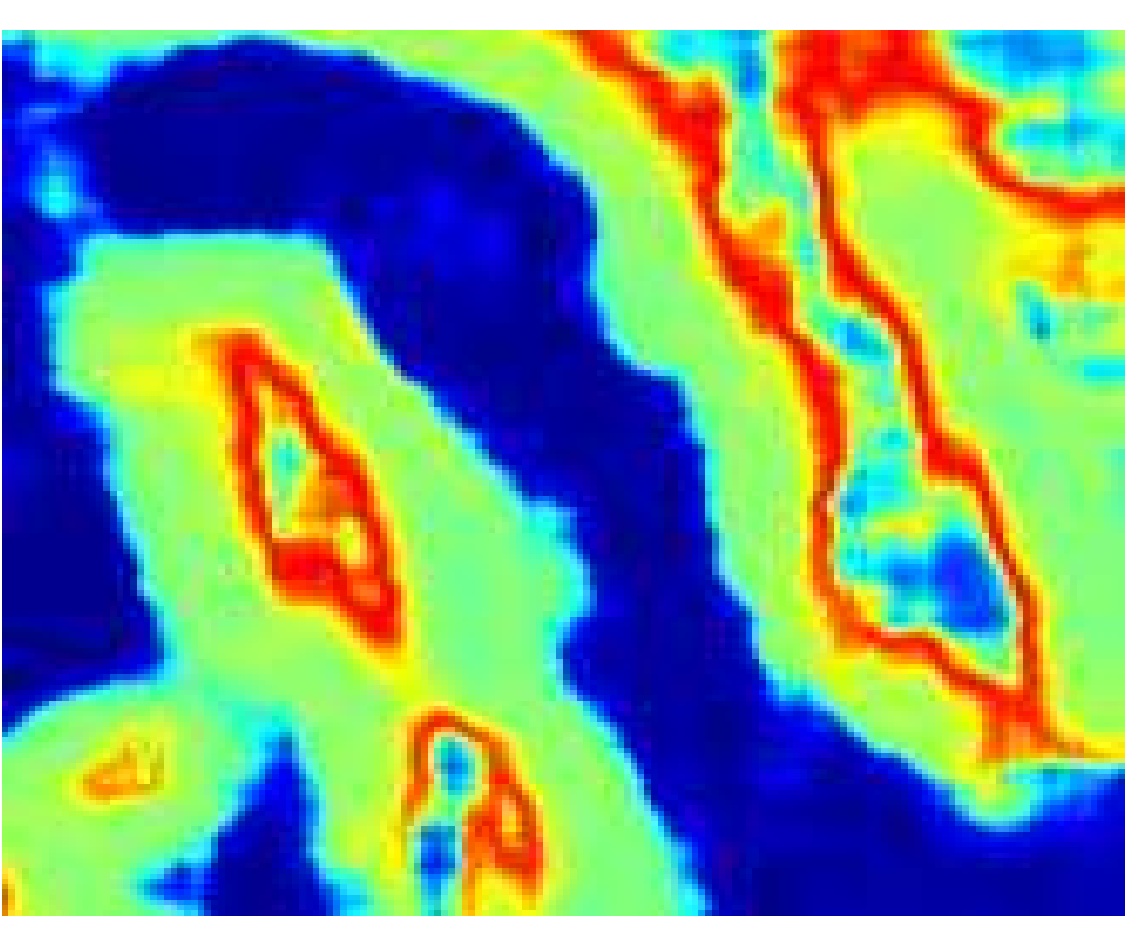}
		\captionsetup{labelformat=empty,skip=0pt}
		\caption{(ad) FCM:1}
	\end{subfigure}
	\begin{subfigure}[t]{0.11\textwidth}
		\centering
		\includegraphics[width=1\linewidth,height=0.8\linewidth]{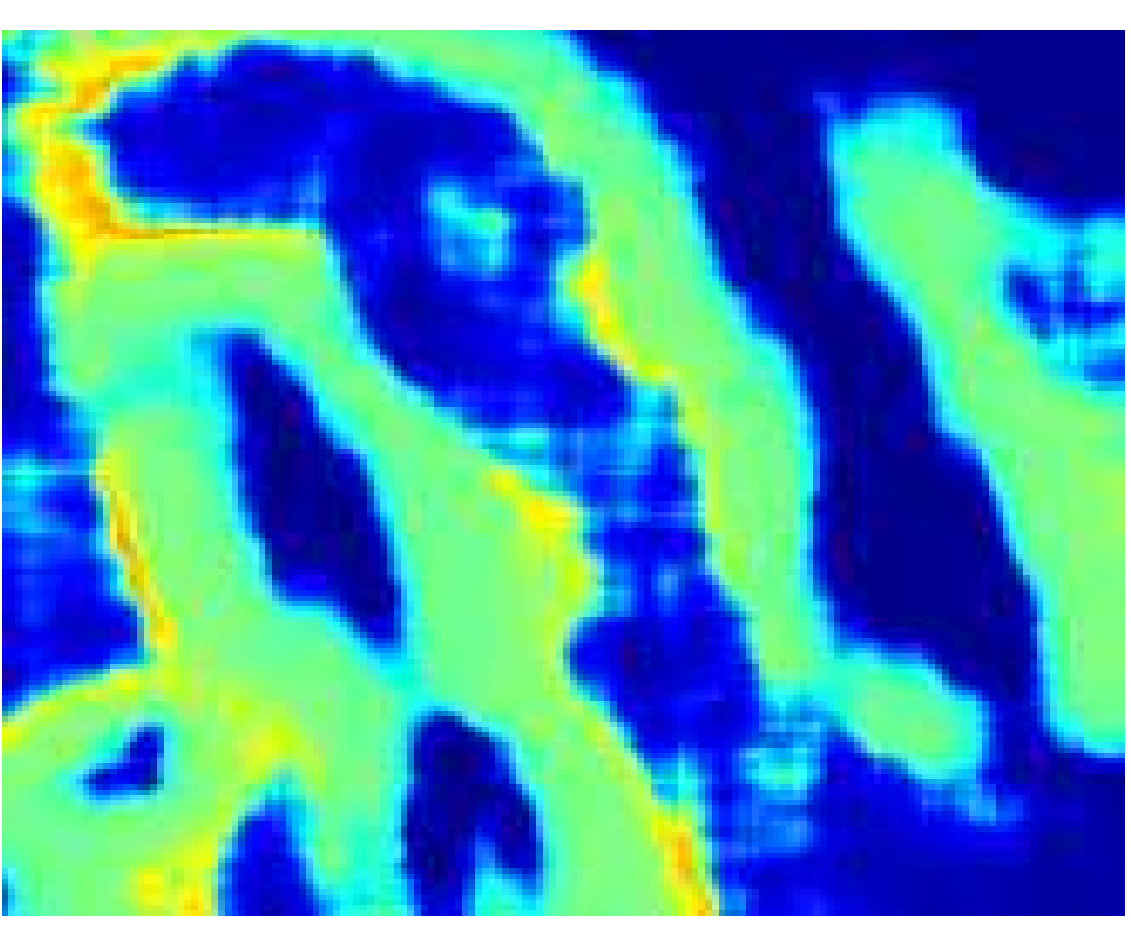}
		\captionsetup{labelformat=empty,skip=0pt}
		\caption{(ae) FCM:2}
	\end{subfigure}
	\begin{subfigure}[t]{0.11\textwidth}
		\centering
		\includegraphics[width=1\linewidth,height=0.8\linewidth]{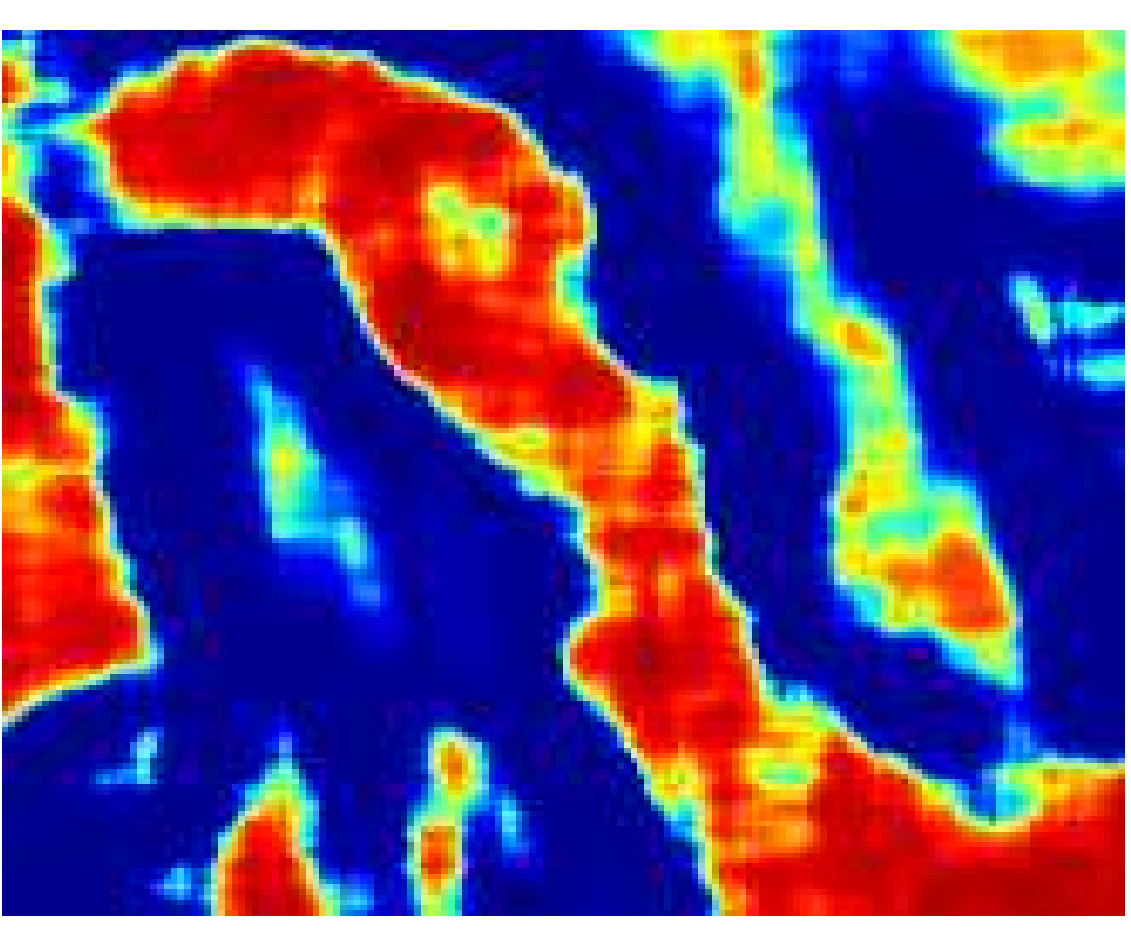}
		\captionsetup{labelformat=empty,skip=0pt}
		\caption{(af) FCM:3}
	\end{subfigure}		
	\begin{subfigure}[t]{0.015\textwidth}
		\centering
		\includegraphics[width=0.91\linewidth]{jetcolorbar.pdf}
		\captionsetup{labelformat=empty,skip=0pt}
	\end{subfigure}

	\caption{Segmentation results of Image 1 - 4 using PM-LDA, FCM and LDA. (a): SAS Image. (b)-(d): PM-LDA partial membership map in the ``dark flat sand,'' ``sand ripple,'' ``bright flat sand'' topics, respectively. (e): LDA result where color indicates topic label. (f)-(h): FCM  partial membership map in the first, second, and third cluster, respectively. Subfigure captions in Row 2 - 4 follow those in Row 1. In PM-LDA and FCM results,  color indicates the degree of membership of a visual word in a topic or cluster. }
	\label{fig:all}
\end{figure*}

\paragraph{Synthetic Aperture Sonar (SAS) Imagery Dataset}  From our SAS image dataset, we selected 4 images (shown in the first column in Fig. \ref{fig:all}) and compute the average intensity value and entropy within a 21 $\times$ 21 window as feature values. The average intensity value is scaled ($\times 10$) to roughly the same magnitude of the average entropy value. Each image is divided into multiple documents using a sliding window approach. A document consists of all of the feature vectors associated with each pixel (\ie, visual words) in the window. The number of topics in this dataset is set to $3$. For LDA, a dictionary of size $100$ is built by clustering all the computed feature values using the K-means.  FCM results with $m=1.5$. Parameters for LDA and FCM were selected manually to provide the best results. Due to the lack of ground-truth, qualitative segmentation results in Fig. \ref{fig:all} is provided. In the first row, Subfigures (b), (c), and (d) show the partial membership maps in the ``dark flat sand" , ``sand ripple" , and  ``bright flat sand'' topics using PM-LDA, respectively.  Subfigures (f), (g), and (h) show the partial membership maps in each of the three clusters using FCM, respectively.  In (b) - (d) and (f) - (h), the color indicates the degree of membership of a visual word in a topic where red corresponds to a full membership of $1$ and dark blue color corresponds to a membership value of $0$. The LDA result is shown in (e) where color indicates topic assignment. Subfigures in Row 2-4 follow the same subfigure captions in Row 1.

From the experimental results,  we can see that PM-LDA achieves much better results than FCM and LDA. As shown in Fig. \ref{fig:1c} and \ref{fig:2c}, the segmentation results of PM-LDA show a gradual change from ``sand ripple" to ``dark flat sand".   FCM captures the gradual transition to some extent but is not able to clearly differentiate between clusters. For example, as shown in Fig. \ref{fig:2g} - \ref{fig:2h} and Fig. \ref{fig:3g} - \ref{fig:3h},  using FCM, the rippled region in Images 2 and 3 is assigned to 2 clusters with nearly equal partial memberships.  As LDA cannot generate partial memberships, in Fig. \ref{fig:1e} and \ref{fig:2e}, Image 1 and 2 are simply partitioned into different topics using LDA. Yet, by comparing Fig. \ref{fig:3e} with \ref{fig:3d} and Fig. \ref{fig:4e} with \ref{fig:4d}, we can see that on Image 3 and 4 that do not contain transition regions, LDA achieves similar segmentation result to PM-LDA.

\begin{figure}[htb]
	\centering
	\includegraphics[height=0.18\linewidth]{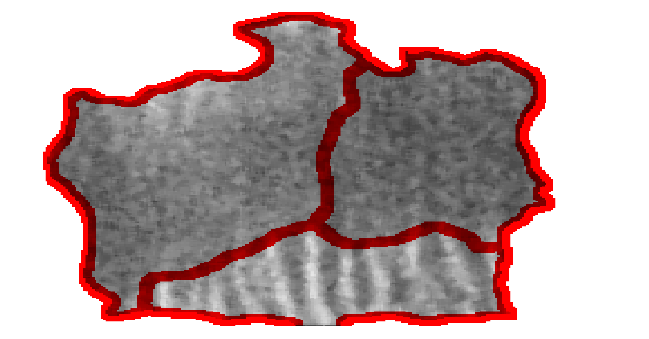}
	\caption{A subregion of three superpixels}
	\label{fig:3sup}
	\vspace{-2mm}
\end{figure}

\begin{figure}[!htb]
	\centering
	\begin{subfigure}[t]{0.12\textwidth}
		\centering
		\includegraphics[height=0.5\linewidth]{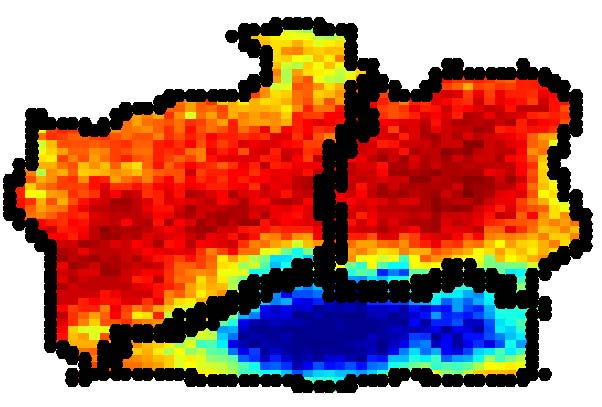}
		\captionsetup{labelformat=empty,skip=0pt}
		\caption{(a) $s=3$}
	\end{subfigure} 
	\begin{subfigure}[t]{0.12\textwidth}
		\centering
		\includegraphics[height=0.5\linewidth]{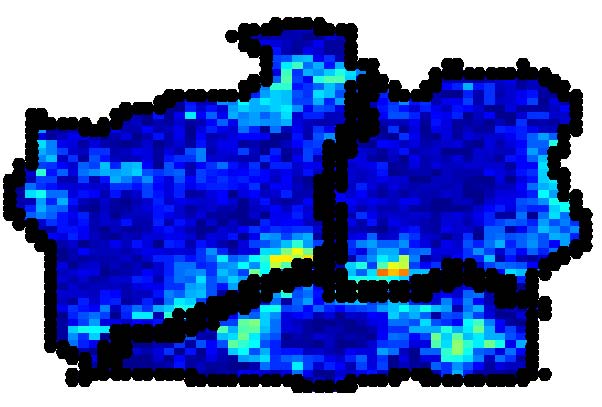}
		\captionsetup{labelformat=empty,skip=0pt}
		\caption{(b) }
	\end{subfigure} 
	\begin{subfigure}[t]{0.12\textwidth}
		\centering
		\includegraphics[height=0.5\linewidth]{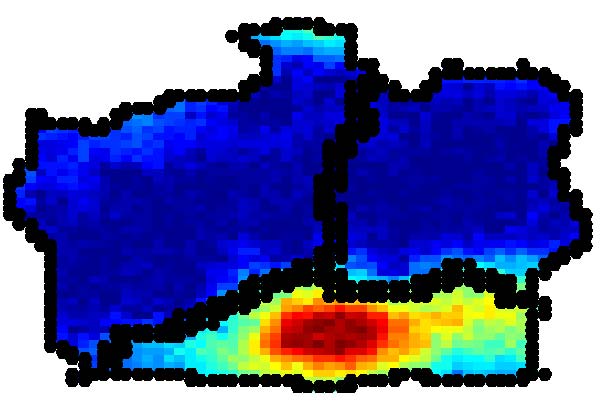}
		\captionsetup{labelformat=empty,skip=0pt}
		\caption{(c) }
	\end{subfigure}

	\begin{subfigure}[t]{0.12\textwidth}
		\centering
		\includegraphics[height=0.5\linewidth]{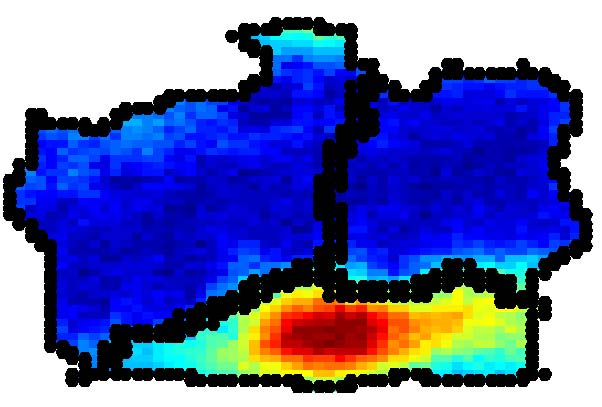}
		\captionsetup{labelformat=empty,skip=0pt}
		\caption{(a) $s=10$}
	\end{subfigure}
	\begin{subfigure}[t]{0.12\textwidth}
		\centering
		\includegraphics[height=0.5\linewidth]{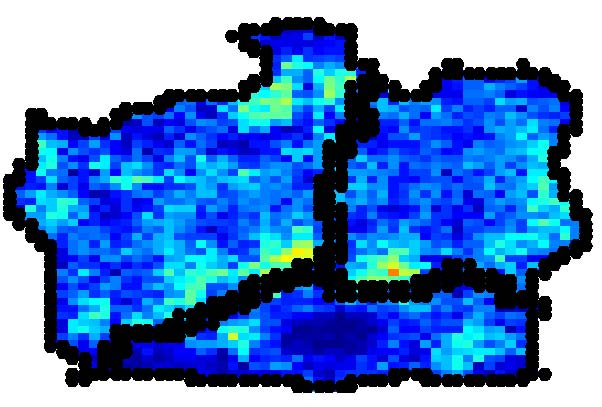}
		\captionsetup{labelformat=empty,skip=0pt}
		\caption{(b)}
	\end{subfigure}
	\begin{subfigure}[t]{0.12\textwidth}
		\centering
		\includegraphics[height=0.5\linewidth]{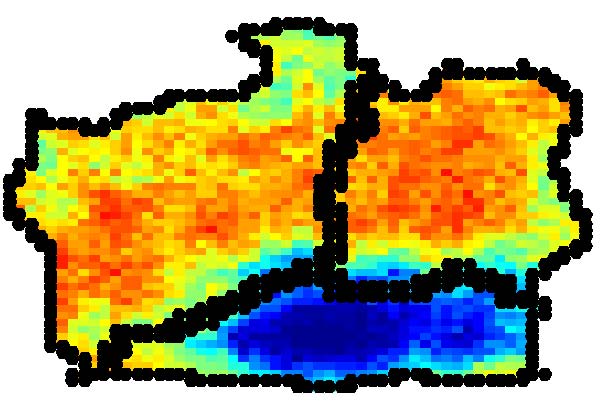}
		\captionsetup{labelformat=empty,skip=0pt}
		\caption{(c)}
	\end{subfigure}

	\begin{subfigure}[t]{0.12\textwidth}
		\centering
		\includegraphics[height=0.5\linewidth]{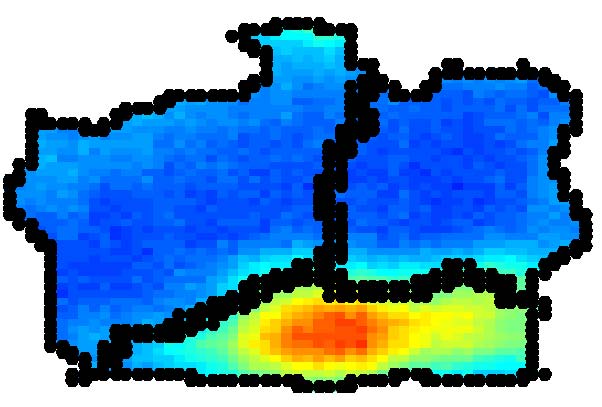}
		\captionsetup{labelformat=empty,skip=0pt}
		\caption{(a) $s=300$}
	\end{subfigure}
	\begin{subfigure}[t]{0.12\textwidth}
		\centering
		\includegraphics[height=0.5\linewidth]{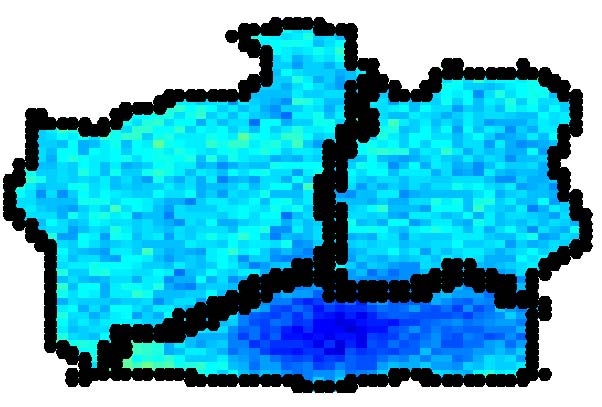}
		\captionsetup{labelformat=empty,skip=0pt}
		\caption{(b) }
	\end{subfigure}
	\begin{subfigure}[t]{0.12\textwidth}
		\centering
		\includegraphics[height=0.5\linewidth]{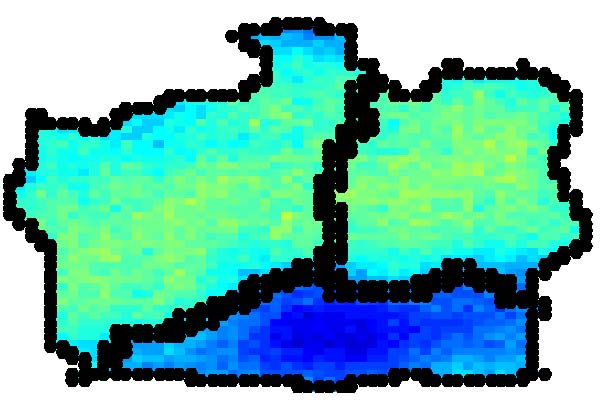}
		\captionsetup{labelformat=empty,skip=0pt}
		\caption{(c) }
	\end{subfigure}

	\begin{subfigure}[t]{0.12\textwidth}
		\centering
		\includegraphics[height=0.5\linewidth]{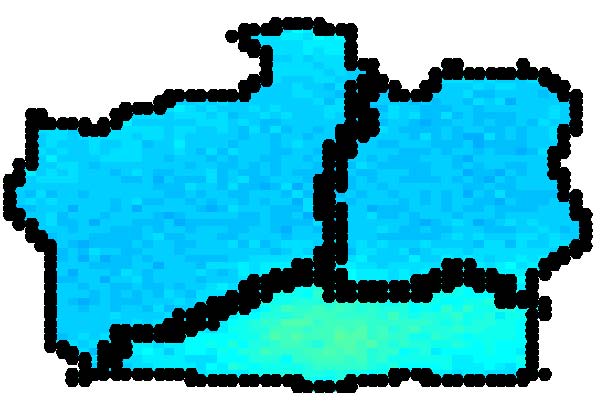}
		\captionsetup{labelformat=empty,skip=0pt}
		\caption{(a) $s=30000$}
	\end{subfigure}
	\begin{subfigure}[t]{0.12\textwidth}
		\centering
		\includegraphics[height=0.5\linewidth]{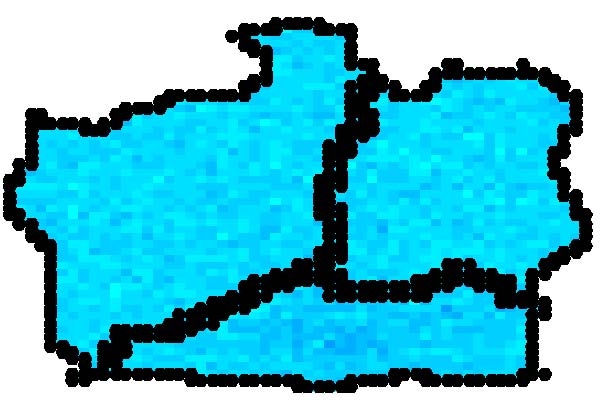}
		\captionsetup{labelformat=empty,skip=0pt}
		\caption{(b)}
	\end{subfigure}
	\begin{subfigure}[t]{0.12\textwidth}
		\centering
		\includegraphics[height=0.5\linewidth]{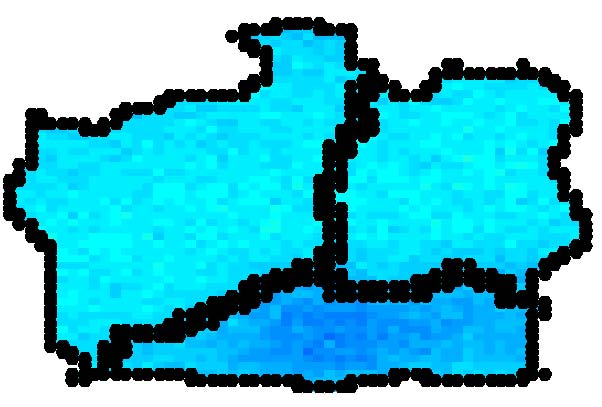}
		\captionsetup{labelformat=empty,skip=0pt}
		\caption{(c) }
	\end{subfigure}	

	\caption[Partial membership maps  with varying $s$]{Partial membership maps  with varying $s$. Each row shows the estimated membership maps of the three estimated topics. The black contour indicates the superpixel boundary. The superpixels are results published in \cite{cobb:2013multi}.}
	\label{fig:varyings}
	\vspace{-4mm}
\end{figure}

As discussed in Section \ref{sec:PMLDA}, the scaling factor $s^d$ determines the similarity of the partial membership vector of each word, $\mathbf{z}_n^d$, to the topic proportion $\boldsymbol{\pi}^d$.  In this experiment, we investigated the effect of $s$ by estimating the memberships and topics with fixed topic proportion. A subregion consisting of three superpixels \cite{cobb:2013multi} are used in this experiment and shown in Fig. \ref{fig:3sup}. Each superpixel is treated as a document. The topic proportion $\boldsymbol{\pi}^d$ is set to be $[1,1,1]/3$ and the scaling factor $s$ is varied to be $3, 10, 300, 30000$. The membership estimation results are shown in Figure \ref{fig:varyings}. As can be seen, as the scaling factor $s$ increases, the partial memberships gradually approach the topic proportion $[1,1,1]/3$ and become more smooth.

\paragraph{Sunset Dataset} Experimental results on Sunset dataset show the ability of PM-LDA to perform partial membership segmentation given visual natural imagery. Two sunset themed images from Flickr (with the necessary permissions)\footnote[2]{\scriptsize{Photo can be found at: https://www.flickr.com/photos/aoa-/6104409480/}}  \footnote[3]{\scriptsize{Photo can be found at: https://www.flickr.com/photos/frenchdave/8482336933/}} were used in this experiment. For each visual word, the first order Gaussian gradient ($\sigma=2$)  with respect to $y$-axis and R and B channels are used as the feature vectors. The number of topics is set to be $3$. Experiments are run on each image individually and results are shown in Fig. \ref{fig:sunset}. Columns 2-4 show the segmentation results of PM-LDA.  Column 5 is the LDA results with $3$ topics.  Comparing Column 3 and Column 5 in Fig. \ref{fig:sunset}, we can see that PM-LDA can generate continuous partial membership according to the extent to which the sky is colored by sunlight. The partial membership map illustrates how the topic gradually shift from one to the other. In contrast, LDA can only produce $0$-$1$ segmentation.

\begin{figure}[!htb]
	\centering
	\begin{subfigure}[b]{0.09\textwidth}
		\centering
		\includegraphics[width=1\linewidth]{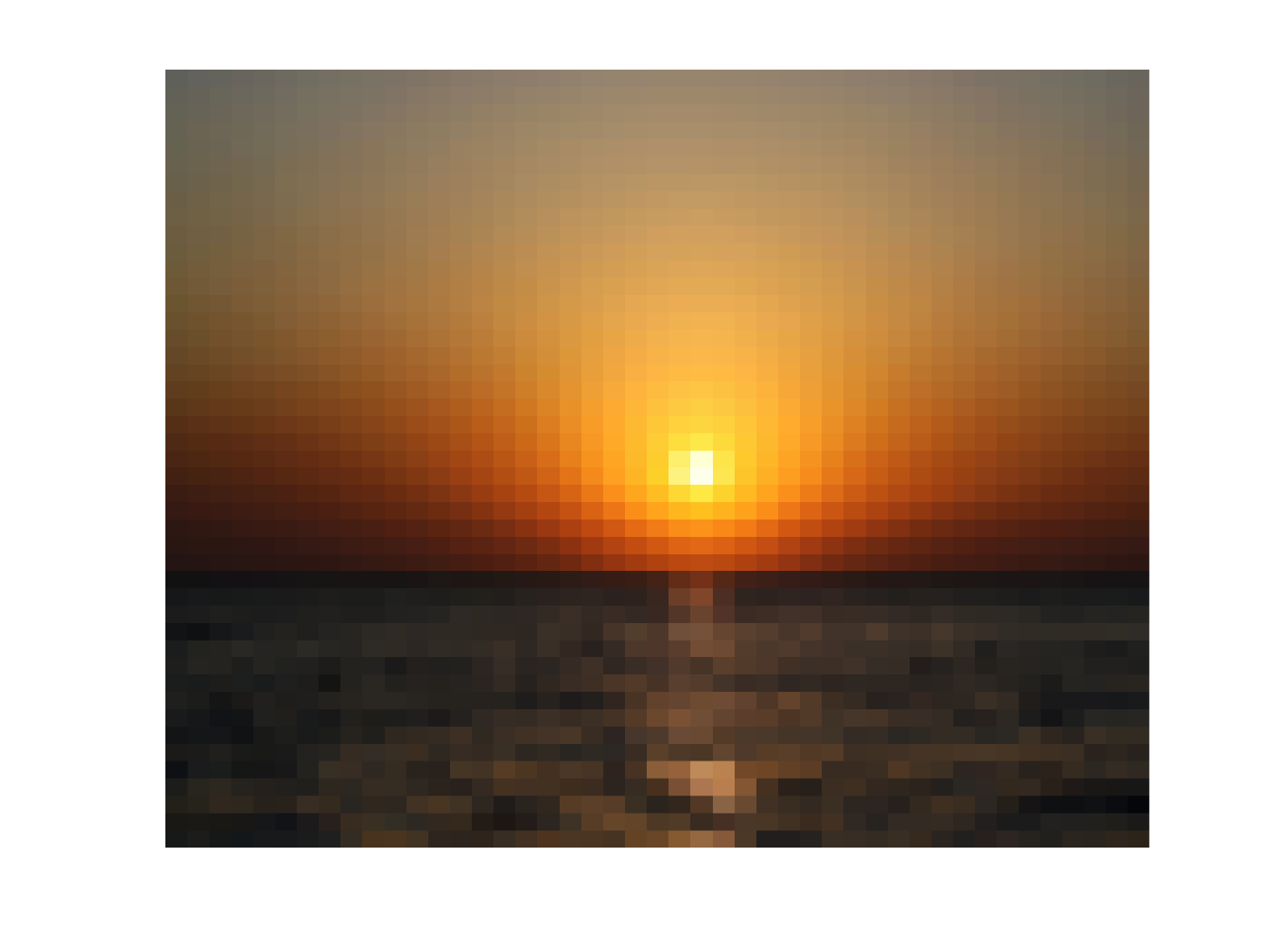}
		\caption{}
	\end{subfigure}
	\begin{subfigure}[b]{0.09\textwidth}
		\centering
		\includegraphics[width=1\linewidth]{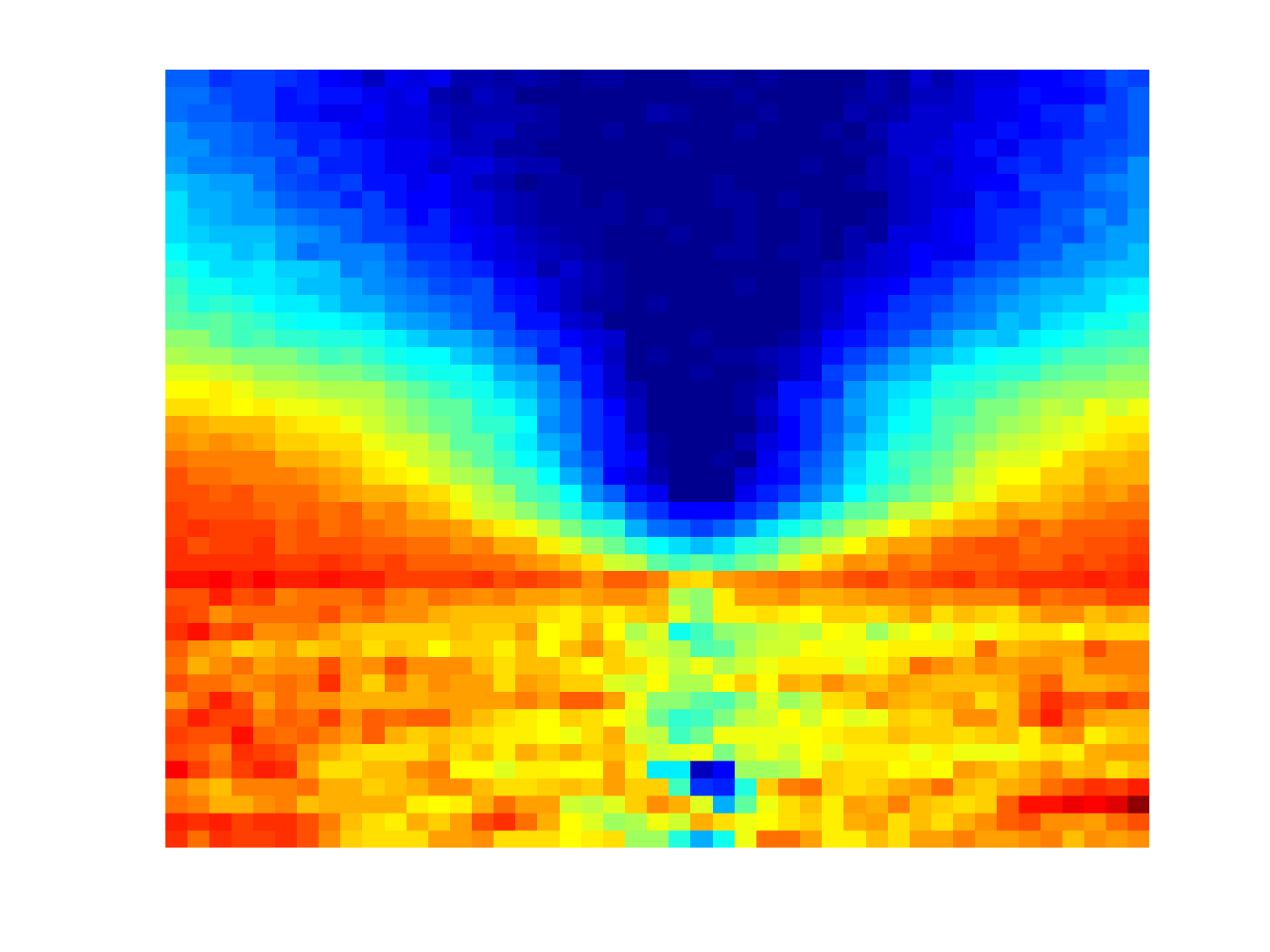}
		\caption{}
	\end{subfigure}
	\begin{subfigure}[b]{0.09\textwidth}
		\centering
		\includegraphics[width=1\linewidth]{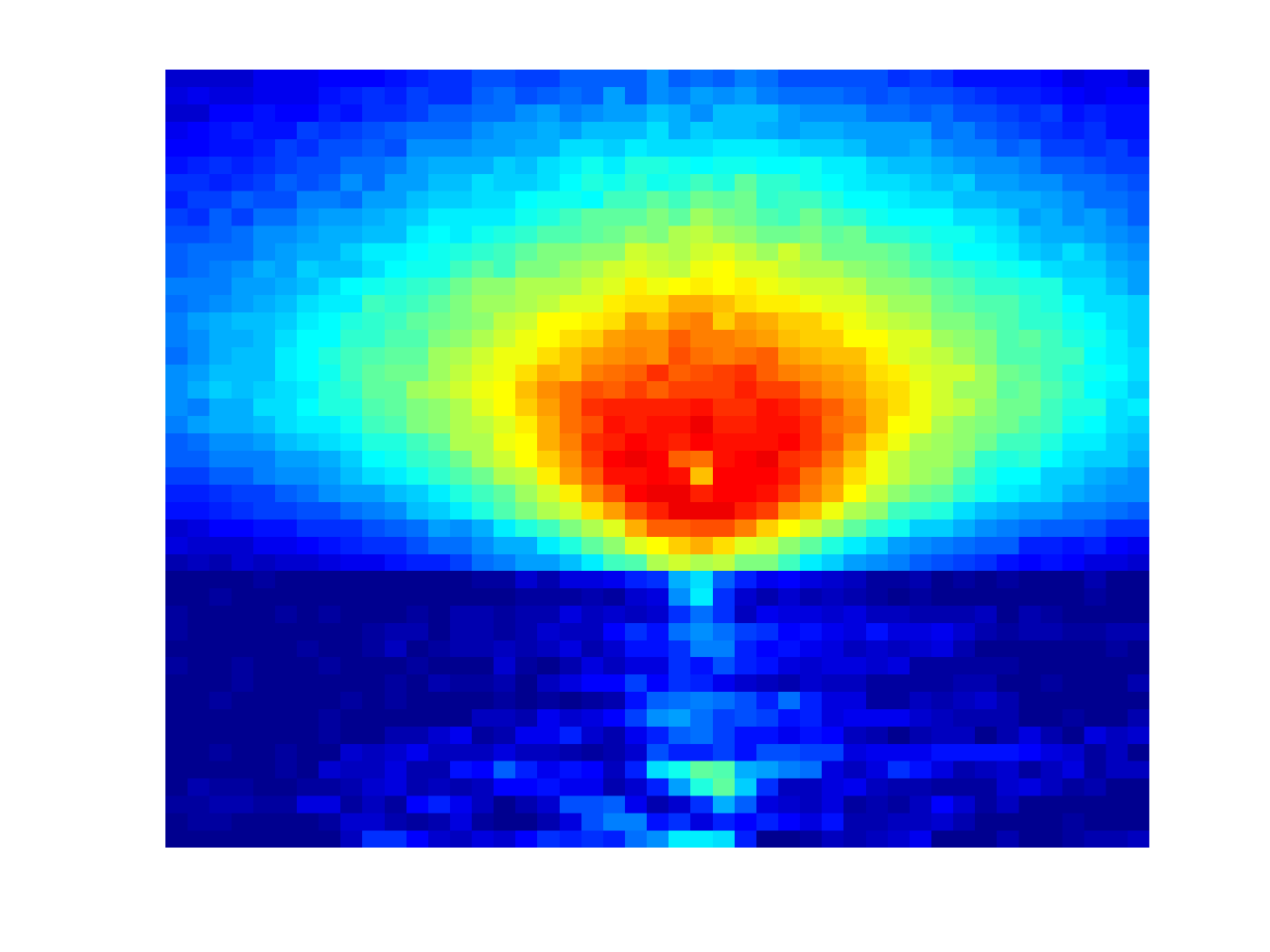}
		\caption{}
	\end{subfigure}
	\begin{subfigure}[b]{0.09\textwidth}
		\centering
		\includegraphics[width=1\linewidth]{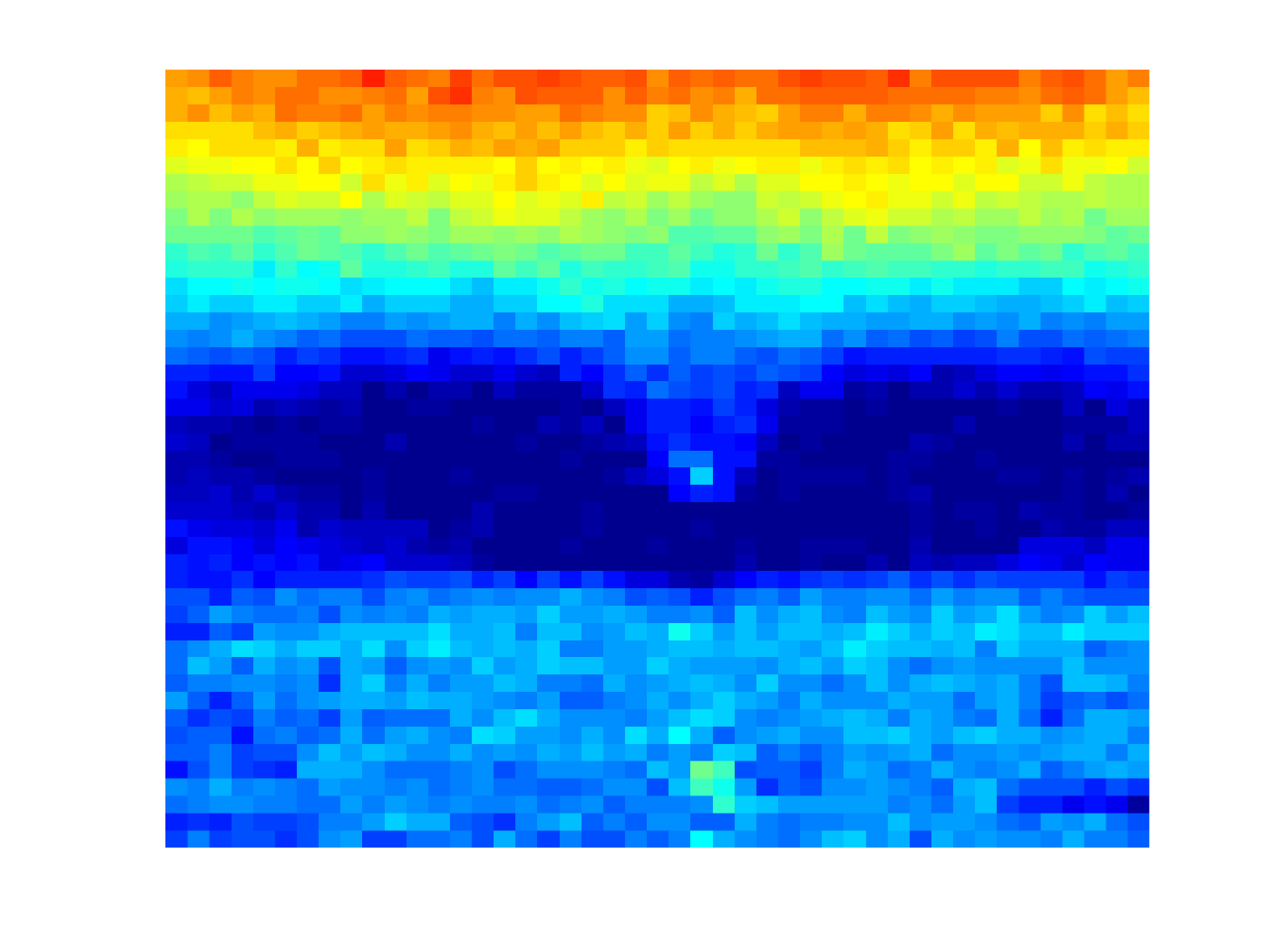}
		\caption{}
	\end{subfigure}
	\begin{subfigure}[b]{0.09\textwidth}
		\centering
		\includegraphics[width=1\linewidth]{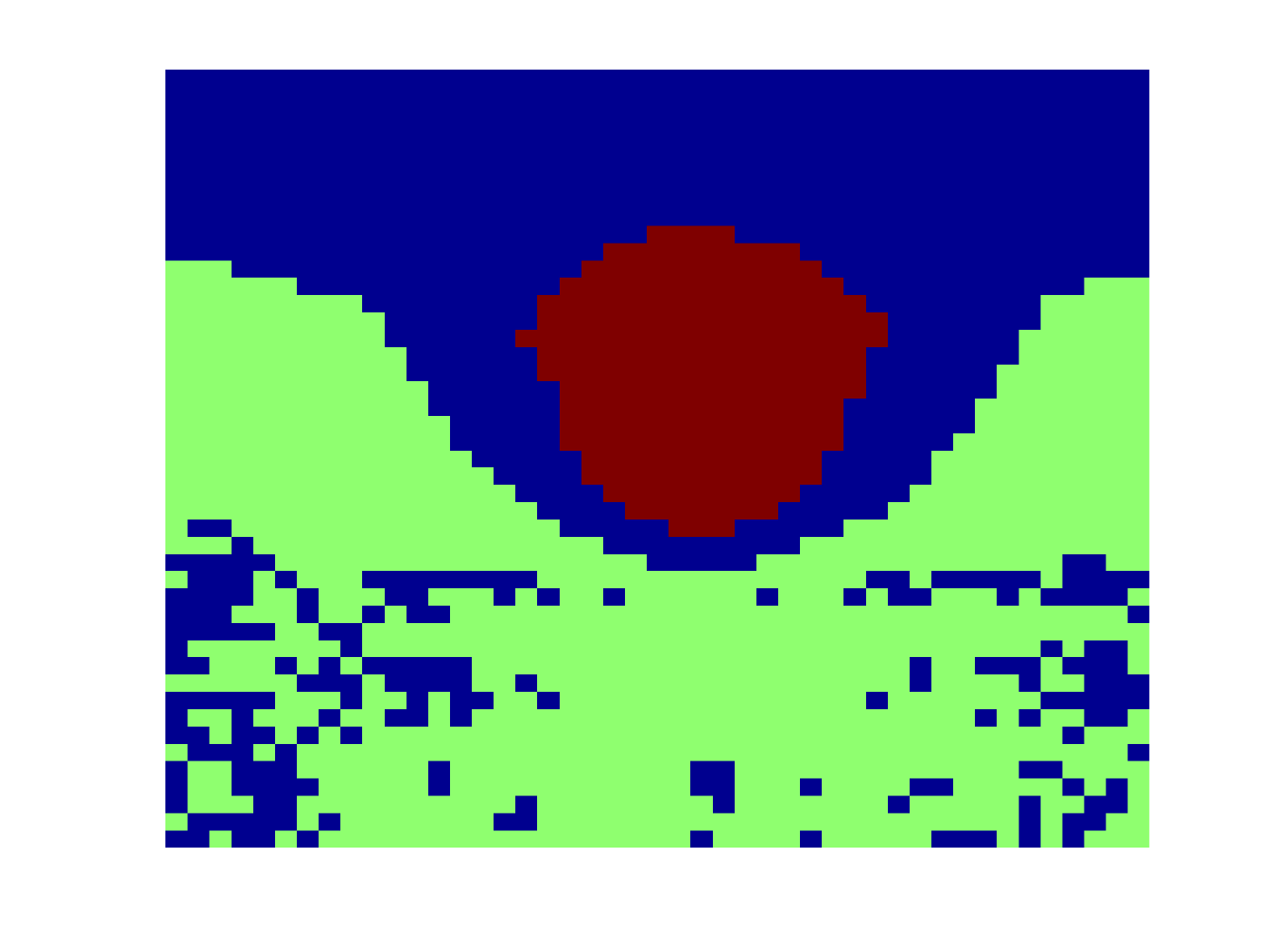}
		\caption{}
	\end{subfigure}
	
	\begin{subfigure}[b]{0.09\textwidth}
		\centering
		\includegraphics[width=1\linewidth]{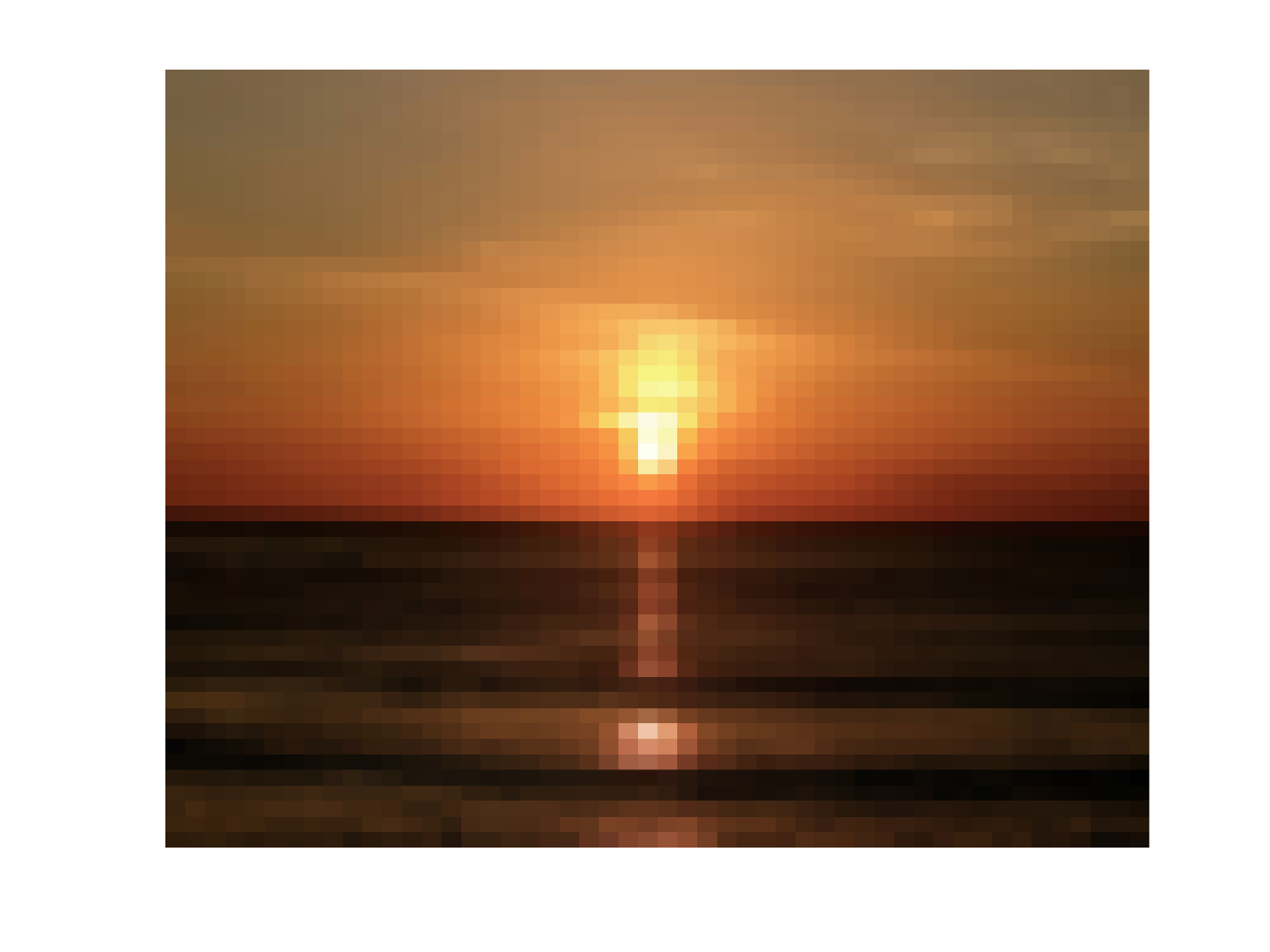}
		\caption{}
	\end{subfigure}
	\begin{subfigure}[b]{0.09\textwidth}
		\centering
		\includegraphics[width=1\linewidth]{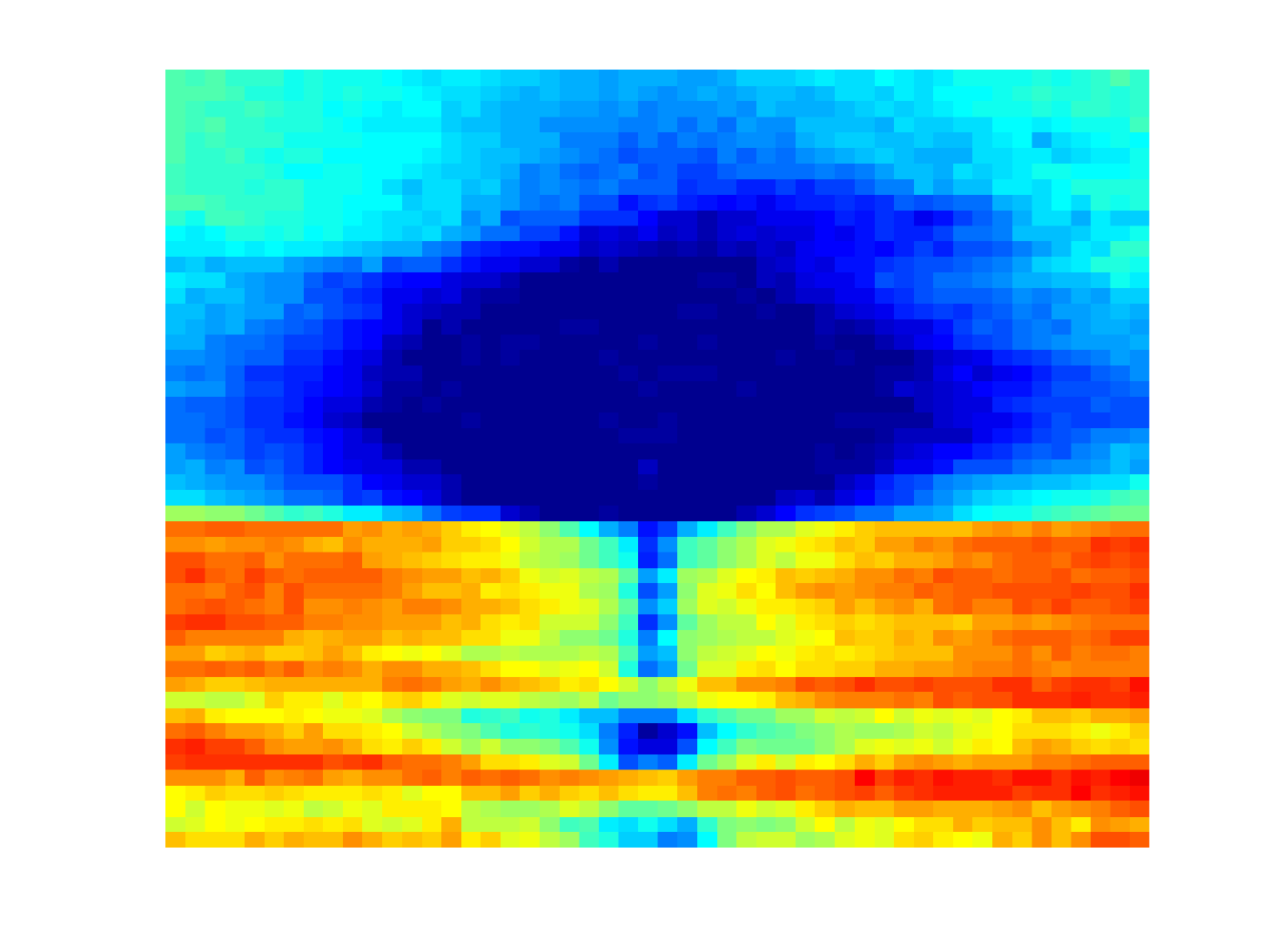}
		\caption{}
	\end{subfigure}
	\begin{subfigure}[b]{0.09\textwidth}
		\centering
		\includegraphics[width=1\linewidth]{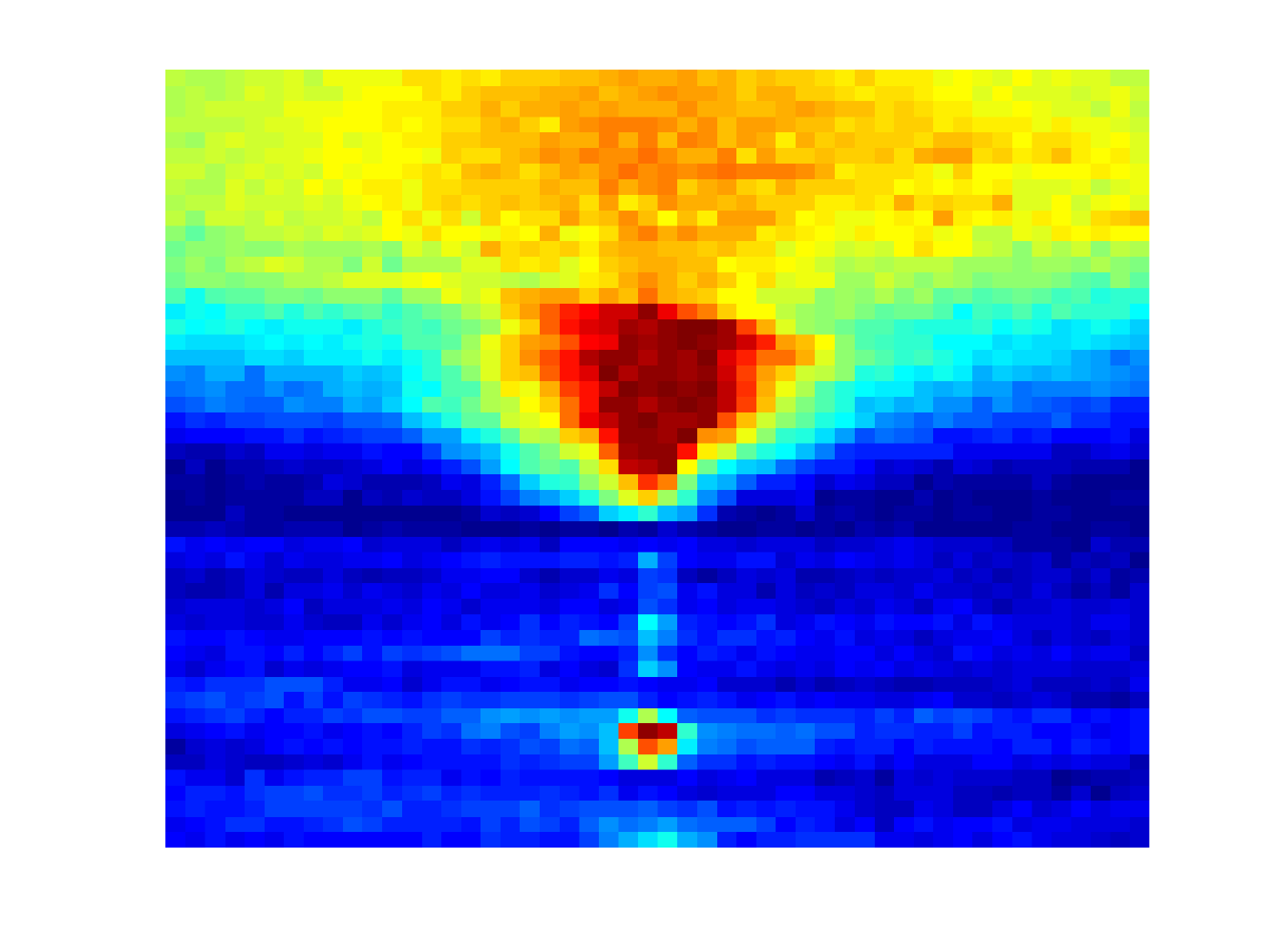}
		\caption{}
	\end{subfigure}
	\begin{subfigure}[b]{0.09\textwidth}
		\centering
		\includegraphics[width=1\linewidth]{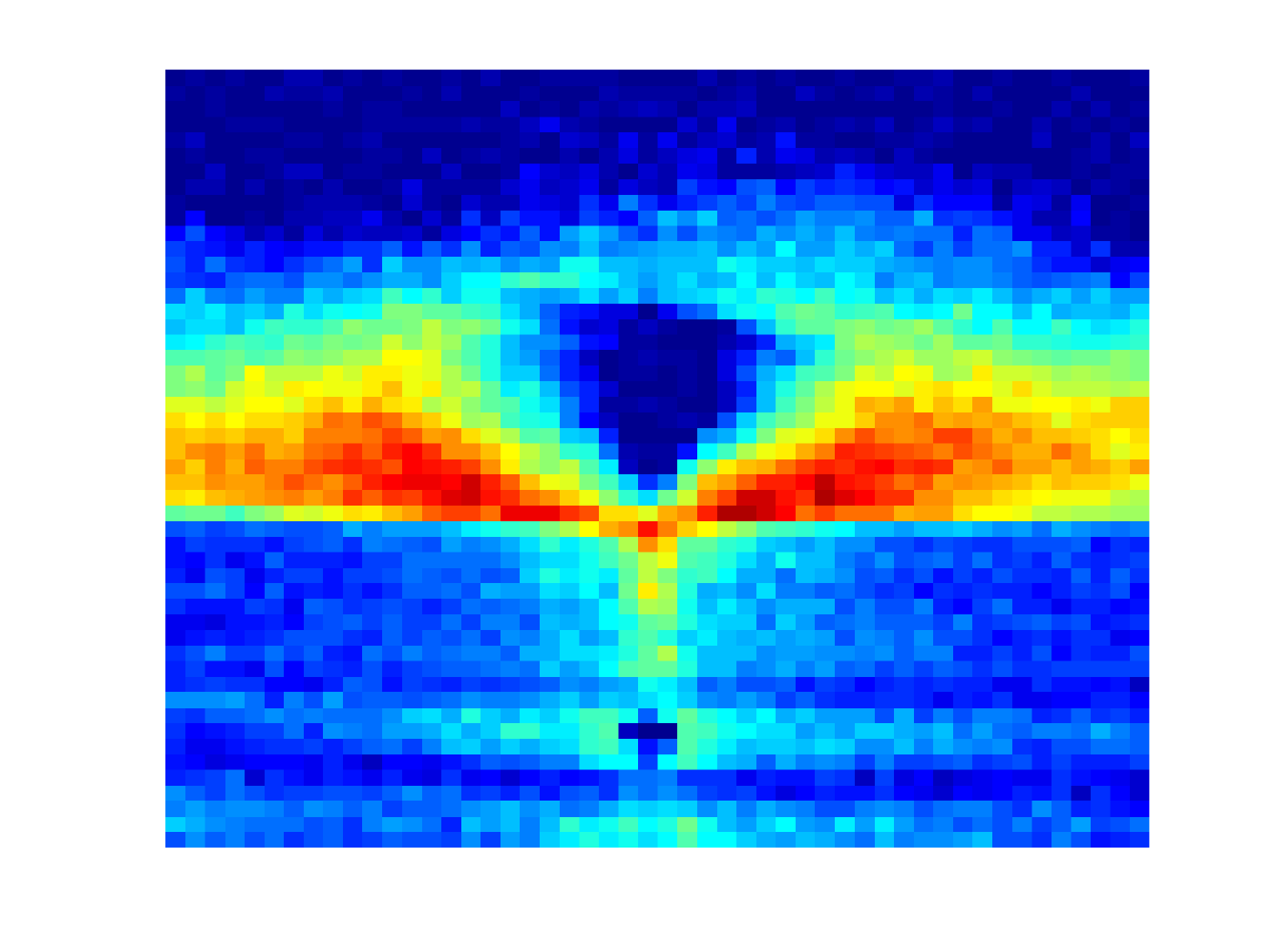}
		\caption{}
	\end{subfigure}
	\begin{subfigure}[b]{0.09\textwidth}
		\centering
		\includegraphics[width=1\linewidth]{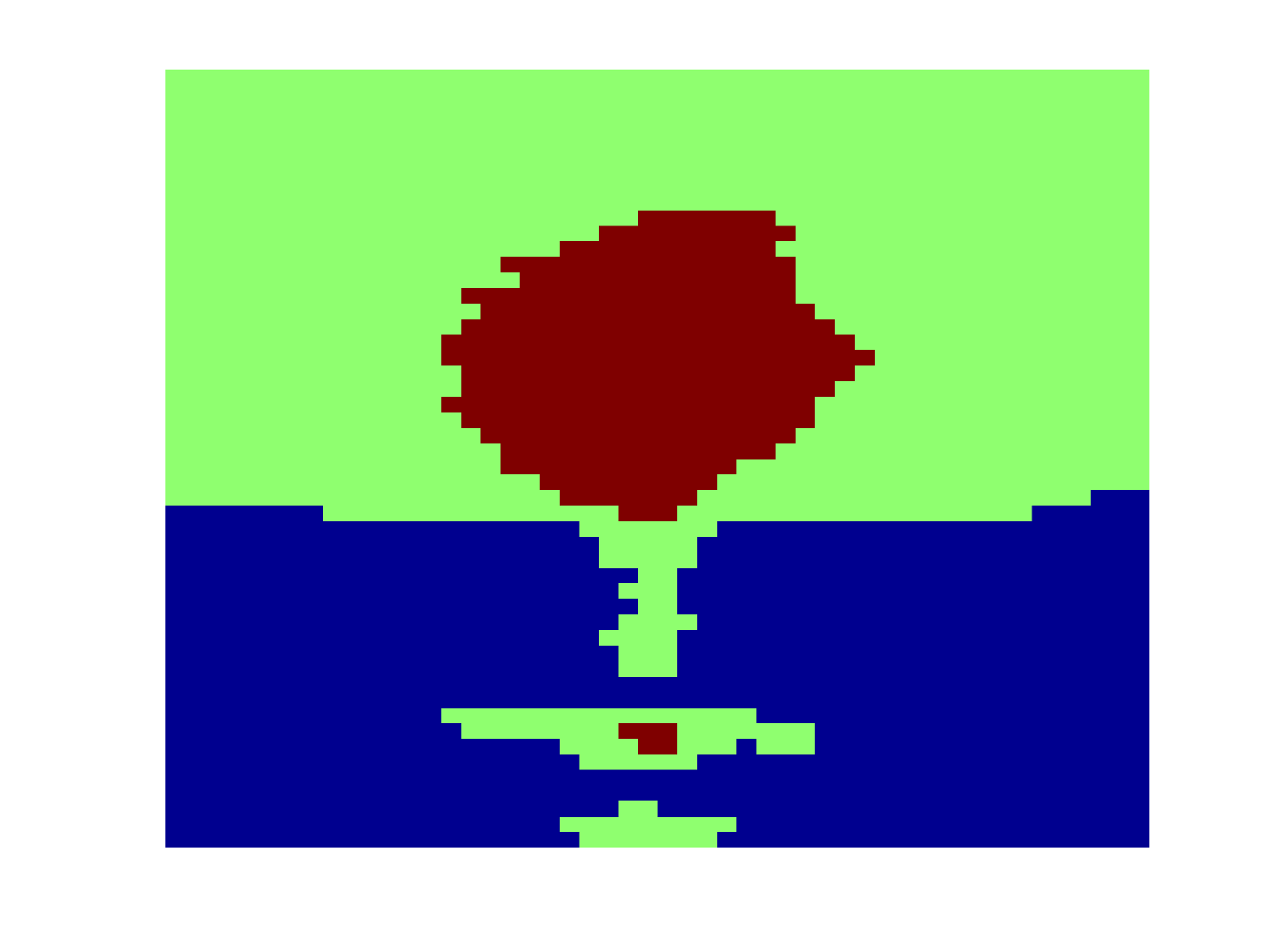}
		\caption{}
	\end{subfigure}
	\caption{Examples of segmentation result on Sunset dataset. (a): Sunset Image 1. (f): Sunset Image 2. (b)-(d) and (g)-(i) are the PM-LDA partial membership maps in the estimated three topics for Sunset Image 1 and Sunset Image 2, respectively. The color indicates the degree of membership of a visual word in a topic or cluster. (e) and (j) are the LDA results where color indicates the topic.}
	\label{fig:sunset}
\end{figure}

\begin{figure*}[!htb]
	\centering
	\begin{subfigure}[t]{0.13\textwidth}
		\centering
		\includegraphics[width=1\linewidth]{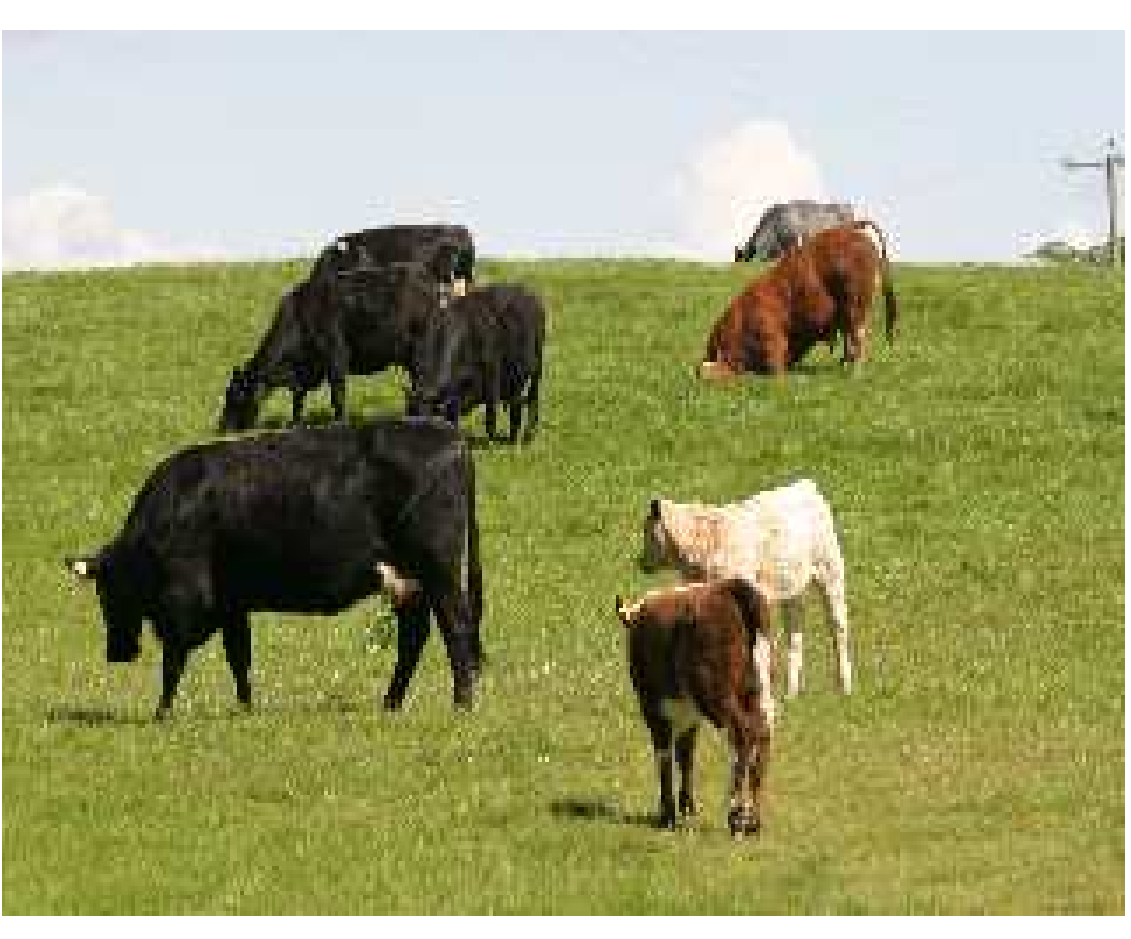}
		\captionsetup{labelformat=empty,skip=0pt}
		\caption{(a) cow1}
	\end{subfigure}
	\begin{subfigure}[t]{0.13\textwidth}
		\centering
		\includegraphics[width=1\linewidth]{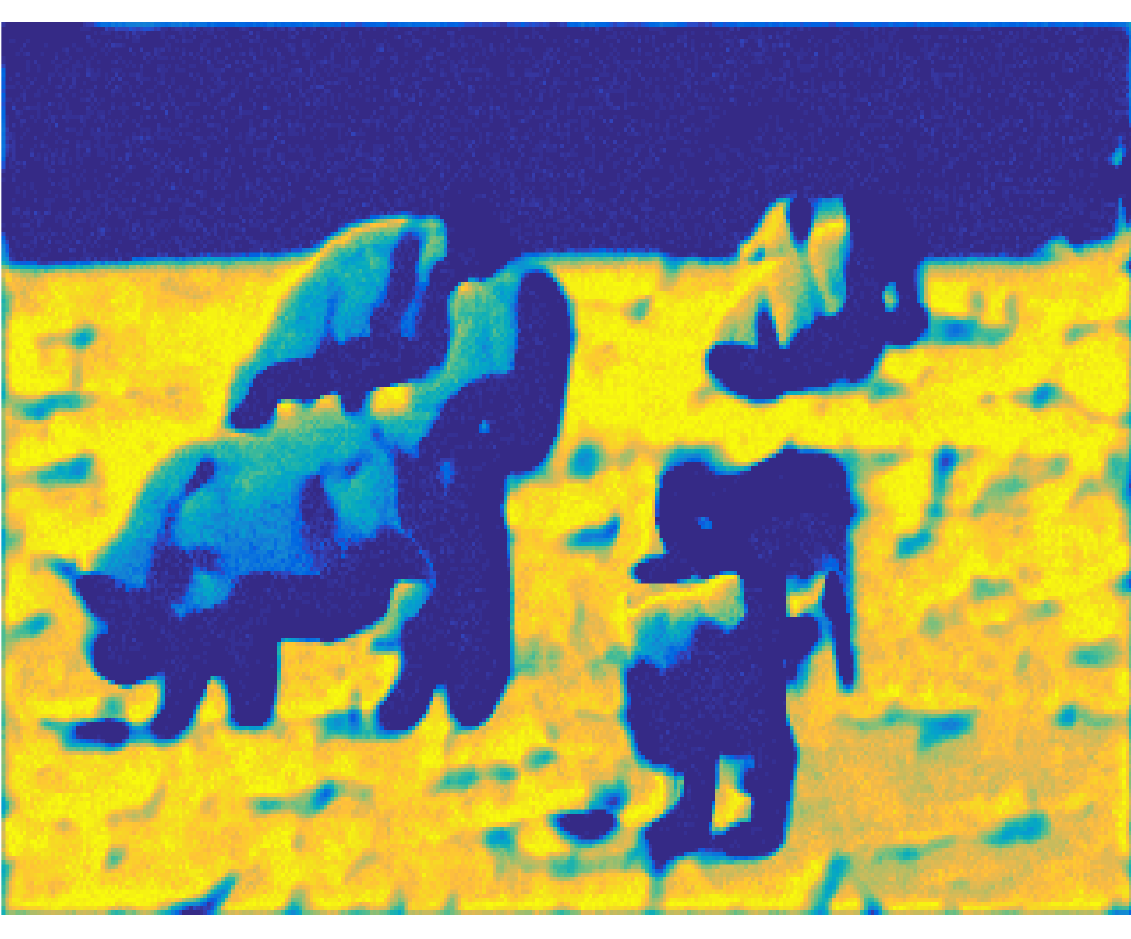}
		\captionsetup{labelformat=empty,skip=0pt}
		\caption{(b) ``grass''}
	\end{subfigure}
	\begin{subfigure}[t]{0.13\textwidth}
		\centering
		\includegraphics[width=1\linewidth]{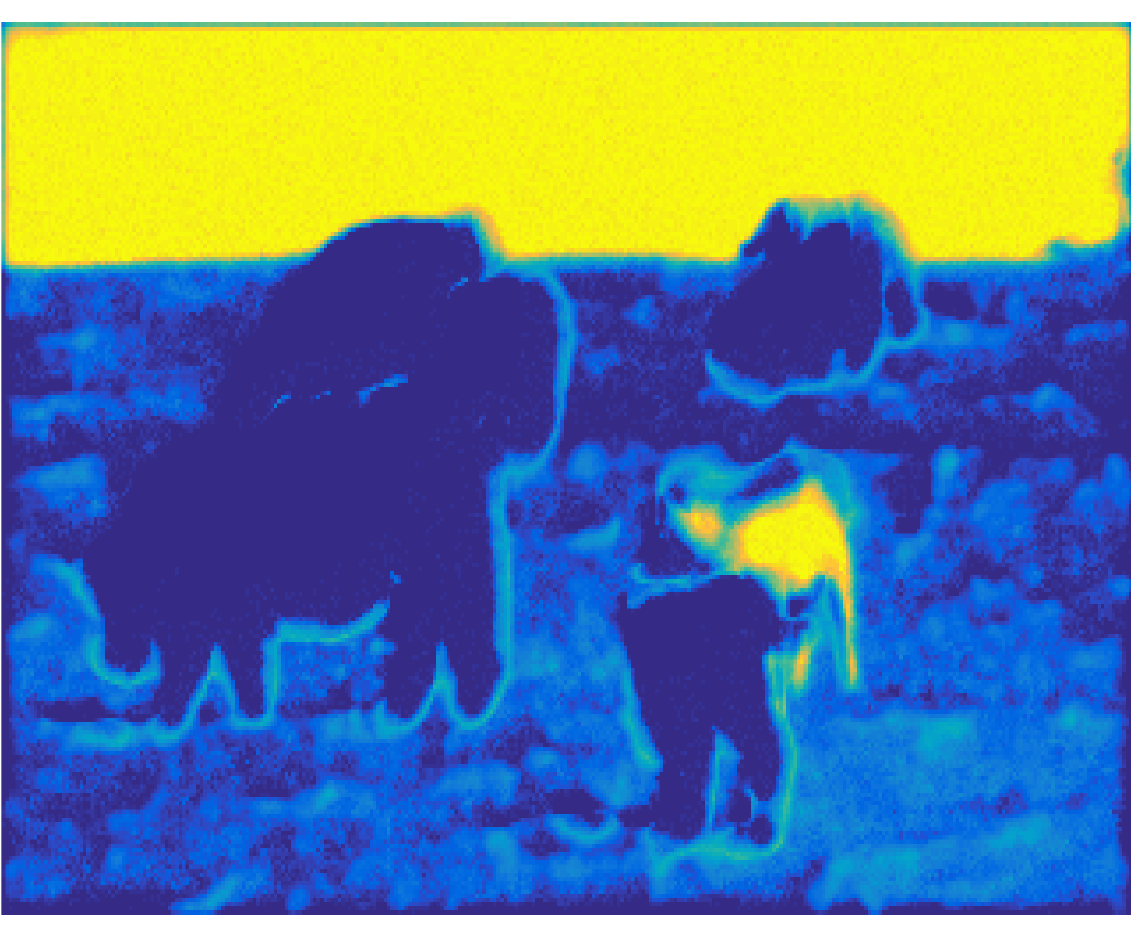}
		\captionsetup{labelformat=empty,skip=0pt}
		\caption{(c) ``sky''}
	\end{subfigure}
	\begin{subfigure}[t]{0.13\textwidth}
		\centering
		\includegraphics[width=1\linewidth]{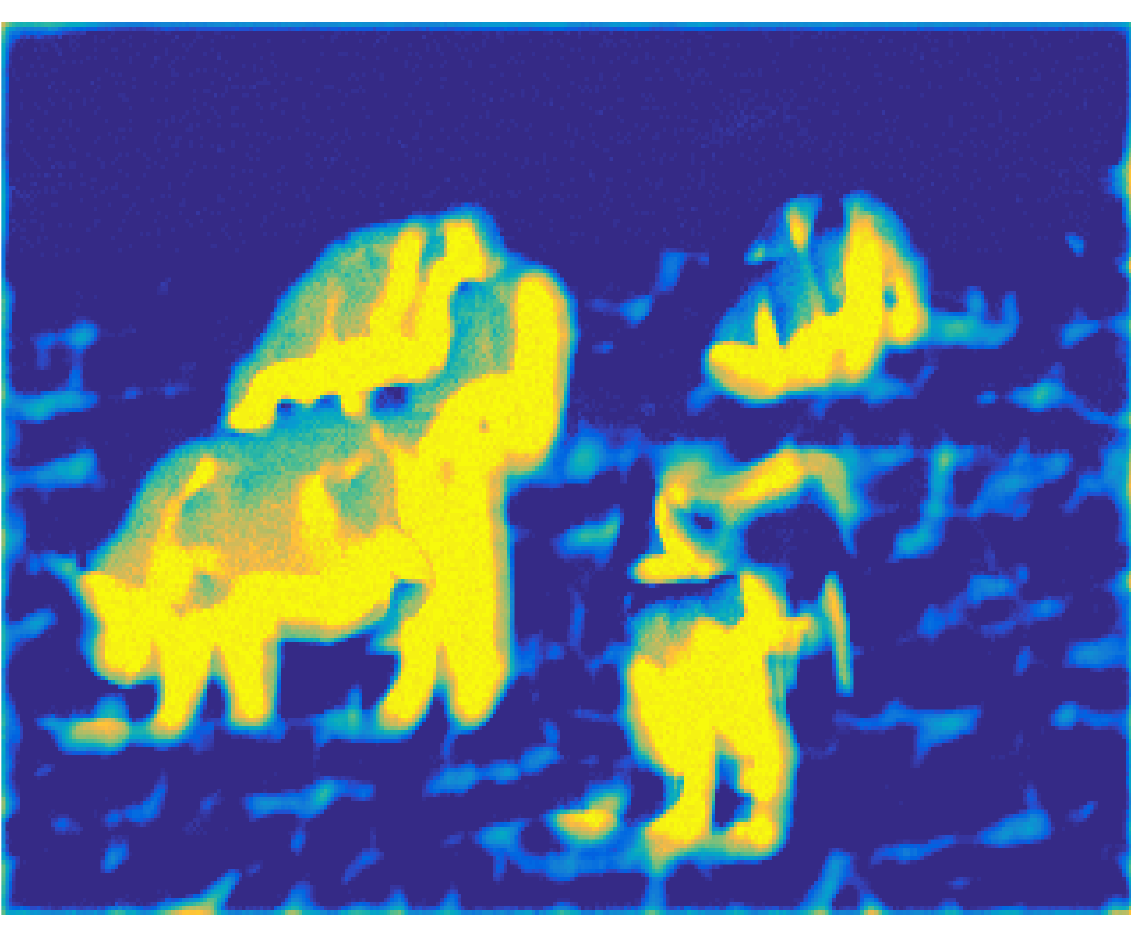}
		\captionsetup{labelformat=empty,skip=0pt}
		\caption{(d) ``cow''}
	\end{subfigure}
	\hspace{-0.2cm}
	\begin{subfigure}[t]{0.029\textwidth}
		\centering
		\includegraphics[width=0.58\linewidth]{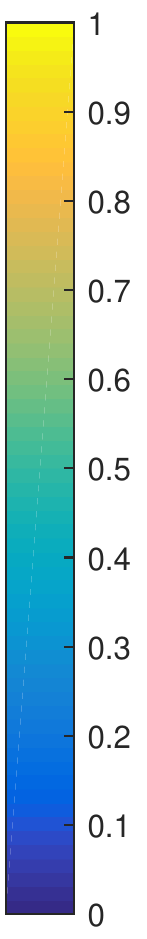}
		\captionsetup{labelformat=empty,skip=0pt}
		\caption{}
	\end{subfigure}
	\begin{subfigure}[t]{0.13\textwidth}
		\centering
		\includegraphics[width=1\linewidth]{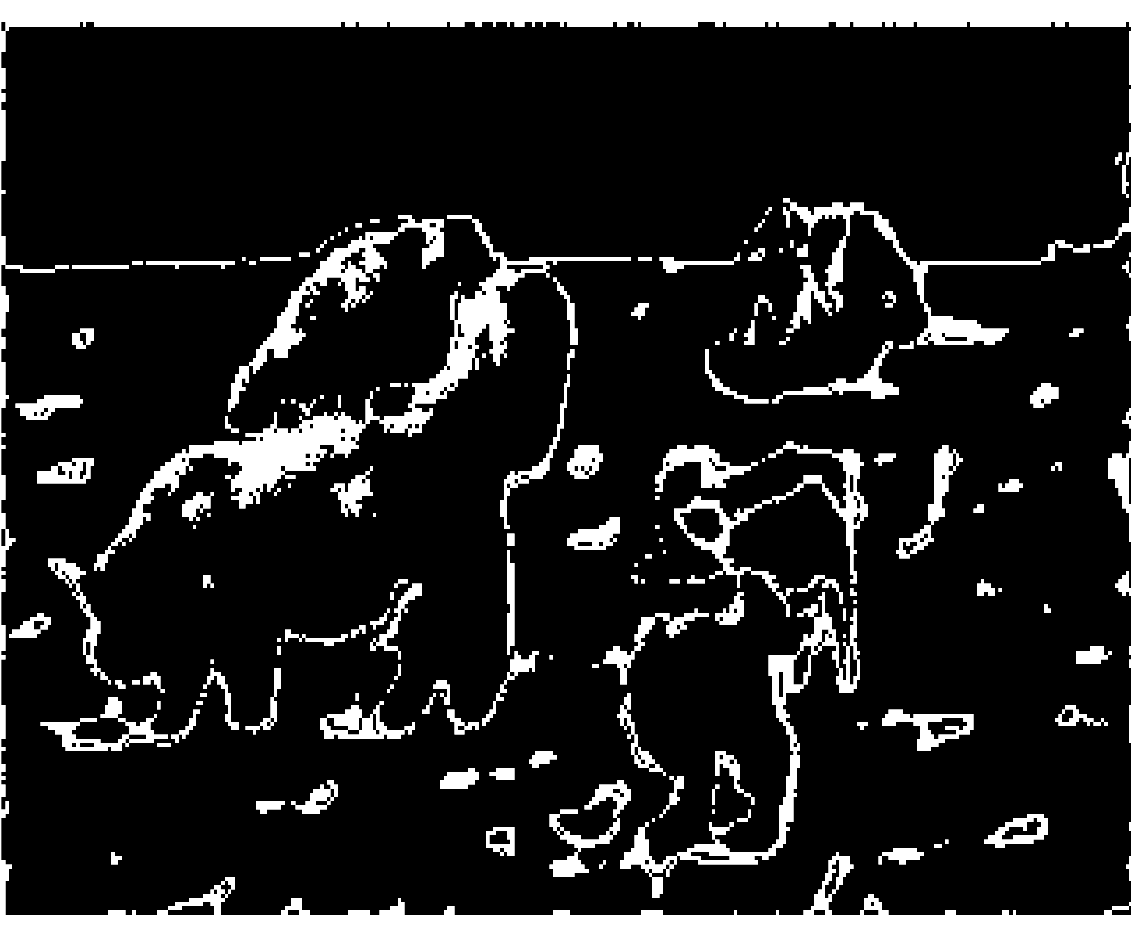}
		\captionsetup{labelformat=empty,skip=0pt}
		\caption{(e) transition}
	\end{subfigure}
	\begin{subfigure}[t]{0.13\textwidth}
		\centering
		\includegraphics[width=1\linewidth]{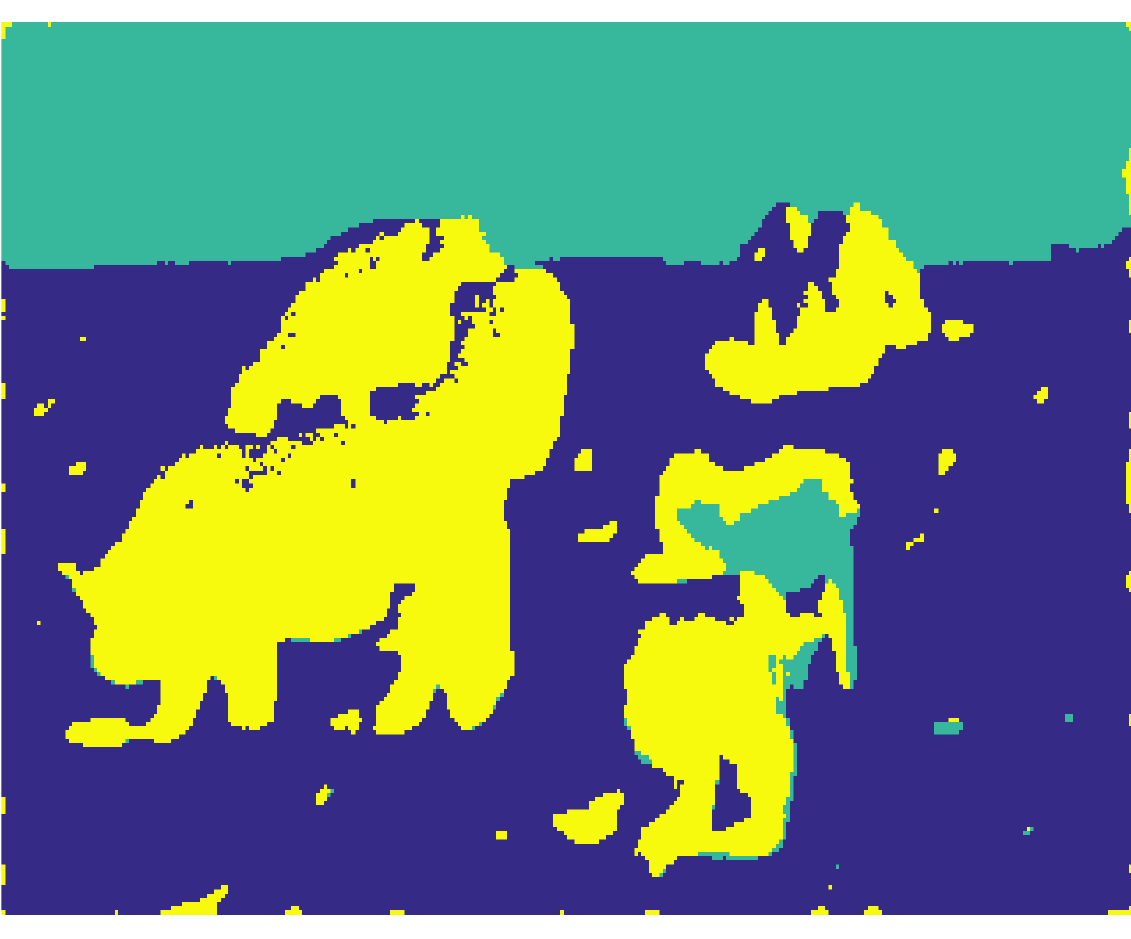}
		\captionsetup{labelformat=empty,skip=0pt}
		\caption{(f) max}
	\end{subfigure}
	\begin{subfigure}[t]{0.13\textwidth}
		\centering
		\includegraphics[width=1\linewidth]{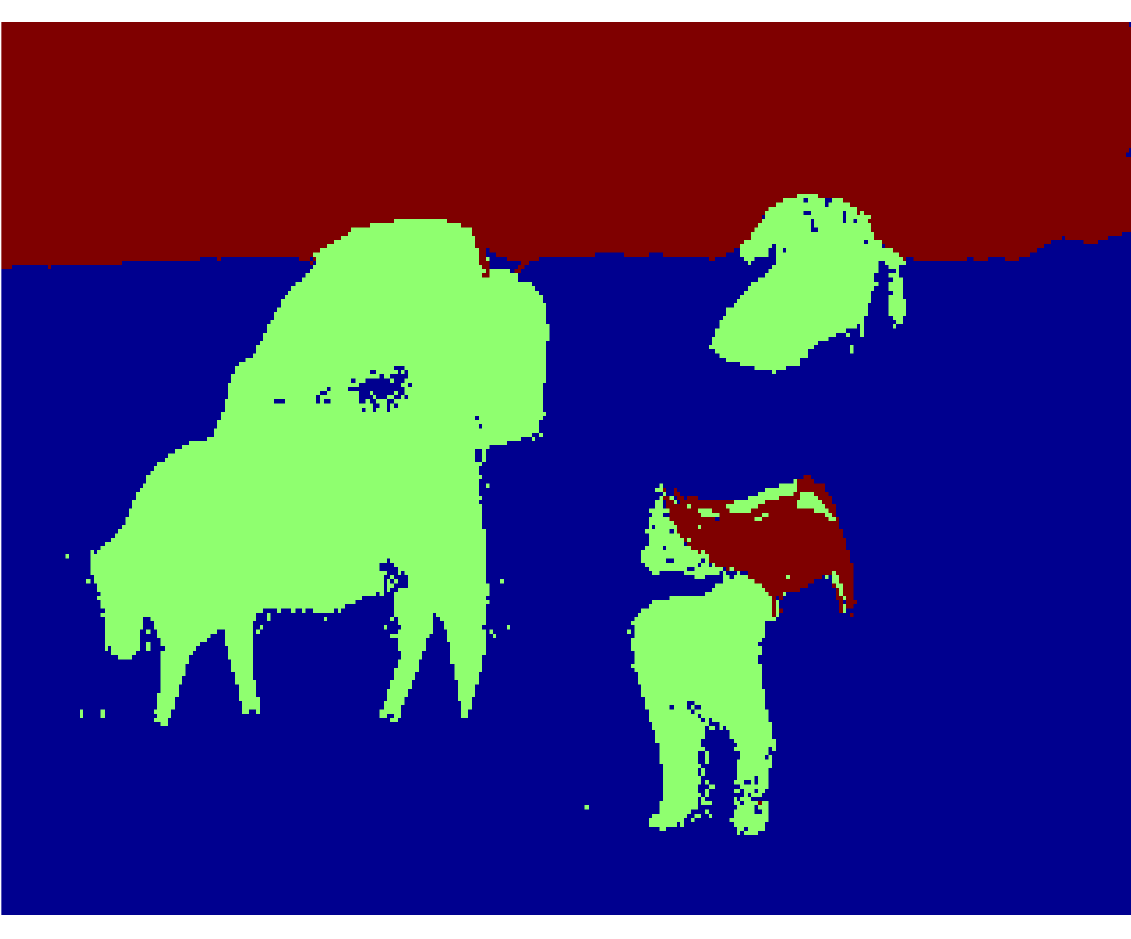}
		\captionsetup{labelformat=empty,skip=0pt}
		\caption{(g) LDA}
	\end{subfigure}

	\begin{subfigure}[t]{0.13\textwidth}
		\centering
		\includegraphics[width=1\linewidth]{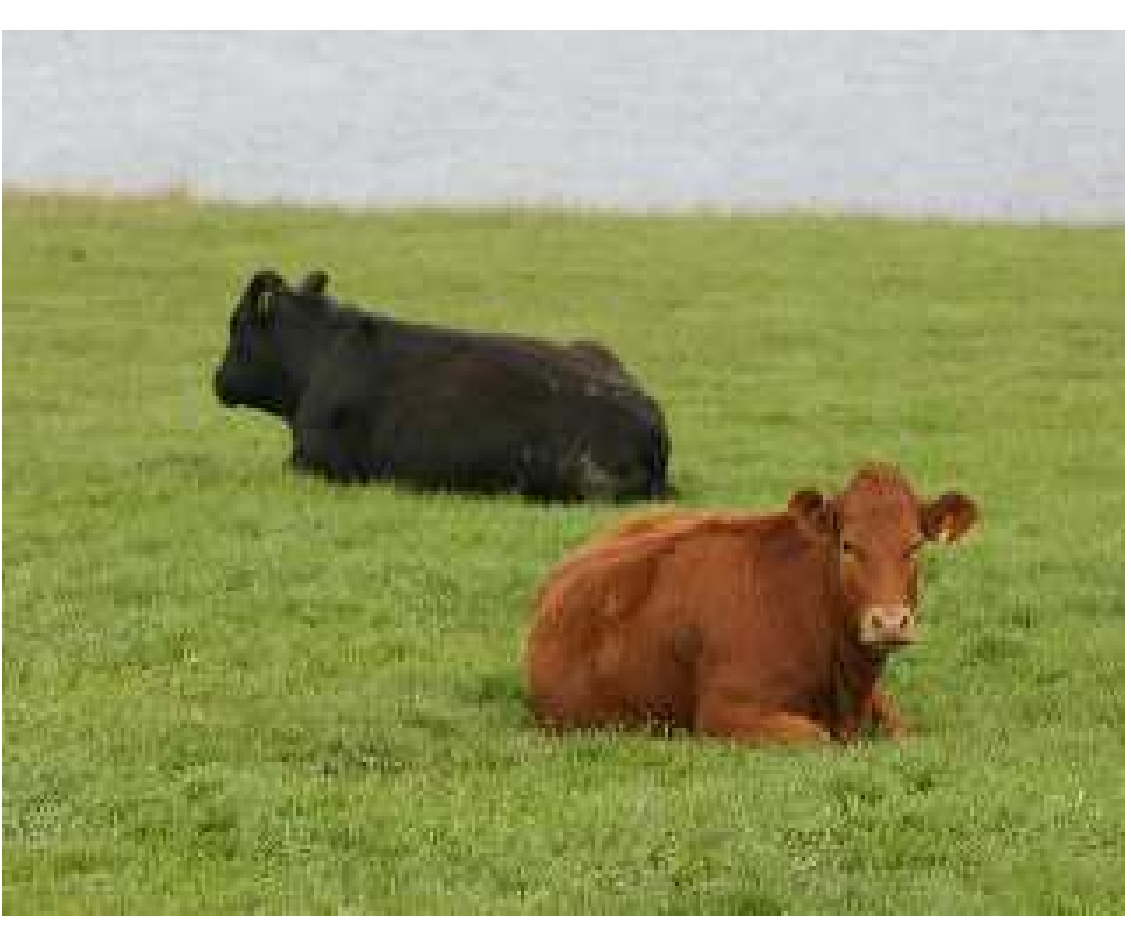}
		\captionsetup{labelformat=empty,skip=0pt}
		\caption{(a) cow2}
	\end{subfigure}
	\begin{subfigure}[t]{0.13\textwidth}
		\centering
		\includegraphics[width=1\linewidth]{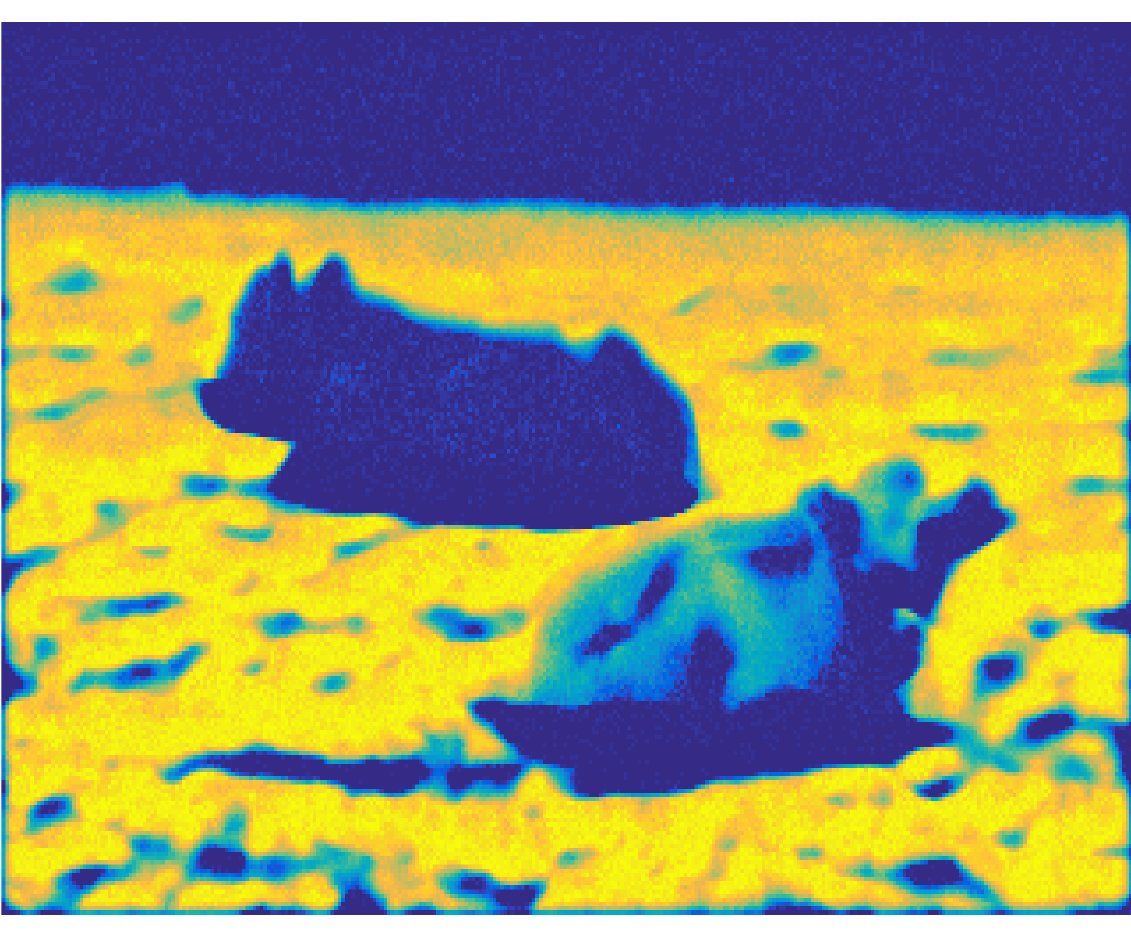}
		\captionsetup{labelformat=empty,skip=0pt}
		\caption{(t) ``grass''}
	\end{subfigure}
	\begin{subfigure}[t]{0.13\textwidth}
		\centering
		\includegraphics[width=1\linewidth]{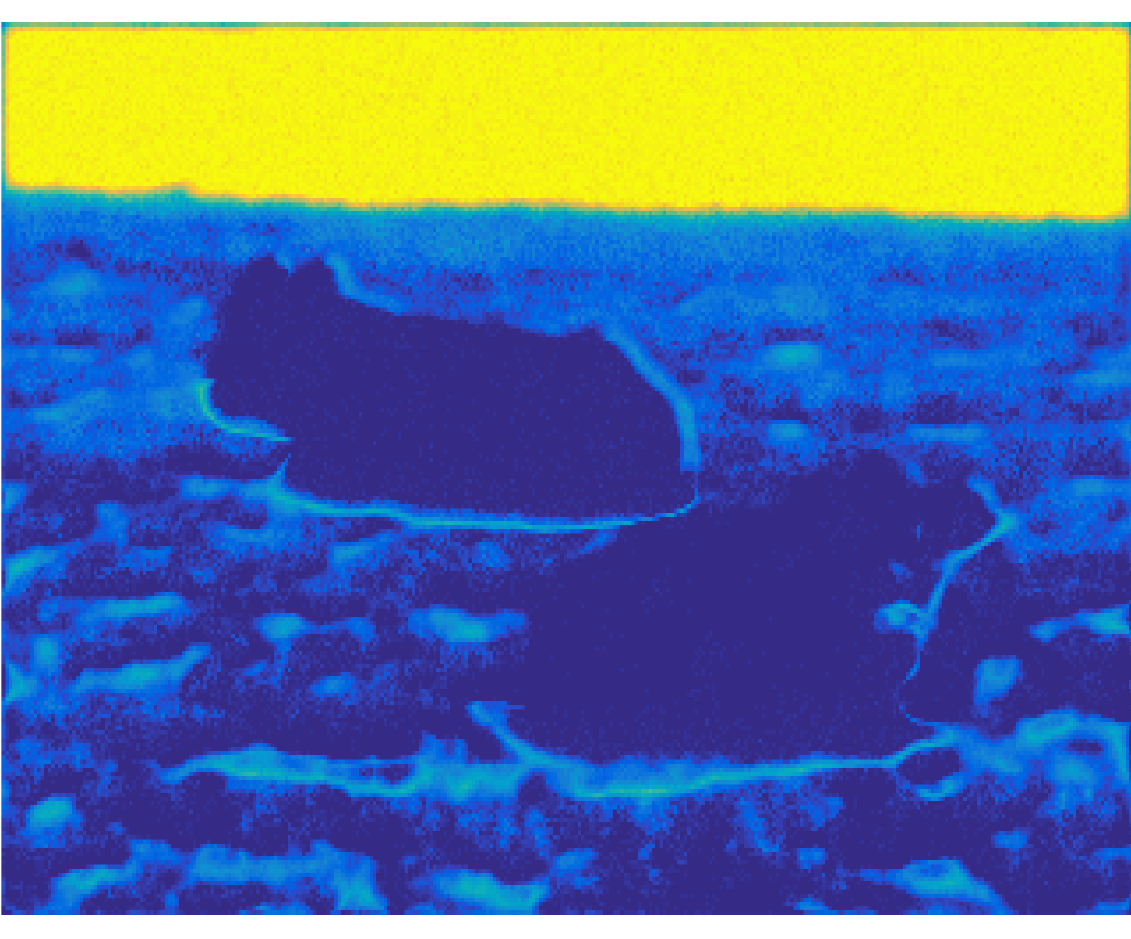}
		\captionsetup{labelformat=empty,skip=0pt}
		\caption{(c) ``sky''}
	\end{subfigure}
	\begin{subfigure}[t]{0.13\textwidth}
		\centering
		\includegraphics[width=1\linewidth]{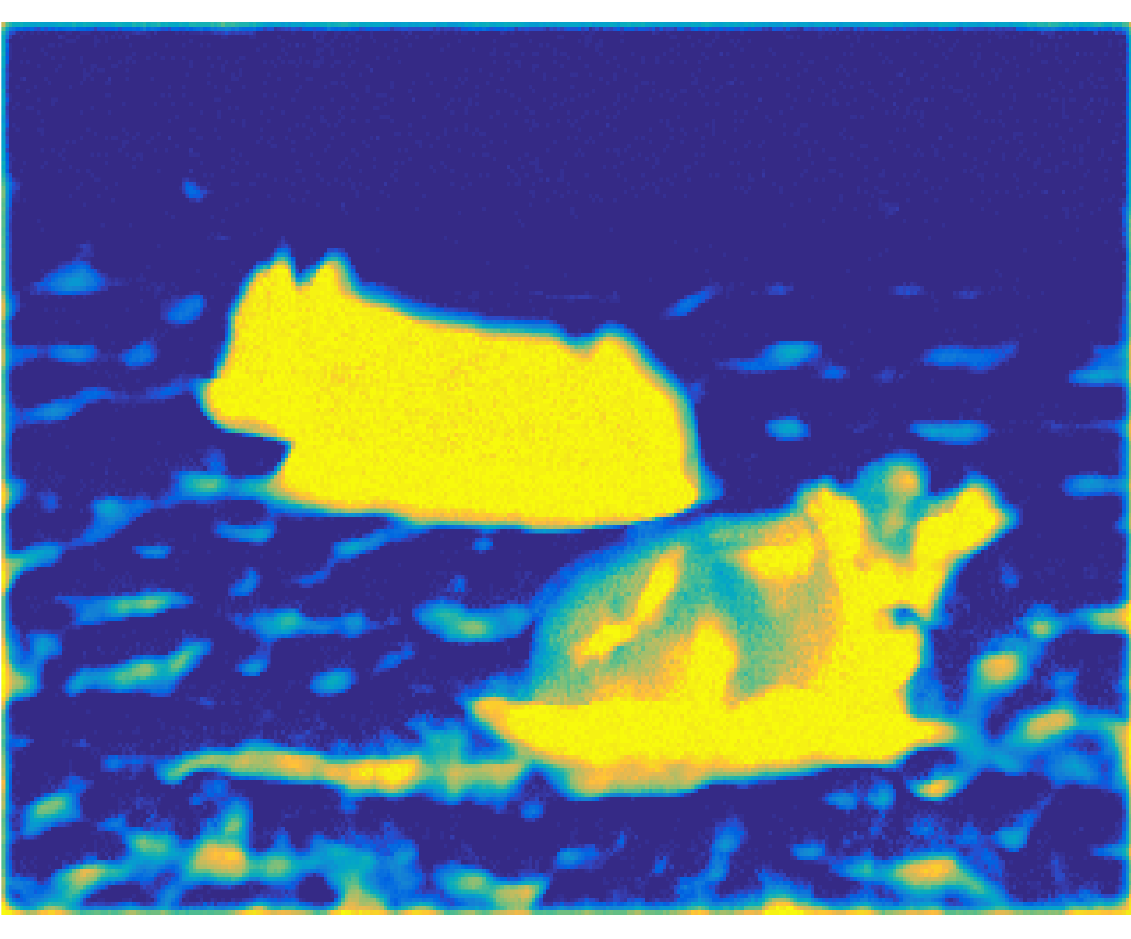}
		\captionsetup{labelformat=empty,skip=0pt}
		\caption{(d) ``cow''}
	\end{subfigure}
	\hspace{-0.2cm}
	\begin{subfigure}[t]{0.029\textwidth}
		\centering
		\includegraphics[width=0.58\linewidth]{colorbar1.pdf}
		\captionsetup{labelformat=empty,skip=0pt}
		\caption{}
	\end{subfigure}
	\begin{subfigure}[t]{0.13\textwidth}
		\centering
		\includegraphics[width=1\linewidth]{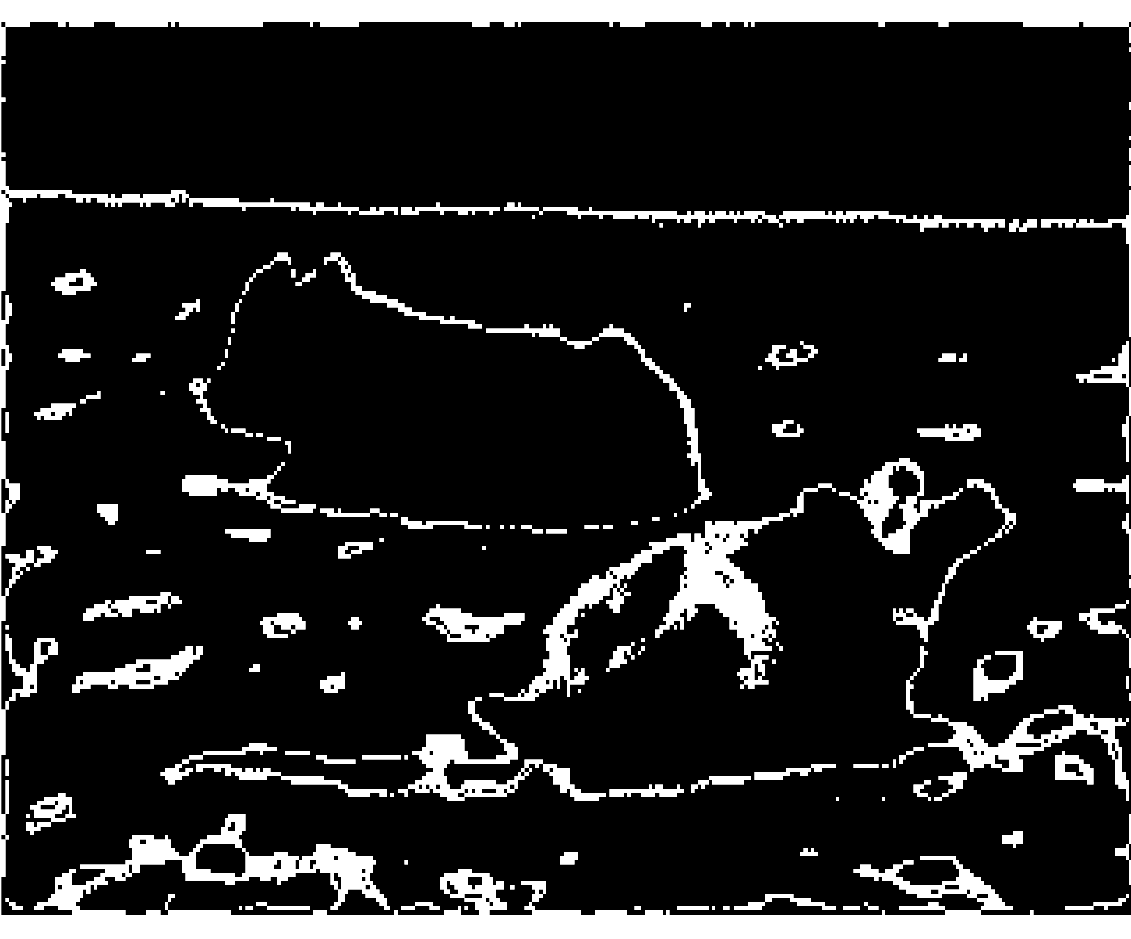}
		\captionsetup{labelformat=empty,skip=0pt}
		\caption{(e) transition}
	\end{subfigure}
	\begin{subfigure}[t]{0.13\textwidth}
		\centering
		\includegraphics[width=1\linewidth]{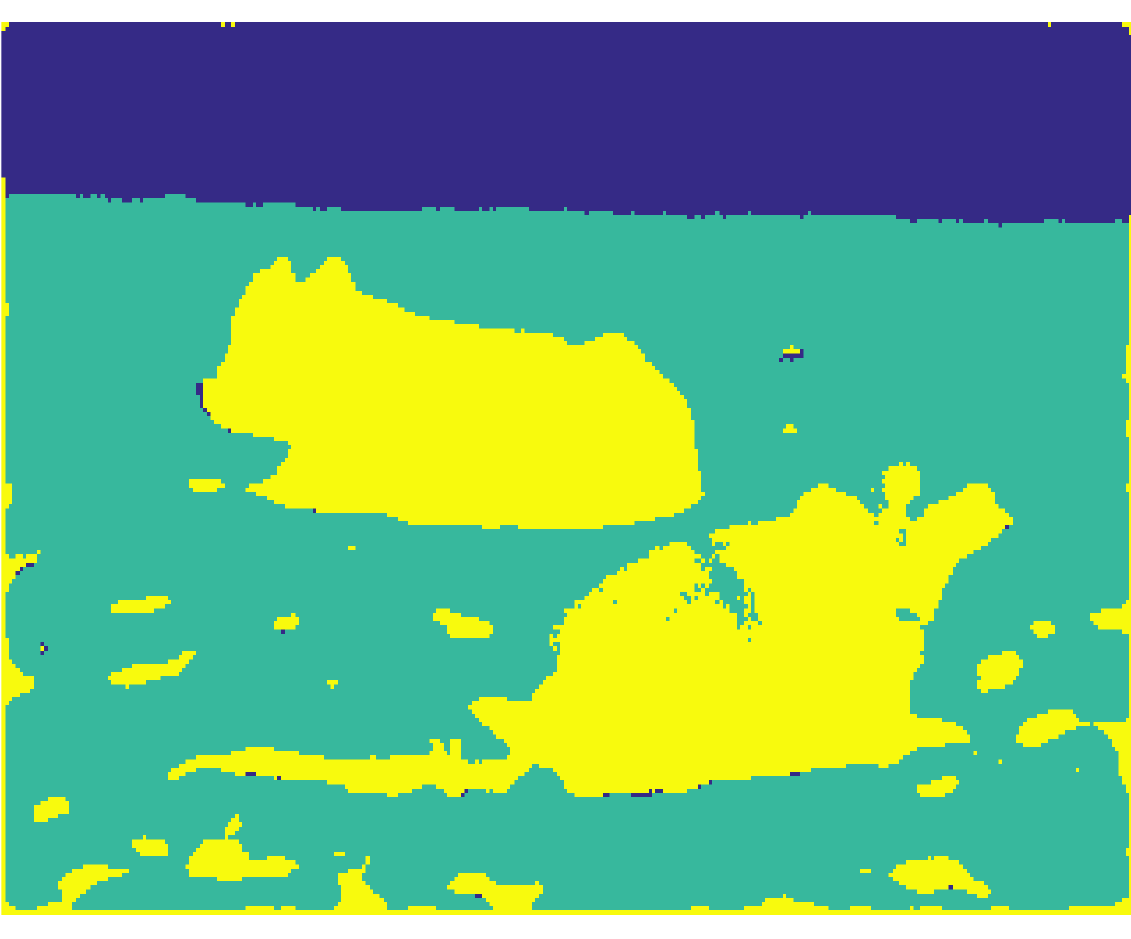}
		\captionsetup{labelformat=empty,skip=0pt}
		\caption{(f) max}
	\end{subfigure}
	\begin{subfigure}[t]{0.13\textwidth}
		\centering
		\includegraphics[width=1\linewidth]{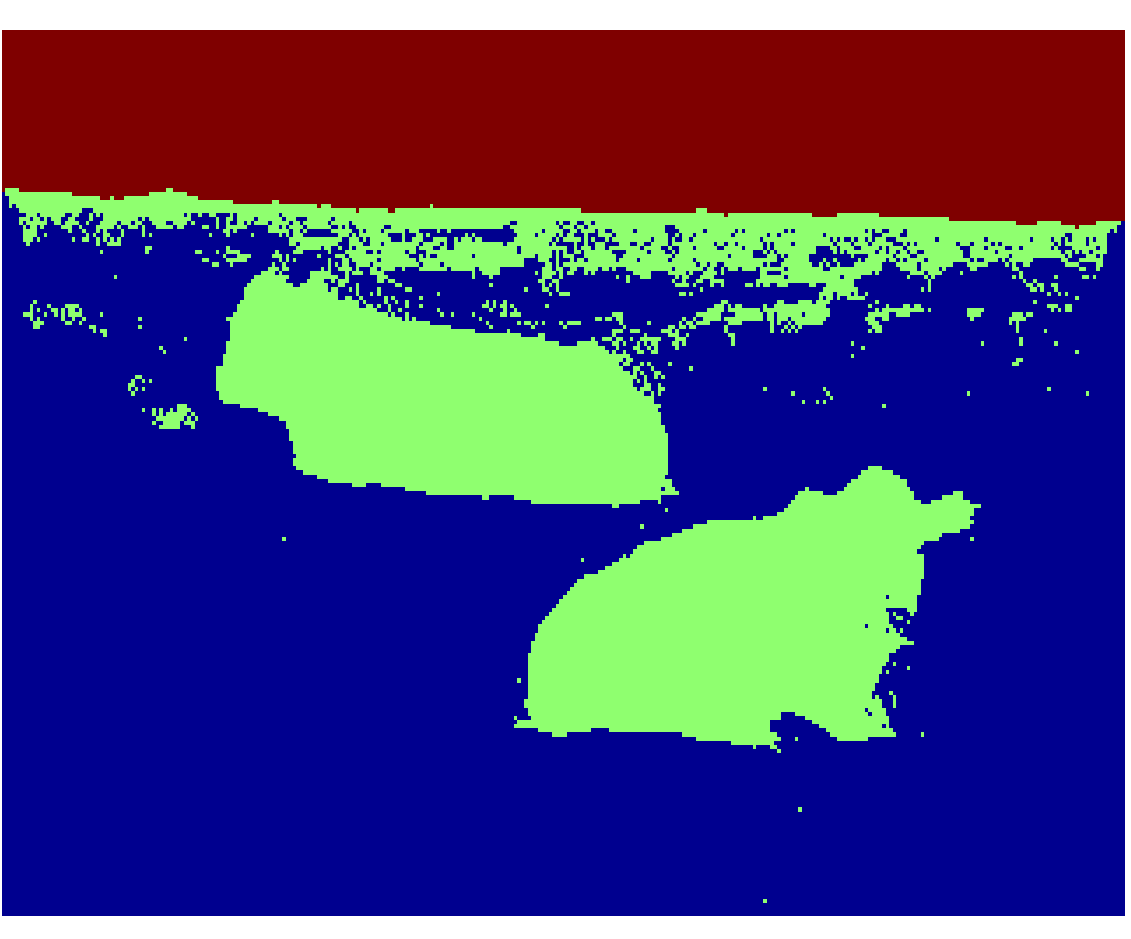}
		\captionsetup{labelformat=empty,skip=0pt}
		\caption{(g) LDA}
	\end{subfigure}
	
	\caption{Example of PM-LDA and LDA results. (a) Original image. (b)-(d) Results of PM-LDA,  the partial membership maps in ``grass'', ``sky'', and ``cow'' topics, respectively. The color indicates the degree of membership in a topic. (e) Transition regions consisting of visual words with at lease one partial membership value in range $[0.4, 0.6]$ (f) Modified segmentation result of PM-LDA by assigning each visual word to the topic with the largest membership. The colors indicate topics. (g) Result of LDA. The colors indicate topics. }
	\label{fig:cow1}
\end{figure*}

\begin{figure}[!htb]
	\centering
	\includegraphics[height=0.4\linewidth,width=0.5\linewidth]{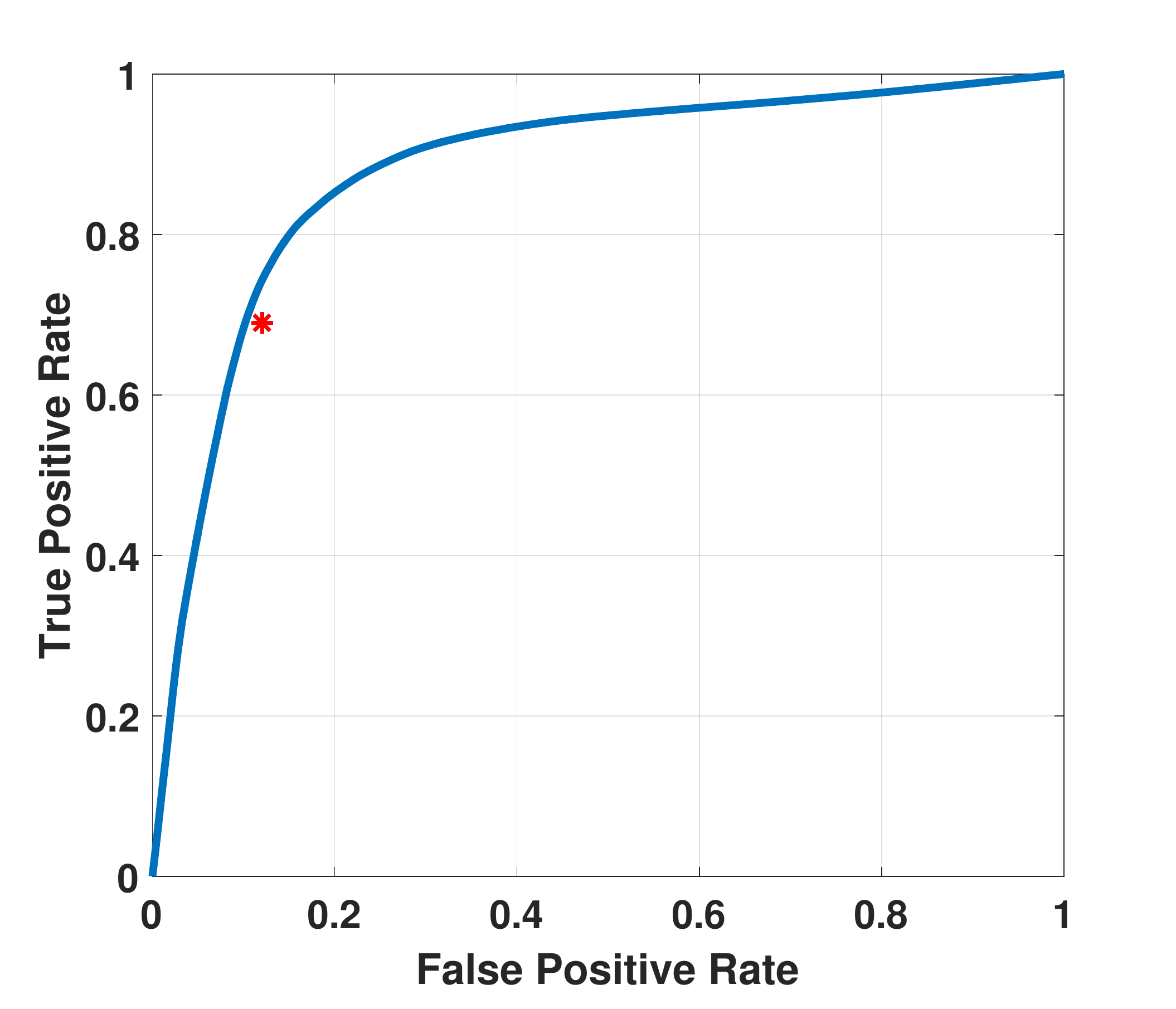}
	\caption{ROC curve of PM-LDA for cow detection evaluated at the pixel level. The red star represents the LDA cow detection results. }
	\label{fig:roc}
	\vspace{-2mm}
\end{figure}

\paragraph{Microsoft Research Cambridge data set version one (MSRCv1)} The MSRCv1 database consists of 240, $213\times320$ pixel images. A subset from this database consisting of all images that include the ``grass'', ``cow'', and ``sky'' topics was used in this experiment. The local descriptors proposed in \cite{winn:2005}, the output of a set of filter bank responses made of 3 Gaussians, 4 Laplacian of Gaussians (LoG) and 4 first order derivatives of Gaussians were used as the feature vectors. The filter window size used was $15\times 15$.  In this experiment, instead of using each image as a document, we apply normalized cuts method to get $40$ super-pixel segments from each image and treat each super-pixel as a document. The topic number is set to be $3$, and for LDA, we densely sample the filter bank output and build a dictionary of size 200. Quantitative comparison results on accurate detection of the ``cow'' class are shown in Fig. \ref{fig:roc}. ROC curve analysis of PM-LDA using the ``cow'' membership map was conducted. The red star indicates the quantitative LDA result (a ROC curve cannot be generated due to the crisp segmentation of the LDA method).  Example results are shown in Fig. \ref{fig:cow1}.  In Subfigure (e), transition regions are highlighted by indicating the pixels with at least one membership value in range $[0.4, 0.6]$. As shown in (e), these partial membership values mostly occur at the boundary between two topics. Thus, PM-LDA is able to identify when the feature vector contains information from multiple topics (as the feature vector is being computed over a window that contains more than one topic).  This is a powerful result showing the effectiveness of PM-LDA to provide semantic image understanding.  For comparison with LDA, we modified the segmentation result of PM-LDA by assigning each visual word to the topic with the largest membership. As shown in (f) and (g), PM-LDA can achieve similar results to LDA. So on these images with crisp boundaries, PM-LDA can generate binary membership values, and learn the three semantic topics comparable to LDA. Thus, PM-LDA also is effective for use in crisp labeling problems. 

{\small
\bibliographystyle{IEEETran}
\bibliography{egbib}
}

\end{document}